\RequirePackage{fix-cm}
\documentclass[smallcondensed]{svjour3}     
\smartqed  
\usepackage{xfrac}
\usepackage{graphicx}
\usepackage{hhline}
\usepackage{graphicx}
\usepackage{adjustbox}
\usepackage[table,xcdraw]{xcolor}
\usepackage{amssymb}
\usepackage{mathtools}

\usepackage{multirow}
\usepackage{bm}
\usepackage{subfig}

\usepackage[titletoc,title]{appendix}

\usepackage{slashbox}

\usepackage{hhline}

\usepackage[pagebackref=true,breaklinks=true,colorlinks,bookmarks=false]{hyperref}

\newcommand{\softmaxt}{SoftMax$^{\mathcal{T}}$}

\begin{document}

\title{Fully Convolutional Open Set Segmentation 
}


\author{Hugo Oliveira \and Caio Silva \and Gabriel L. S. Machado \and Keiller Nogueira \and Jefersson A. dos~Santos 
}

\institute{H. Oliveira \and Caio Silva \and Gabriel L. S. Machado \and Keiller Nogueira \and Jefersson A. dos Santos \at
           Av. Ant\^{o}nio Carlos, 6627 -- ICEx Building,\\
           Pampulha, Belo Horizonte, Brazil\\
           \email{oliveirahugo@dcc.ufmg.br, caiosilva@dcc.ufmg.br, gabriel.lucas@dcc.ufmg.br, keiller.nogueira@dcc.ufmg.br, jefersson@dcc.ufmg.br}\\
}

\date{Received: 15/06/2020 / Accepted: }

\maketitle

\begin{abstract}
In semantic segmentation knowing about all existing classes is essential to yield effective results with the majority of existing approaches. However, these methods trained in a Closed Set of classes fail when new classes are found in the test phase. It means that they are not suitable for Open Set scenarios, which are very common in real-world computer vision and remote sensing applications. 
In this paper, we discuss the limitations of Closed Set segmentation and propose two fully convolutional approaches to effectively address Open Set semantic segmentation: OpenFCN and OpenPCS. OpenFCN is based on the well-known OpenMax algorithm, configuring a new application of this approach in segmentation settings. OpenPCS is a fully novel approach based on feature-space from DNN activations that serve as features for computing PCA and multi-variate gaussian likelihood in a lower dimensional space. Experiments were conducted on the well-known Vaihingen and Potsdam segmentation datasets. OpenFCN showed little-to-no improvement when compared to the simpler and much more time efficient SoftMax thresholding, while being between some orders of magnitude slower. OpenPCS achieved promising results in almost all experiments by overcoming both OpenFCN and SoftMax thresholding. OpenPCS is also a reasonable compromise between the runtime performances of the extremely fast SoftMax thresholding and the extremely slow OpenFCN, being close able to run close to real-time. Experiments also indicate that OpenPCS is effective, robust and suitable for Open Set segmentation, being able to improve the recognition of unknown class pixels without reducing the accuracy on the known class pixels.







\end{abstract}

\newcommand{\currprop}{}

\section{Introduction}
\label{sec:intro}

The development of new technologies for the acquisition of aerial images onboard satellites or aerial vehicles has made it possible to observe and study various phenomena on the Earth's surface, both on a small and large scale.
A highly requested task, in this sense, is automated geographic mapping, which gives an easier and faster approach of monitoring cities, regions, countries, or entire continents. 
Automatic mapping of remote sensing images is typically modeled as a supervised classification task, commonly known as semantic segmentation, in which a model is first trained using labeled pixels and then used to classify other pixels in a new region.
Commonly, this process is based on the Closed Set (or Closed World) assumption: it assumes that all training and testing pixels come from the same label space, i.e., train and test sets have the same set of classes.
It is easy to notice that this assumption does not hold in real-world scenarios, mainly for Earth Observation applications, such as geographic mapping, given the huge size of the images and the (possible) elevated number of distinct objects (classes).
In these scenarios, the model is likely to observe, during the prediction phase, samples of classes not seen during the training.
In these cases, Closed Set semantic segmentation methods are error-prone to unknown classes given that they will wrongly recognize it as one of the known classes during the inference.
This limits the use of such approaches to real-world Earth Observation applications, such as automated geographic mapping.


Towards solving this, Open Set Recognition (OSR) can be described as the set of algorithms that address the problem of identifying, during the inference phase, samples of unknown classes, i.e., instances of classes not seen during the training.
Using the same definitions of Geng \textit{et al.}~\cite{Geng:2020} and Scheirer \textit{et al.}~\cite{Scheirer:2014}, during the inference phase, an OSR system, such as an Open Set semantic segmentation approach, should be able to correctly classify the instances/pixels of classes employed during the training (Known Known Classes (KKCs)) whereas recognizing the samples/pixels of classes \textbf{not} seen during training (Unknown Unknown Classes (UUCs)).

\begin{figure}[!ht]
    \centering
    \includegraphics[clip, trim=0.0cm 1.5cm 0.0cm 0.0cm, width=\textwidth]{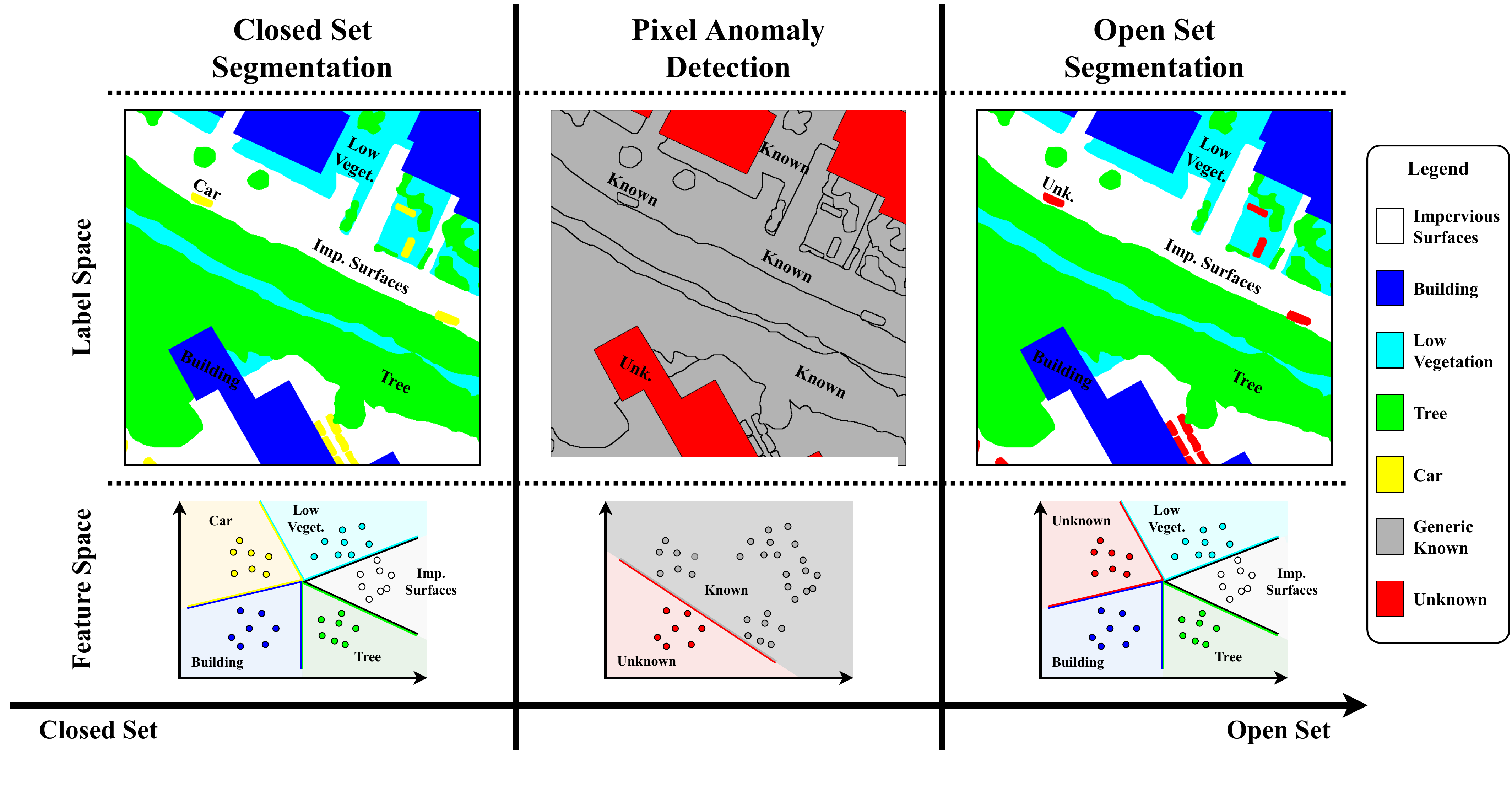}
    \caption{Graphical depiction of the problem settings of Closed Set, Anomaly Detection and Open Set Recognition in dense labeling scenarios. The $x$-axis ranges from fully Closed Set (that is, assuming full knowledge of the world) on the left side to Open Set on the rightmost example. In the middle, there is the binary task of Pixel Anomaly Detection, wherein pixels are segmented either into KKCs or UUCs without discerning between distinct KKCs. We also depict the Label Space (Semantic Segmentation map) and a sample of some pixels' representation in a 2D manifold of the Feature Space separated by labels and decision boundaries for each class.}
    \label{fig:problem_setting}
\end{figure}

By this definition, it is possible to state that the main difference between the closed and Open Set scenarios is related to the knowledge of the world, i.e., the knowledge of all possible classes. 
Specifically, while in the Closed Set scenario the methods should have full knowledge of the world, Open Set approaches must assume that they do not know all the possible classes during the training.
Obviously, different approaches may have a distinct knowledge of the world depending on the problem.
A visual example of this difference, in terms of knowledge of the world, is depicted in Figure~\ref{fig:problem_setting}.



Technically, OSR is usually achieved via transductive learning~\cite{Li:2005} by trying to infer samples from UUCs using only the test data distribution.
Another approach is to adapt Anomaly Detection techniques for OSR by using auxiliary data to try to learn a generative model via supervision, such as the Outlier Exposure (OE)~\cite{Hendrycks:2018}.
However, neither of these approaches is ideal for Open Set semantic segmentation.
This is because transduction requires continuous updates of a model in order to cope with new data, presenting an overhead that might be too expensive in most real-world applications, and the OE~\cite{Hendrycks:2018} depends on the existence and availability of auxiliary Out-of-Distribution (OOD) data for training, which may not be possible or even useful in semantic segmentation.
Therefore, an Open Set semantic segmentation algorithm must rely solely on inductive learning and use only the available training data and KKCs, while also addressing performance concerns, given the higher output dimensions of segmentation.

In this paper, we proposed two novel approaches for Open Set semantic segmentation of remote sensing image. We evaluate our methods and compare it with baselines by simulating Open Set situations in well known urban scenes such as Vaihingen and Potsdam datasets.
It is important to mention that the concept of Open Set semantic segmentation is still very little explored in the literature. The first and unique work that introduces this problem is~\cite{Silva:2020}, which proposes a method based on OpenMax~\cite{Bendale:2016} for pixelwise classification, which have many limitations concerning both effectiveness and efficiency.

The main contributions of this work are:
\begin{enumerate}
    \item The first fully convolutional methodology for semantic segmentation in RS imagery or otherwise adapted from OpenMax~\cite{Bendale:2016}, explained in Section~\ref{sec:openfcn};
    \item Proposal of a completely novel fully convolutional methodology for identifying UUCs in dense labeling scenarios using Principal Components from the internal feature space of the DNNs, further described in Section~\ref{sec:openpcs};
    \item Definition of a benchmark evaluation protocol with standard threshold-dependent and threshold-independent metrics for testing OSR Semantic Segmentation tasks;
    \item Extensive evaluations of architectures and thresholding for the proposed approaches and both Open and Closed Set baselines.
\end{enumerate}


This paper is organized in sections, as follow: Section \ref{sec:related_work} presents the related work, other papers that try to solve the Open Set semantic segmentation problem; Section \ref{sec:method} describes the proposed methods, explaining in details how it works; 
Section \ref{sec:experiments} shows the setup used for this paper, the datasets evaluated and the metrics; Section \ref{sec:results} presents the obtained results from the experimental setup; Finally, section \ref{sec:conclusion} contains the conclusion over the obtained results and the proposed method.
\section{Related Work and Background}
\label{sec:related_work}

Convolutional Neural Networks (CNNs) have become the backbone of visual recognition for the last decade. AlexNet~\cite{Krizhevsky:2012} reintroduced image feature learning, allowing for better scalability than the first CNNs (i.e. LeNet~\cite{Lecun:1998}) in order to perform inference over harder tasks (i.e. ImageNet~\cite{Deng:2009} and CIFAR~\cite{Krizhevsky:2009}). AlexNet took advantage of larger convolutional kernels in the earlier layers and contained a total of eight layers, between convolutional and fully-connected ones. VGG~\cite{Simonyan:2014} simplified CNN architectures by using the same $3\times3$ kernels in all convolutions and max-poolings for downscaling. In contrast to VGG, the GoogleNet architecture~\cite{Szegedy:2015} -- also known as Inception -- studied a diverse set of kernel sizes to enforce disentanglement in activations. Inception modules mix combinations of multiple kernel sizes and poolings. Both VGG and Inception allow for deeper networks with smaller convolutional kernels in each module, which proved to be more efficient than shallower networks with larger convolutions, at least up to around 20 layers.

It was observed that adding layers beyond a total of 20 was detrimental to the training of CNNs, as the gradients did not reach the earlier layers, effectively preventing their training. Residual Networks (ResNets)~\cite{He:2016} based on residual identity functions that allow for shortcuts in the backpropagation were then introduced. ResNets with between 18 and 151 convolutional blocks were investigated by He \textit{et al.}~\cite{He:2016}, with little benefit being observed beyond that. With time, some tweaks were proposed to the standard ResNet architecture, with the more noteworthy ones being Wide ResNets (WRNs)~\cite{Zagoruyko:2016} and ResNeXt~\cite{Xie:2017}, which yielded considerable improvements to the traditional ResNets. However, ResNets were observed to be highly inefficient, as the activations of most convolutions throughout the network could be dropped with little-to-no effect on classification performance~\cite{Srivastava:2015}. Densely Connected Convolutional Networks (DenseNets)~\cite{Huang:2017} improved on the parameter efficiency of ResNets by replacing the identity function by concatenation and adding bottleneck and transition layers, which lowered the parameter requirements of the architecture. Huang \textit{et al.}~\cite{Huang:2017} tested DenseNet with between 121 and 264 layers, observing them to be more efficient than ResNets in both parameter and flops, when similar error values were compared in the validation set.

The remainder of this section presents the main concepts to the understand of this work and the recent literature about semantic segmentation (Section~\ref{sec:related_segmentation}) and Open Set recognition (Section~\ref{sec:related_open_set}).

\subsection{Deep Semantic Segmentation}
\label{sec:related_segmentation}

\begin{figure}[!ht]
    \centering
    \includegraphics[width=\textwidth]{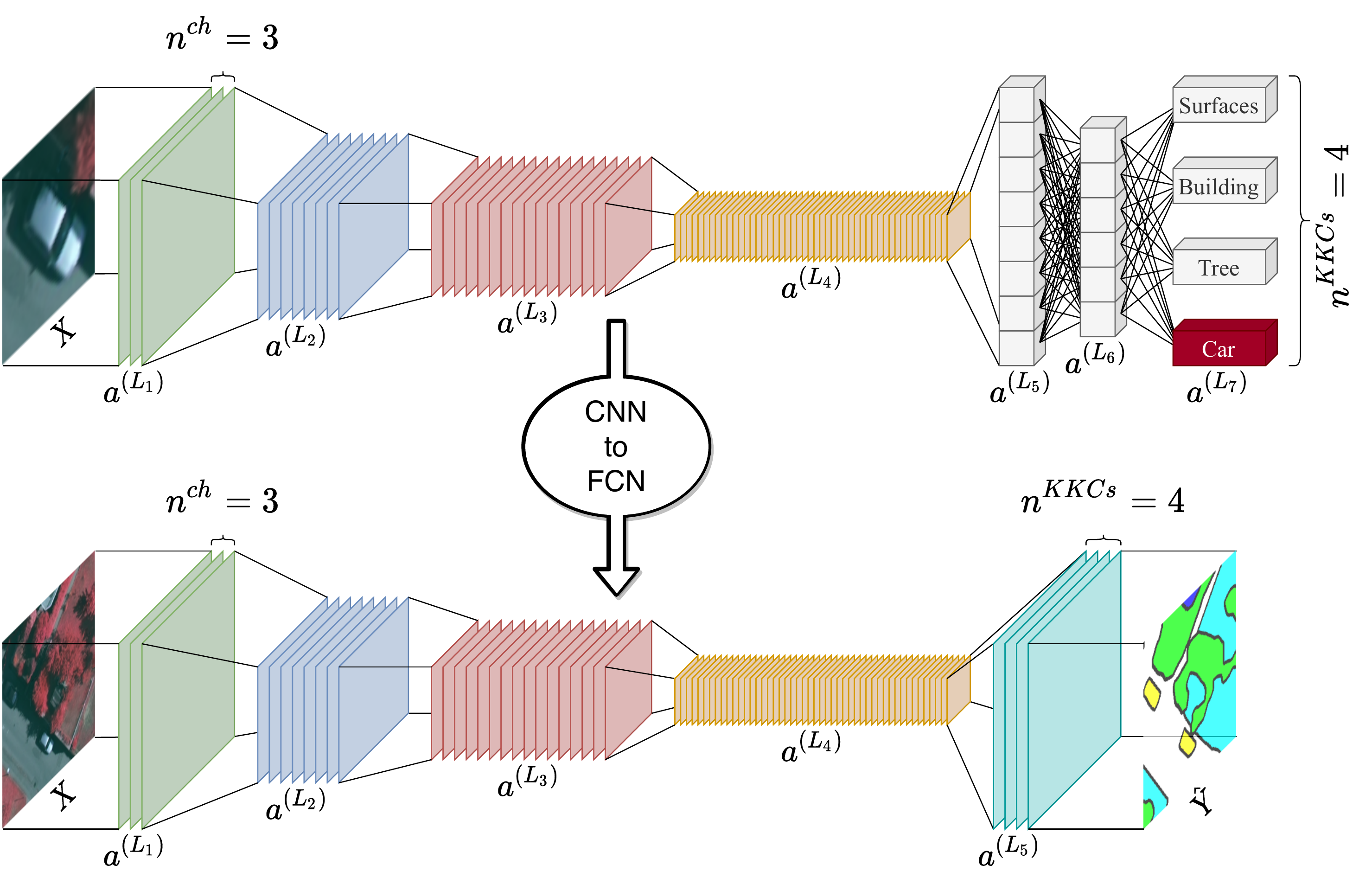}
    \caption{Architecture example of a CNN for image classification and its equivalent FCN architecture with the same backbone for semantic segmentation. Activations from layer $l$ are depicted as $a^{(l)}$ for all layers in the network ($L_{1}$ through $L_{7}$ for the CNN and $L_{1}$ through $L_{5}$ for the FCN). One should notice that in both architectures the input layer $a^{(L_{1})}$ has the number of channels $n^{ch}$ depending on the input data's number of channels (in the case of RGB images, $n^{ch} = 3$). In the CNN, the number of neurons in the output layer must match the number of KKCs ($n^{KKCs}$) in the data. Equivalently, in the FCN, the number of channels in the output layer $n^{KKCs}$, as suggested by the notation, depends on the number of KKCs of the dataset. In this example, $n^{KKCs} = 4$ as there are 4 known classes in this example.
    }
    \label{fig:cnn_fcn}
\end{figure}

CNNs~\cite{Krizhevsky:2012} are considered the state-of-the-art for sparse labeling tasks as object/scene classification due to their feature learning capabilities. The literature quickly learned to adapt CNNs for dense labeling tasks by patchwise training, using the label for the central pixel~\cite{Farabet:2012,Pinheiro:2014}. However, Fully Convolutional Networks (FCNs)~\cite{Long:2015} were shown to be considerably more efficient than patchwise training, providing the first end-to-end framework for semantic segmentation. Besides the accuracy and efficiency benefits of fully convolutional training, any traditional CNN architecture could be converted into an FCN by adding a bilinear interpolation to the activations and replacing the dense layers by convolutional ones, as shown in Figure~\ref{fig:cnn_fcn}. This simple scheme also allowed for transfer learning from large labeled datasets as ImageNet~\cite{Deng:2009} to relatively smaller semantic segmentation datasets as Pascal VOC~\cite{Everingham:2015} and COCO~\cite{Lin:2014}. FCNs are also shown to benefit from skip connections that merge the high semantic level activations at the end of the network with the high spatial resolution information from earlier layers~\cite{Long:2015}.

More recently, several semantic segmentation methods have been proposed specifically to deal with specific aspects of remote sensing images such as spatial constraints~\cite{Nogueira:2016,Maggiori:2017,marmanis2016classification,wang2017gated,Audebert:2016,Nogueira2019:TGRS} or non-RGB data~\cite{Kanan2018:IJPRS,Guiotte2020:GRSL}. 
Nogueira et al.~\cite{Nogueira:2016} use patchwise semantic segmentation in RS imaging for both urban and agricultural scenarios. 
Maggiori et al.~\cite{Maggiori:2017} proposed a multi-context method based on upsampled and concatenated features extracted from distinct layers of a fully convolutional network.
In~\cite{marmanis2016classification}, the authors proposed multi-context methods that combine boundary detection with deconvolution networks.
In~\cite{Audebert:2016}, the authors fine-tuned a deconvolutional network using $256\times256$ fixed size patches.
To incorporate multi-context knowledge into the learning process, they proposed a multi-kernel technique at the last convolutional layer.
Wang et al.~\cite{wang2017gated} proposed to extract features from distinct layers of the network to capture low- and high-level spatial information. 
In~\cite{Kanan2018:IJPRS}, the authors adapt state-of-the-art semantic segmentation approaches to work with multi-spectral images. Guiote et al.~\cite{Guiotte2020:GRSL} proposed an aprooach for semantic segmentation from LiDAR point clouds.

\subsection{Open Set Recognition}
\label{sec:related_open_set}

The Open Set recognition problem was first introduced by Scheirer et al.~\cite{Scheirer:2012}. They discussed about the notion of ``openness'', that occurs when we do not have knowledge of the entire set of possible classes during supervised training, and must account for unknowns during predicting phase. 
The first studies and applications involving Open Set recognition were adaptations of ``shallow'' methods that acted in the feature space of visual samples and consisted mainly of threshold-based or support-vector-based methods~\cite{Scheirer:2012}. More recent work has extended the concept to deep neural networks~\cite{Bendale:2016,Cardoso2017:ML}.

A recent survey by Geng \textit{et al.}~\cite{Geng:2020} splits Open Set methods mainly between discriminative and generative approaches. Discriminative approaches usually use 1-vs-All Support Vector Machines (SVMs)~\cite{Scheirer:2012} in order to delineate the space between valid samples from the training classes and outliers -- which ideally would identify UUCs. Meta-recognition can also be used to predict failures in visual learning tasks~\cite{Scheirer:2012,Scheirer:2014}. Extreme Value Theory (EVT) is one of the most common modelings for meta-recognition using DNNs~\cite{Bendale:2016,Ge:2017,Oza:2019} for classification.

Earlier deep OSR methods~\cite{Bendale:2016,Ge:2017,Liang:2017} aimed to incorporate the prediction of UUCs directly onto the prediction of the DNN output layer. Bendale \textit{et al.}~\cite{Bendale:2016} and Ge \textit{et al.}~\cite{Ge:2017} perform this by reweighting the output probabilities of the SoftMax activation to accomodate a UUC into the prediction during test time. This approach is known as OpenMax~\cite{Bendale:2016}, and further developments to it have been proposed; for instance, aiding the computation of OpenMax with synthetic images from a Generative Adversarial Network (GAN) in G-OpenMax~\cite{Ge:2017}. OpenMax~\cite{Bendale:2016} will be further detailed in the methodology (Section~\ref{sec:openfcn}), as it is the basis for one of the proposed methods in this paper.

Inspired by adversarial attacks~\cite{Goodfellow:2014}, Out-of-Distribution Detector for Neural Networks (ODIN)~\cite{Liang:2017} insert small perturbations in the input image $x$ in order to increase the separability in the SoftMax predictions between in- and out-of distribution data ($\mathcal{D}^{in}$ and $\mathcal{D}^{out}$, respectively). This separability allows ODIN to work similarly to OpenMax~\cite{Bendale:2016} and operate close to the label space, using a threshold over class probabilities to discern between KKCs and UUCs. The manuscript reports Area Under ROC curve metrics between 0.90 and 0.99 for CIFAR-10~\cite{Krizhevsky:2009} as $\mathcal{D}^{in}$ and between 0.70 and 0.85 for CIFAR-100~\cite{Krizhevsky:2009} as $\mathcal{D}^{in}$, depending on the $\mathcal{D}^{out}$ (i.e. TinyImageNet\footnotemark\footnotetext{\url{https://tiny-imagenet.herokuapp.com/}}, LSUN~\cite{Yu:2015}, random noise, etc). Extensive hyperparameter tuning experiments are reported in the paper as well. As will be further discussed in Section~\ref{sec:openpcs}, restricting the information used for OSR to the activations in the last layers has severe limitations. Thus, modern methods have employed different strategies than simply thresholding the output probabilities to split samples into KKCs and UUCs.

A recent trend in both Anomaly Detection and OSR for deep image classification has been to incorporate input reconstruction error in supervised DNN training as a way to identify OOD samples~\cite{Yoshihashi:2019,Oza:2019,Sun:2020}. These approaches fall under the branch of generative OSR, according to the taxonomy by Geng \textit{et al.}~\cite{Geng:2020}. Classification-Reconstruction learning for Open-Set Recognition (CROSR)~\cite{Yoshihashi:2019} trains conjointly a supervised DNN for classification model ($x \rightarrow y$) and an AutoEncoder (AE) to encode the input ($x$) into an bottleneck embedding ($z$) and then decode it to reconstruct $\tilde{x}$. Conjoint training allows the DNN to optimize a compound loss function that minimizes both the classification and reconstruction errors. During the test phase, the reconstruction error magnitude between $err(x, \tilde{x}) = ||x - \tilde{x}||$ dictates if the input $x$ is indeed from the predicted class $\hat{y}$ or an OOD sample.

Class Conditional AutoEncoder (C2AE)~\cite{Oza:2019}, similarly to CROSR, uses the reconstruction error of the input ($||x - \tilde{x}||$) from an AE and EVT modeling to determine a threshold in order to discern between KKC and UUC samples. Following the same trend of thresholding a certain point in the density function of the reconstruction error from the inputs, Conditional Gaussian Distribution Learning (CGDL)~\cite{Sun:2020} uses a Variational AutoEncoder (VAE) to model the bottleneck representation of the input images according to a vector of gaussian means $\mu$ and standard deviations $\sigma$ in a lower-dimentional high semantic-level space. This modeling allows CGDL to unsupervisedly discriminate between KKCs and UUCs by thresholding the likelihood of the embedding $z_{i}$ generated from a novel sample $x_{i}$ pertaining to the multivariate gaussians $\mathcal{N}(z_{i}, \mu_{k}, \sigma^{2}_{k})$, where $k$ represents the predicted class for sample $x_{i}$.

\subsubsection{Open Set Semantic Segmentation}
\label{sec:related_open_set_segmentation}

OpenPixel~\cite{Silva:2020} is based on patchwise training of classification DNNs for image classification~\cite{Nogueira:2016} of RS images. The method builds on top of OpenMax~\cite{Bendale:2016} in order to recognize out-of-distribution pixels in urban scenarios. However, OpenPixel is highly inefficient during both training and test times due to the patchwise training using a customly built CNN.

As we have discussed in the introduction, to the best of our knowledge,
the unique work in the literature that address the Open Set segmentation problem was proposed by Silva et al~\cite{Silva:2020}. The authors introduce the concept and proposes two methods based on OpenMax~\cite{Bendale:2016} for pixelwise classification. Although promising, the approaches proposed in~\cite{Silva:2020} have several limitations both in effectiveness and efficiency aspects. In this work we have extended the OpenPixel method proposed in~\cite{Silva:2020} to be feasible in pratical situations. We better explain the improvements and adaptations in Section~\ref{sec:openfcn}.

As far as the authors are aware, there are no fully convolutional architectures for deep Open Set semantic segmentation in neither the remote sensing nor computer vision communities. Section~\ref{sec:method} bridges this gap with the proposal of two approaches based on OpenMax~\cite{Bendale:2016} (Section~\ref{sec:openfcn}) and Principal Component likelihood scoring~\cite{Tipping:1999} (Section~\ref{sec:openpcs}) in the domain of urban scene segmentation.
\section{Proposed Methods}
\label{sec:method}

This section details the two proposed methods presented in this work: 1) Open Fully Convolutional Network (OpenFCN), a fully convolutional extension of OpenMax~\cite{Bendale:2016,Silva:2020} for dense labeling tasks (Section~\ref{sec:openfcn}); and 2) Open Principal Component Scoring (OpenPCS), a novel approach that uses feature-level information to fit multivariate gaussian distributions to a low-dimensional manifold of the data in order to obtain a score based on the data likelihood for identifying failures in recognition (Section~\ref{sec:openpcs}). 

\subsection{OpenFCN}
\label{sec:openfcn}

\begin{figure}[!ht]
    \centering
    \includegraphics[width=0.95\columnwidth]{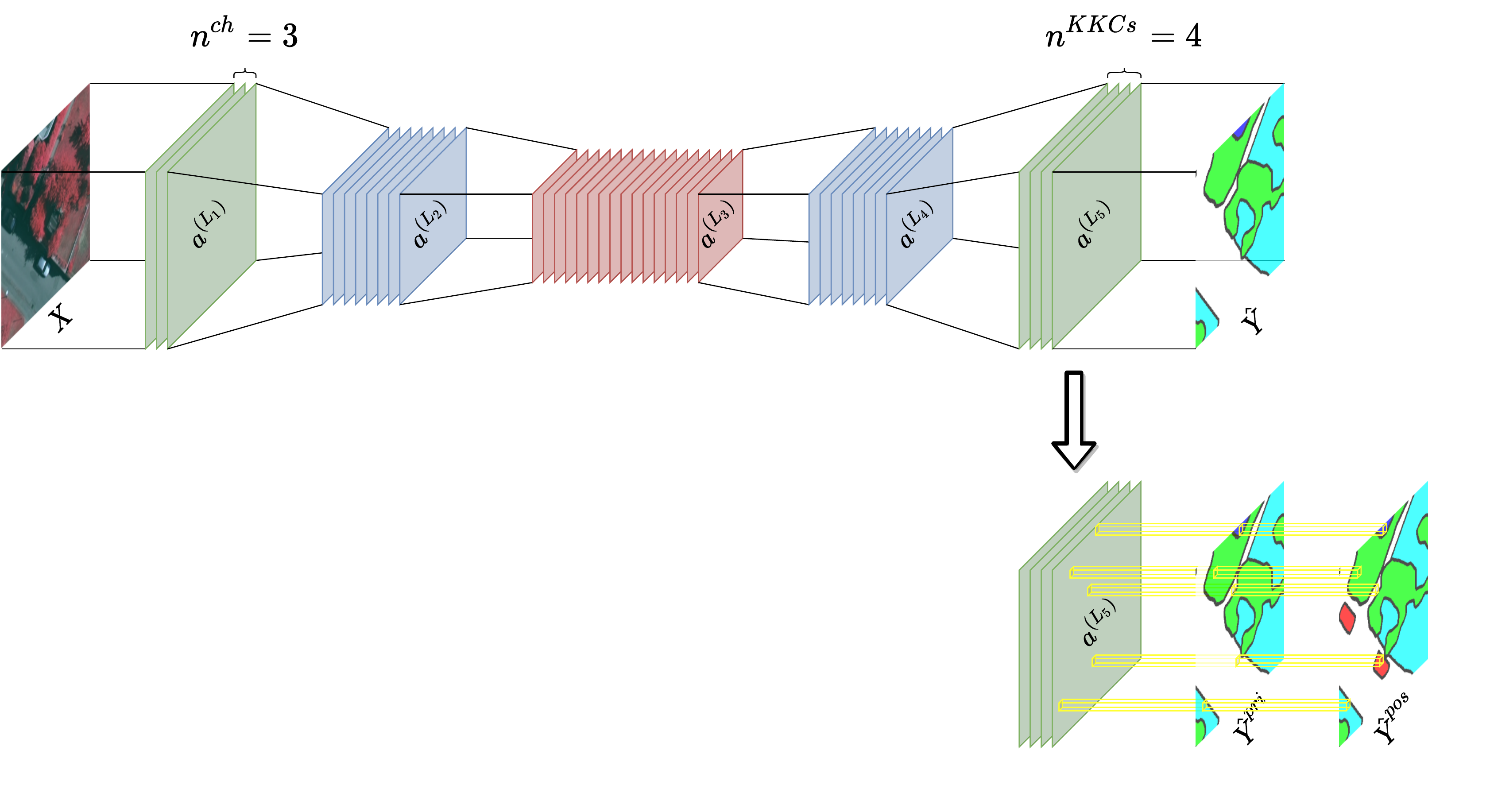}
    \caption{OpenFCN scheme for Open Set Semantic Segmentation. During training, OpenFCN behaves like the traditional Closed Set FCN, with only  Known  Known  Classes (KKCs) being fed to a supervised loss, such as Cross Entropy. OpenFCN differs from FCN only during validation and testing, when OpenMax is computed and the probabilities are thresholded in order to predict Unknown  Unknown  Classes (UUCs).}
    \label{fig:openfcn}
\end{figure}

OpenFCN relies on traditional FCN-based architectures, which are normally composed of traditional CNN backbones with the inference layers replaced by bilinear interpolation and more convolutions. As the dense prediction is treated at training time as a classification task, the distinction between OpenFCN and FCN can be seen more clearly during validation and predicting. A meta-recognition module based on OpenMax~\cite{Bendale:2016} is added to the prediction procedure of traditional FCNs, as can be seen in Figure~\ref{fig:openfcn}.

Let $\{X,\ Y\}$ be a paired set of image pixels and semantic labels from a dataset containing $C$ KKCs. A deep model $\mathcal{M}$ can be trained in a stochastic manner by feeding samples $X_{i}$ to a gradient descent optimizer as Adam~\cite{Kingma:2014} with a loss function as Cross Entropy, given by:
\begin{equation}
    \mathcal{L}_{CE}(Y, \hat{Y}^{pri}) = -Y\log{(\hat{Y}^{pri})} - (1 - Y) \log{(1 - \hat{Y}^{pri})}
    \label{eq:cross_entropy}
\end{equation}
This strategy ultimately yields an activation $a_{i} \in \mathbb{R}^{C}$ for each pixel $i$ after the last layer. Thus, $\mathcal{M}$ can be seen as a function $\mathbb{R}^{3} \rightarrow \mathbb{R}^{C}$ that converts the input space $X_{i}$ into the prior SoftMax prediction $\sigma_{i}$ for sample $i$. Obtaining the class prediction $\hat{Y}^{pri}_{i}$ can be easily done by finding the class with the larger probability in $\sigma_{i}$ across all KKCs. Thus, OpenFCNs are trained using the exact same procedure as traditional FCNs for Closed Set semantic segmentation. Posteriori predictions $\hat{Y}^{pos}$ are only computed on validation and testing, as described in the following paragraphs.

OpenMax ($\mathcal{O}$) relaxes the traditional SoftMax requirement that prediction probabilities for KKCs must add to 1, introducing an extra class to the posterior prediction $\hat{Y}^{pos}$ to the prediction set for $X$. The function $\mathcal{O}(X, Y, \mathcal{M})$ is, therefore, able to reweight the SoftMax predictions, aggregating the probability of misclassifications due to UUCs. Following the protocol of OpenMax~\cite{Bendale:2016}, during OpenFCN's validation procedure each KKC $c_{k}, k \in \{0, 1, ..., C-1\}$ yields one Weibull distribution $\mathcal{W}_{k}$. $\mathcal{W}_{k}$ is fit to the deviations from the mean $\mu_{k}$ of $a^{(L_{5})}$ according to some distance (i.e. euclidean, cosine, hybrid distances). Averages $\mu_{k}$ are computed in the validation set according only to the correctly classified pixels of class $c_{k}$.

\begin{figure}[!ht]
    \centering
    \includegraphics[width=0.8\textwidth, clip, trim=0.8cm 0cm 0.8cm 0cm]{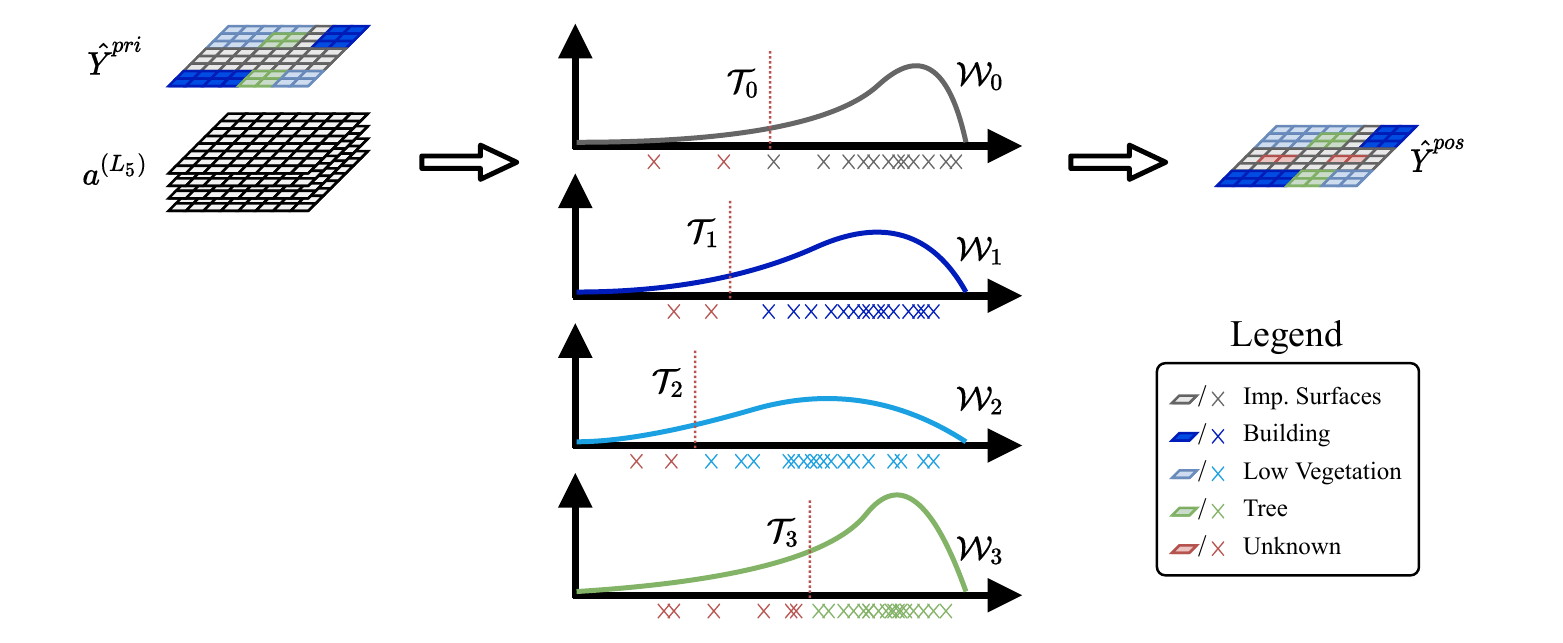}
    \caption{OpenFCN's generative modeling for detecting UUC samples. A Weibull model $\mathcal{W}_{c}$ for KKC $c$ is fit using the last layer's activations ($a^{(L_{5})}$) from correctly classified samples of this class in the training set. According to the CDF for each class' Weibull distribution, a threshold $\mathcal{T}_{c}$ is set and samples that do not reach this threshold are classified as pertaining from a UUC.}
    \label{fig:openfcn_generative}
\end{figure}

Finally, in order to identify Out-of-Distribution (OOD) samples, quantiles from the Cumulative Distribution Function (CDF) for $\mathcal{W_{k}}$ are computed, with all pixels in the set $\hat{Y}^{pri}$ predicted to be from $c_{k}$ and with less confidence than $\mathcal{T}_{k}$ being attributed to be from a UUC. Thus, the posterior OpenMax prediction $\hat{Y}^{pos}_{i}$ for a specific pixel $X_{i}$ is given by:
\begin{equation}
    \hat{Y}^{pos}_{i} =
    \begin{cases*}
        c_{k}\text{, if }max(a^{(l)}_{i}) \ge \mathcal{T}_{k} \\ 
        c_{unk}\text{, if }max(a^{(l)}_{i}) < \mathcal{T}_{k},
    \end{cases*}
    \label{eq:posterior_openfcn}
\end{equation}
where $l$ is the output layer in a DNN and $c_{unk}$ is the identifier for the UUC in the Open Set scenario. This scheme is shown in Figure~\ref{fig:openfcn_generative}.

\begin{figure}[!ht]
    \centering
    \includegraphics[width=\textwidth]{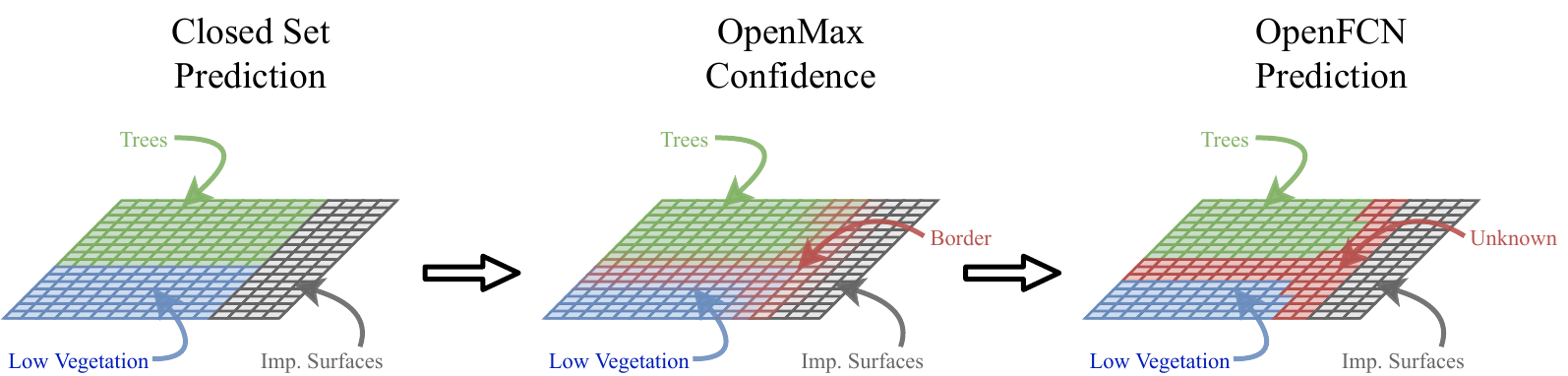}
    \caption{Depiction of OpenFCN's prediction confidence degradation on object boundaries due to the use of information close to the label space. SoftMax and OpenMax probabilities on dense labeling tasks are naturally lower on boundary regions between objects from distinct classes.}
    \label{fig:openfcn_borders}
\end{figure}

Dense labeling tasks (i.e. instance/semantic segmentation or detection) inherently have higher dimensional outputs than sparse labeling tasks (i.e. classification or single-target regression). Dense predictions also present their particular set of difficulties, including how to handle boundaries between adjacent objects. Early in our experiments, we have seen a large number of border artifacts in OpenFCN predictions. As depicted in Figure~\ref{fig:openfcn_borders}, adjacent areas between objects with distinct classes naturally yield class predictions with smaller certainties than the central pixels of these objects, resulting in warped Weibull distributions. This happens because the last layer in the network tries to model directly the label space distribution, retaining little-to-no information about the original input. Hence, activations from later layers in the network are more affected by object boundary uncertainties.

In order to mitigate this limitation of OpenFCNs in dense labeling tasks, we propose a completely novel method that is able to fuse information from low and high semantic level information into one single model. This method will be presented in Section~\ref{sec:openpcs}.

\subsection{OpenPCS}
\label{sec:openpcs}

OpenPCS works similarly to CGDL~\cite{Sun:2020}, but with three important differences: 1) we fit gaussian priors using not only the input images $X$, but also the intermediary activations (i.e. $a^{(L_{3})}$, $a^{(L_{4})}$, $a^{(L_{5})}$, etc); 2) we use a PCA instead of a VAE; and 3) the training is purely supervised and the low dimensional gaussians are fitted only during the validation phase, and not conjointly with the training, as CGDL.
This was all done for simplicity and aiming to ease computational  complexity, as  open  set  semantic  segmentation  is  a very recent field of research.

It is well known that the deeper a certain layer $l$ is placed in a DNN, the closer to the label space the activation features $a^{(l)}$ are~\cite{Shwartz:2017}. In fact, Shwartz and Tishby~\cite{Shwartz:2017} argue that any supervised DNN can be seen as a Markov chain of sequential tensorial representations that gradually morph the information processed by the network from the input space (in the input layer) to the label space (in the output layer). Thus, by using only the last layer's activations to fit Weibull distributions to each KKC, OpenFCNs -- and, by extension, OpenMax~\cite{Bendale:2016} -- limit themselves to work with information close to the label space.

\begin{figure}[!ht]
    \centering
    \includegraphics[width=0.95\textwidth]{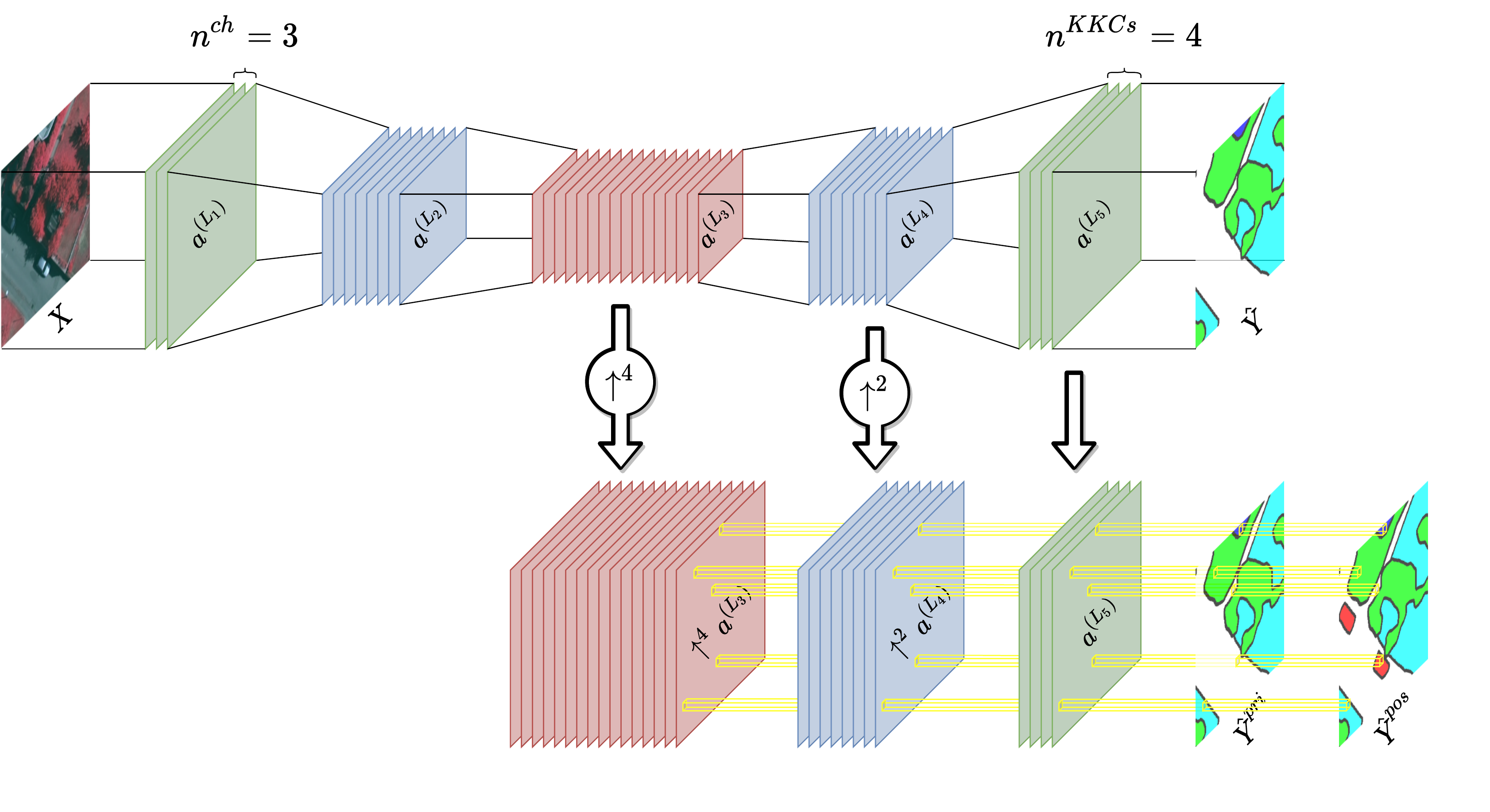}
    \caption{OpenPCS general schematics. Subsequent activations ($a^{(L_{1})}$, $a^{(L_{2})}$, ..., $a^{(L_{4})}$) from an FCN with a certain CNN backbone (as in Figure~\ref{fig:cnn_fcn}) is shown. Activations from the last layers (i.e. $a^{(L_{5})}$, $a^{(L_{4})}$ and $a^{(L_{3})}$, in this case) are concatenated to form column vectors for each predicted output pixel in $\hat{Y}^{pri}$. The prior prediction $\hat{Y}^{pri}$ is then processed according to the scheme shown in Figure~\ref{fig:openpcs_generative} using a generative model $\mathcal{G}$ in order to detect OOD pixels and, thus, classify them as unknowns.}
    \label{fig:openpcs}
\end{figure}

\begin{figure}[!ht]
    \centering
    \includegraphics[width=\textwidth, clip, trim=0.8cm 0cm 0.8cm 0cm]{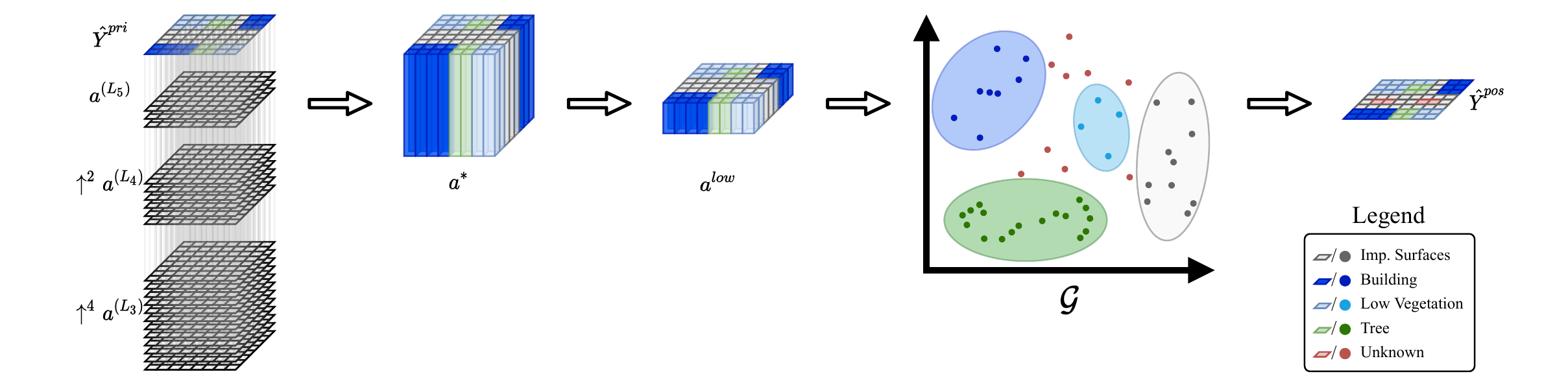}
    \caption{Graphical scheme for the generative modeling in OpenPCS. In contrast to OpenFCN's Weibull fitting, OpenPCS uses a gaussian modeling in a low-dimensional representation $a^{low}$ of the activations $a^{*}$. The Principal Components for each pixel are computed according to the concatenation of activations $a^{*}$ from multiple layers (i.e. $a^{(L_{3})}$, $a^{(L_{4})}$, $a^{(L_{5})}$, etc) for the corresponding region of each specific pixel. One multivariate gaussian is fit for each KKC and thresholds defined according to likelihoods from these models are used in order to identify OOD pixels.}
    \label{fig:openpcs_generative}
\end{figure}

Unlike OpenFCN, OpenPCS takes into account feature maps from earlier layers, which encode information closer to the input space, and combine them with activations from the last layers, fusing low and high semantic level information. This can be seen in Figures~\ref{fig:openpcs} and~\ref{fig:openpcs_generative} in the form of the yellow columns shown in the lower part of the image. Each output pixel in $\hat{Y}^{pri}$ gets a correspondent activation vector ($a^{*}$) made by the concatenation of earlier layer activations for the corresponding prediction map region in the channel axis. As earlier layers ($a^{(L_{4})}$ and $a^{(L_{3})}$) have lower spatial resolution due to the network's bottleneck, the activations from these layers are upscaled using the $\uparrow^{t}$ function, where $t$ indicates the number scaling factor. In the example shown in Figure~\ref{fig:openpcs}, $a^{*} = (a^{(L_{5})}, \uparrow^{2} a^{(L_{4})}, \uparrow^{4} a^{(L_{3})})$, so, for instance, if $a^{(L_{5})} \in \mathbb{R}^{4 \times MN}$, $\uparrow^{2} a^{(L_{4})} \in \mathbb{R}^{8 \times MN}$ and $\uparrow^{4} a^{(L_{3})} \in \mathbb{R}^{16 \times MN}$, then $a^{*}$ has dimensionality $28 \times MN$. The concatenated feature vector for each input/output pixel of index $i$ in this example would be, therefore, $a^{*}_{i} \in \mathbb{R}^{(28)}$.

Concatenating activations from multiple layers into $a^{*}$ yields high dimensionality feature vectors for each pixel, as modern CNNs/FCNs easily output hundreds or thousands of activation channels from each layer. The large redundancy found in activation maps from convolutional layers~\cite{Srivastava:2015,Huang:2017} should also render $a^{*}$ to be highly redundant. OpenPCS mitigates both problems by computing a lower dimension manifold $a^{low}$ of each pixel's activation with Principal Component Analysis (PCA) previously to fitting a generative model ($\mathcal{G}$), as shown in Figure~\ref{fig:openpcs_generative}. This approach grants two desirable properties to OpenPCS: 1) faster inference time during testing, as PCA implementations can be highly parallelized via vectorial operations and low-dimensional gaussian likelihood scoring can be computed in a fast manner; and 2) PCA feature selection guarantees that only the most important activation channels are used to compute a scoring function to detect OOD samples and, consequently, UUCs.

\subsubsection{Open Set Scoring with Principal Components}
\label{sec:openpcs_scoring}

Besides being purely a tensorial operation used for dimensionality reduction, PCA can be seen as a probability density estimator with gaussian priors. As described by Tipping and Bishop~\cite{Tipping:1999}, this allows PCA to be used as a generative model for novelty detection. In other words, the low dimensional Principal Components generated by PCA using a multivariate gaussian prior can yield likelihoods that allow for OOD recognition in new data.

PCA reduces dimensionality by finding latent variables composed of combinations of features in the original input space such that the reconstruction error when returning to this original space is minimized. This operation works by computing eigenvalues ($\lambda$) and eigenvectors ($v$) of a covariance matrix ($A$). Those values can be calculated using the equation $Av = \lambda v$. In the PCA procedure, the eigenvalues represent how impacting their correspondent eigenvector direction is to the data variability. Since reconstructions provided by PCA emphasize variation, in order to perform dimensionality reduction, the standard procedure is to project your data into an orthogonal basis composed by the eigenvectors that have the biggest correspondent eigenvalues, i.o.w., a subspace with the highest variance possible.

Exemplifying the application of PCA in OpenPCS, first we compute a covariance matrix using feature vector $a^{*}$ -- that is, a concatenation of activations from different layers in a DNN for a specific set of pixels. After that, we calculate the eigenvalues and eigenvectors of this matrix and select the set of $low$ eigenvectors ($v^{low}$) that have the correspondent $low$ largest eigenvalues ($\lambda^{low}$). Finally, we use those eigenvectors as a basis and project our input vectors ($a^{*}$) on it, resulting in $a^{low} = a^{*} \cdot v^{low}$.


\begin{figure}[!ht]
    \centering
    \includegraphics[width=0.75\textwidth]{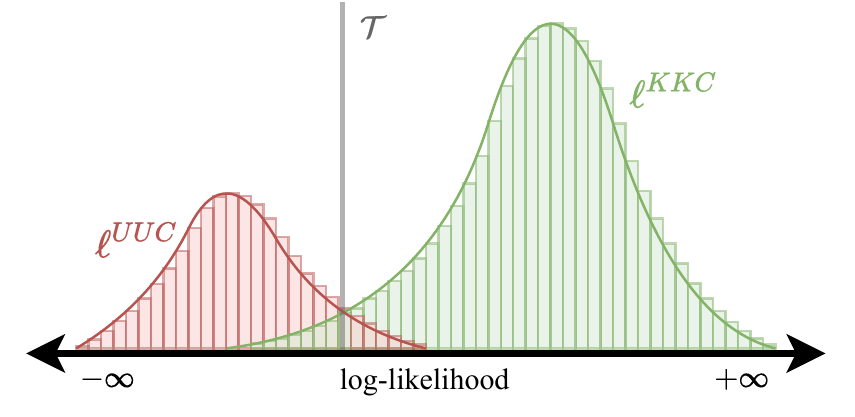}
    \caption{Depiction of log-likelihood probability density functions. This toy example's bimodal distribution is composed of two distinct peaks from the log-likelihoods of KKCs ($\ell^{KKC}$), which tend to have higher values, as the low dimensional gaussians were fit to them; and log-likelihoods from UUC samples ($\ell^{UUC}$), which assume lower values, as they were not fed to the gaussian density estimator.}
    \label{fig:openpcs_thresholding}
\end{figure}

\subsubsection{OpenIPCS}
\label{sec:openipcs}

OpenPCS is highly memory intensive, as fitting $n^{KKC}$ PCA models using millions of pixels with feature vectors in the scale of hundreds or thousands of bins requires all this data to be stored in RAM. In practice, for training the traditional OpenPCS we used a subsample of randomly selected patches to fit the models using a reasonable amount of memory. Empirically, we found that 150 patches with $224 \times 224$ pixels each resulted in an acceptable trade-off between memory and enough training data for the computation of Principal Components for each class. Even with this approach, OpenPCS required between 20 and 30 GB of memory, depending on the dimensionality of the feature vectors $a^{*}$. In addition, PCA models trained using subrepresented classes in the imbalanced datasets (i.e. Cars) did not fit correctly due to a low number of correctly classified samples used for training the generative model. Ultimately, OpenPCS for large and highly imbalanced datasets can result in underperformance on the task of UUC identification due to the subsampling required by the large memory usage of the method.

In an effort to minimize this problem, in addition to the traditional OpenPCS we also evaluate an Incremental PCA (IPCA) for generative modeling and likelihood scoring. The full OSR segmentation methodology will be henceforth referred to as OpenIPCA. In contrast to the traditional offline training of the standard PCA, IPCA allows for mini-batch online training, which is highly useful for correctly computing the Principal Components in large datasets. OpenIPCA also allows for the model to be further updated in an efficient manner whenever new data might arise, not requiring a full retraining of the standard PCAs from scratch, but instead updating the IPCA model by feeding newly acquired patches. We emphasize that, apart from the incremental/online training that allows all the training dataset to be used in the computation of the Principal Components, all other aspects of the implementation of OpenIPCS were identical to the standard OpenPCS.
\section{Experimental Setup}
\label{sec:experiments}

This section describes the experimental setup for the evaluation of OpenFCN against open and Closed Set baselines. In order to encourage reproducibility, we provide details regarding the datasets and evaluation protocol (Section~\ref{sec:datasets}), fully convolutional architectures and baselines (Section~\ref{sec:dnns}). In addition, we are publicizing the code for OpenFCN, OpenPCS and OpenIPCS in this project's webpage\footnotemark\footnotetext{\url{http://patreo.dcc.ufmg.br/2020/03/20/openfcn/}} in a conscious effort to encourage reproducibility of our results and follow-up research on OSR segmentation. OpenFCN's Weibull fitting and distance computation was based on libMR\footnotemark\footnotetext{\url{https://github.com/Vastlab/libMR}} and OpenPCS' Principal Components and likelihood scoring were computed using the scikit-learn\footnotemark\footnotetext{\url{https://scikit-learn.org/stable/}} library.

\subsection{Datasets and Evaluation Protocol}
\label{sec:datasets}

In order to validate the effectiveness of OpenFCN and OpenPCS on RS image segmentation, we used two urban scene 2D semantic labeling datasets from the International Society for Photogrammetry and Remote Sensing (ISPRS) with pixel-level labeling: Vaihingen\footnotemark\footnotetext{\url{http://www2.isprs.org/commissions/comm3/wg4/2d-sem-label-vaihingen.html}} and Potsdam\footnotemark\footnotetext{\url{http://www2.isprs.org/commissions/comm3/wg4/2d-sem-label-potsdam.html}}. Vaihingen presents a spatial resolution of 5cm/pixel with patches ranging from 2,000$\sim$2,500 pixels in each axis and Potsdam has 9cm/pixel samples with $6,000\times6,000$ patches. Both datasets contain IR-R-G-B spectral channels paired with semantic maps divided into 6 KKCs: impervious surfaces, buildings, low vegetation, high vegetation, cars and miscellaneous; and 1 KUC: segmentation boundaries between objects.

The data also allows for 3D information to be incorporated into the models via Digital Surface Model (DSM) images. However, in order to follow standard procedures in the RS literature~\cite{Sherrah:2016,Audebert:2016,Silva:2020}, we ignored the DSM and blue channel in our evaluation, limiting the experiments to the IR-R-G channels. Aiming to ease the computation of our experiments, we also ignored the miscellaneous class, as it contains a rather small number of samples in Vaihingen. For both datasets we used the standard procedure in the literature of training with most patches, while separating some specific patches for testing: 11, 15, 28, 30 and 34 for Vaihingen; and 2\_11, 2\_12, 4\_10, 5\_11, 6\_7, 7\_8 and 7\_10 in the case of Potsdam.

\begin{figure}[!ht]
    \centering
    \includegraphics[width=\textwidth]{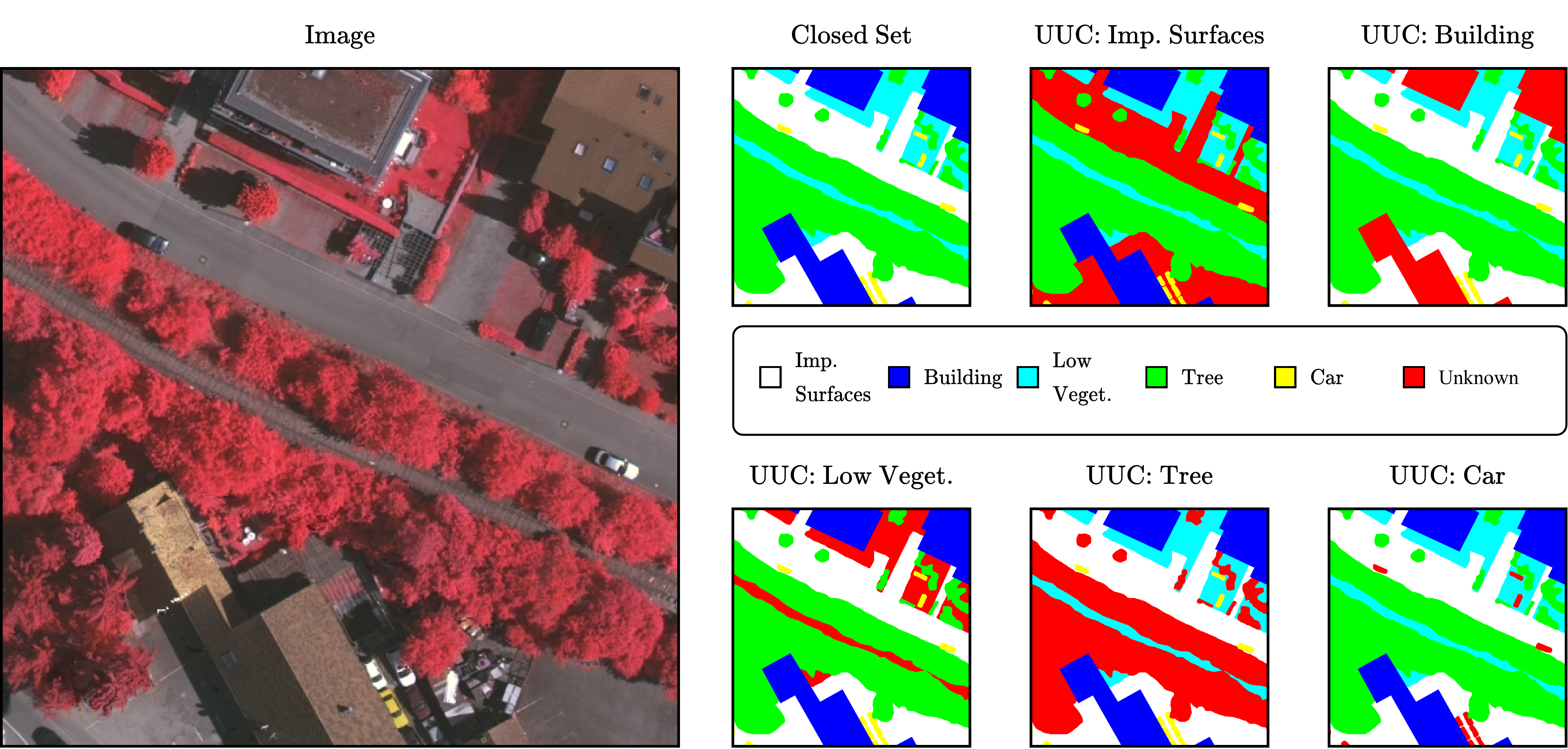}
    \caption{LOCO procedure for Open Set evaluation. On the left one can see a mini-patch chosen from one of the larger patches that compose sample 11 of Vaihingen. On the right we present the Closed Set labels and different KKCs marked as Unknown using the LOCO protocol.}
    \label{fig:loco}
\end{figure}

In order to simulate the Open Set environments on the Closed Set RS datasets, we employed a Leave-One-Class-Out (LOCO) protocol. LOCO works by selecting one class as UUC and ignoring its samples during training as shown in Figure~\ref{fig:loco}. This protocol allows for verifying both the overall performances of the segmentation architectures and class-by-class metrics, as there is a large class imbalance mainly when comparing classes as impervious surfaces and cars in the datasets.

One should notice that all samples in the data used in our experiments pertains to Closed Set Semantic Segmentation datasets, which allows for an objective evaluation of the effect of OSR segmentation on well-known classification metrics. Aiming to capture all nuances of training and testing Open Set scenarios, we propose a new standard set of test metrics that accounts for KKCs and UUCs, while also capturing the overall performance over all known and unknown classes. In order to evaluate the impact of adding the meta-recognition step to the inference procedure of KKCs, we computed the accuracy for known classes ($Acc^{K}$). The main metric adopted by Silva \textit{et al.}~\cite{Silva:2020} for evaluating the performance of OSR segmentation on UUCs is the binary accuracy between KKCs and UUCs ($Acc^{U}$), as in the case of Anomaly Detection. However, $Acc^{U}$ is artificially inflated due to the large imbalance between the number of pixels from KKCs and UUC in most cases. We solved this by adding the precision of unknown detection ($Pre^{U}$) to the evaluation in order to observe the trade-off in performance in the binary meta-recognition task. The Kappa ($\kappa$) metric was used to evaluate the overall performance in both KKCs and UUC of the algorithms, as it is a common metric for the evaluation of semantic segmentation in remote sensing. All of the metrics above are, however, threshold dependent, making it hard to compare methods with distinct threshold ranges. As the thresholds do not represent equal statistical entities in \softmaxt{}, OpenFCN and OpenPCS, we define the thresholds for the methods based on preset values of True Positive Rate (TPR). This evaluation protocol also has the benefit of yielding information about the recall of the methods, as TPR and recall are equivalent in this case. Also, in order to evaluate the methods in a threshold-independent fashion, we complement our analysis with the Receiver Operating Characteristic (ROC) curve -- which evaluates the whole range of thresholds for the different methods on a TPR vs. False Positive Rate (FPR) 2D plot -- and its corresponding Area Under Curve (AUC).

Due to the large spatial resolution of Vaihingen and Potsdam patches, we sliced them into $224\times224$ mini-patches for training, in order to fit ImageNet's~\cite{Deng:2009} traditional patch size, which is the most commonly expected input size for the CNNs used as backbones in our experiments. During training, we used data augmentation based on random cropping for patch selection, random flips in the horizontal and vertical dimensions and random rotations by multiples of 90 degrees. In order to solve inconsistencies due to lack of context and to compensate for patch border uncertainties, we employed overlapping patches of $224\times224$ pixels during testing. We explicit that this procedure considerably increased the computational budget of our experiments mainly in the Potsdam dataset for OpenFCN due to inefficiencies and non-vectorized operations in libMR. A more thorough evaluation of time complexity can be found in Section~\ref{sec:time_comparison}. 

\subsection{Fully Convolutional Architectures and Baselines}
\label{sec:dnns}

We compared several distinct fully convolutional architectures for OSR semantic segmentation, including five distinct CNN backbones: VGG~\cite{Simonyan:2014}, ResNet-50~\cite{He:2016}, Wide ResNet-50-2~\cite{Zagoruyko:2016} (WRN), ResNeXt-50-32x4d~\cite{Xie:2017} (ResNeXt) and DenseNet-121~\cite{Huang:2017}. 

Based on the methodology of OpenMax~\cite{Bendale:2016}, we also compare OpenFCN and OpenPCS with traditional SoftMax followed by thresholding. SoftMax Thresholding (\softmaxt{}) follows the premise that least certain network predictions may be motivated by outlier classes seen in test time. In addition to these Open Set approaches, we evaluate the proposed methods in comparison with traditional Closed Set fully convolutional architectures for dense labeling using the LOCO protocol. Closed Set FCNs wrongly classify UUCs by design, forcing the pixel to be segmented according to the higher probability prediction among KKCs. One should notice that OpenPCS, OpenFCN, \softmaxt{} and Closed Set FCNs were evaluated according to the same pretrained DNNs, differing only on test time. This protocol allows for direct comparisons in objective metrics disregarding performance variability due to the random nature of gradient descent optimization.
\section{Results and Discussion}
\label{sec:results}

In this section we present and discuss the obtained results.
First, an analysis is performed in Section~\ref{sec:auc} in order to define the most suitable architectures for the proposed approaches. This section contains both overall and per-class threshold-independent analysis. Section~\ref{sec:baseline} compares the proposed methods with the baseline in terms of quantitative metrics (Sections~\ref{sec:quantitative_vaihingen} and~\ref{sec:quantitative_potsdam}), qualitative segmentation maps (Section~\ref{sec:qualitative}), and runtime performance (Section~\ref{sec:time_comparison}).


Note that, for the sake of simplicity and clarity, only the most relevant results were reported in this Section. For a full report of the results, please, check the supplementary material on this project's webpage linked in Section~\ref{sec:experiments}.

\subsection{Architecture Analysis}
\label{sec:auc}

In this Section, we analyze the networks in order to define the most suitable architectures for the proposed techniques. To perform this evaluation, we employed the AUC, a threshold-independent performance measurement that allows comparisons between methods without resorting to (potentially) arbitrary thresholds. Furthermore, it is important to emphasize that we performed this analysis by using only the Vaihingen dataset. This is due to the fact that the Potsdam dataset is very similar to Vaihingen one and, therefore, analysis and decisions made over the latter dataset are also applicable to the former one.



Table~\ref{tab:results_vaihingen_auc} presents the average AUC results over all runs of the LOCO protocol. For \softmaxt{} and OpenFCN, the ResNet-50 architecture~\cite{He:2016} produced the best outcomes, followed closely by WRN-50~\cite{Zagoruyko:2016} and DenseNet-121~\cite{Huang:2017}. OpenPCS showed better results with WRN-50~\cite{Zagoruyko:2016}, followed closely by DenseNet-121~\cite{Huang:2017} and VGG~\cite{Simonyan:2014}. Since WRN~\cite{Zagoruyko:2016} and DenseNet~\cite{Huang:2017} were the most stable networks, producing good results in all approaches, they were selected and used in further experiments shown in the following sections.

\begin{table}[!ht]
    \centering
    \caption{Average AUC values in the LOCO protocol for each evaluated FCN backbone in the \textbf{Vaihingen} dataset. Bold values indicate the best overall results including all methods.}
    \label{tab:results_vaihingen_auc}
    \begin{tabular}{|c|c|c|c|}
        \hline
        \textbf{Backbone}     & \textbf{\softmaxt{}} & \textbf{OpenFCN}           & \textbf{OpenPCS}           \\ \hline
        \textbf{WRN-50}~\cite{Zagoruyko:2016}       & 0.68 $\pm$ 0.09                & 0.69 $\pm$ 0.10          & \textbf{0.82 $\pm$ 0.12} \\ \hline
        \textbf{DenseNet-121}~\cite{Huang:2017} & 0.68 $\pm$ 0.08                & 0.68 $\pm$ 0.09          & 0.79 $\pm$ 0.12          \\ \hline
        \textbf{ResNeXt-50}~\cite{Xie:2017}   & 0.64 $\pm$ 0.06                & 0.65 $\pm$ 0.04          & 0.72 $\pm$ 0.12          \\ \hline
        \textbf{ResNet-50}~\cite{He:2016}    & \textbf{0.69 $\pm$ 0.12}       & \textbf{0.71 $\pm$ 0.12} & 0.69 $\pm$ 0.16          \\ \hline
        \textbf{VGG-19}~\cite{Simonyan:2014}       & 0.67 $\pm$ 0.06                & 0.68 $\pm$ 0.07          & 0.77 $\pm$ 0.12          \\ \hline
    \end{tabular}
\end{table}

Based on the best results for Vaihingen, we also report the threshold independent average AUC analysis over all UUCs on Potsdam. Table~\ref{tab:results_potsdam_auc} shows the results for DenseNet-121 and WRN-50 on the considerably larger Potsdam dataset, wherein extensive architectural comparisons were not feasible.

\begin{table}[!ht]
    \centering
    \caption{Average AUC values in the LOCO protocol for \textbf{DenseNet-121} and \textbf{WRN-50} backbones in the \textbf{Potsdam} dataset. Bold values indicate the best overall results including all methods.}
    \label{tab:results_potsdam_auc}
    \begin{tabular}{|c|c|c|c|c|}
        \hline
        \textbf{Backbone} & \textbf{SoftMax$^{\mathcal{T}}$} & \textbf{OpenFCN} & \textbf{OpenPCS}         & \textbf{OpenIPCS}        \\ \hline
        \textbf{WRN}      & 0.62 $\pm$ 0.06                  & 0.62 $\pm$ 0.07  & \textbf{0.73 $\pm$ 0.13} & \textbf{0.73 $\pm$ 0.13} \\ \hline
        \textbf{DenseNet} & 0.61 $\pm$ 0.11                  & 0.62 $\pm$ 0.12  & 0.66 $\pm$ 0.25          & 0.71 $\pm$ 0.23          \\ \hline
    \end{tabular}
\end{table}

One can easily see that in both tables OpenPCS and OpenIPCS present better AUC results than OpenFCN and \softmaxt{}, while the distinction between OpenFCN and \softmaxt{} is often rather small or nonexistent. These evaluations are averages and standard deviations computed across all classes in the LOCO protocol, therefore, apart from a raw evaluation of the overall performance of each method, the previously mentioned.

\subsection{Baseline Comparison}
\label{sec:baseline}

Based on the analysis performed in previous sections, we have conducted several experiments to investigate the effectiveness of the proposed methods using the most promising architectures: WRN-50 and DenseNet-121. We further investigate the performance on both KKCs and UUCs of the proposed methods (OpenFCN and OpenPCS) and baseline (\softmaxt{}) using the threshold-dependent metrics described in Section~\ref{sec:datasets}. This section also shows qualitative segmentation predictions from our experiments.

\subsubsection{Per-Class Analysis}
\label{sec:per_class}

Table~\ref{tab:results_vaihingen_auc_per_class} presents the results, in terms of AUC per UUC, for the Vaihingen dataset. As can be seen, the OpenPCS obtained most of the best results when evaluating the UUCs and networks independently (bold values). When considering all the networks together and only the UUCs independently (values with $\dagger$), OpenPCS produced the best outcomes for four UUCs while OpenFCN yielded the best result for one UUC.


\begin{table}[!ht]
    \centering
    \caption{AUC values for each UUC of the \textbf{Vaihingen} dataset. Bold values indicate the best results when comparing \softmaxt{}, OpenFCN and OpenPCS for each network and UUC. Results followed by $\dagger$ represent the best overall results for a certain UUC.}
    \label{tab:results_vaihingen_auc_per_class}
    \resizebox{\textwidth}{!}{%
    \begin{tabular}{|c|c|c|c|c|c|c|}
        \hline
        \textbf{Backbone}                  & \textbf{Methods}                 & \textbf{\begin{tabular}[c]{@{}c@{}}UUC:\\ Imp. Surf.\end{tabular}} & \textbf{\begin{tabular}[c]{@{}c@{}}UUC:\\ Building\end{tabular}} & \textbf{\begin{tabular}[c]{@{}c@{}}UUC:\\ Low Veg.\end{tabular}} & \textbf{\begin{tabular}[c]{@{}c@{}}UUC:\\ Tree\end{tabular}} & \textbf{\begin{tabular}[c]{@{}c@{}}UUC:\\ Car\end{tabular}} \\ \hline
        \multirow{3}{*}{\textbf{DenseNet}} & \textbf{\softmaxt{}} & .78 $\pm$ .03                                                      & .63 $\pm$ .05                                                    & \textbf{.73 $\pm$ .05}                                           & .67 $\pm$ .06                                                & .58 $\pm$ .06                                               \\ \cline{2-7} 
                                          & \textbf{OpenFCN}                 & .81 $\pm$ .03                                                      & .64 $\pm$ .05                                                    & .72 $\pm$ .05                                                    & .66 $\pm$ .06                                                & .58 $\pm$ .05                                               \\ \cline{2-7} 
                                          & \textbf{OpenPCS}                 & \textbf{.84 $\pm$ .03}                                             & \textbf{.94 $\pm$ .01}\rlap{$\dagger$}                                  & .70 $\pm$ .08                                                    & \textbf{.81 $\pm$ .07}\rlap{$\dagger$}                             & \textbf{.65 $\pm$ .06}                                      \\ \hline
        \multirow{3}{*}{\textbf{WRN}}      & \textbf{\softmaxt{}} & .80 $\pm$ .03                                                      & .64 $\pm$ .04                                                    & .75 $\pm$ .05                                                    & \textbf{.63 $\pm$ .06}                                       & .58 $\pm$ .05                                               \\ \cline{2-7} 
                                          & \textbf{OpenFCN}                 & .83 $\pm$ .03                                                      & .67 $\pm$ .03                                                    & .74 $\pm$ .05                                                    & .62 $\pm$ .06                                                & .58 $\pm$ .05                                               \\ \cline{2-7} 
                                          & \textbf{OpenPCS}                 & \textbf{.87 $\pm$ .03}                                             & \textbf{.94 $\pm$ .02}\rlap{$\dagger$}                                   & \textbf{.77 $\pm$ .04}                                           & .63 $\pm$ .09                                                & \textbf{.87 $\pm$ .03}\rlap{$\dagger$}                              \\ \hline
        \multirow{3}{*}{\textbf{ResNeXt}}  & \textbf{\softmaxt{}} & .66 $\pm$ .06                                                      & .55 $\pm$ .07                                                    & \textbf{.71 $\pm$ .07}                                           & .66 $\pm$ .04                                                & .65 $\pm$ .04                                               \\ \cline{2-7} 
                                          & \textbf{OpenFCN}                 & .67 $\pm$ .06                                                      & .59 $\pm$ .05                                                    & .71 $\pm$ .07                                                    & .64 $\pm$ .03                                                & .63 $\pm$ .04                                               \\ \cline{2-7} 
                                          & \textbf{OpenPCS}                 & \textbf{.88 $\pm$ .03}\rlap{$\dagger$}                                     & \textbf{.81 $\pm$ .07}                                           & .58 $\pm$ .08                                                    & \textbf{.69 $\pm$ .09}                                       & \textbf{.65 $\pm$ .05}                                      \\ \hline
        \multirow{3}{*}{\textbf{ResNet}}   & \textbf{\softmaxt{}} & .83 $\pm$ .03                                                      & .62 $\pm$ .06                                                    & \textbf{.81 $\pm$ .05}\rlap{$\dagger$}                                   & .60 $\pm$ .05                                                & .60 $\pm$ .05                                               \\ \cline{2-7} 
                                          & \textbf{OpenFCN}                 & \textbf{.86 $\pm$ .02}                                             & .64 $\pm$ .05                                                    & .80 $\pm$ .05                                                    & \textbf{.61 $\pm$ .05}                                       & .62 $\pm$ .05                                               \\ \cline{2-7} 
                                          & \textbf{OpenPCS}                 & .75 $\pm$ .03                                                      & \textbf{.92 $\pm$ .02}                                           & .65 $\pm$ .07                                                    & .50 $\pm$ .07                                                & \textbf{.63 $\pm$ .05}                                      \\ \hline
        \multirow{3}{*}{\textbf{VGG}}      & \textbf{\softmaxt{}} & .72 $\pm$ .02                                                      & .59 $\pm$ .04                                                    & .74 $\pm$ .04                                                    & \textbf{.66 $\pm$ .05}                                       & .66 $\pm$ .05                                               \\ \cline{2-7} 
                                          & \textbf{OpenFCN}                 & .74 $\pm$ .03                                                      & .59 $\pm$ .04                                                    & \textbf{.74 $\pm$ .03}                                           & .65 $\pm$ .05                                                & .65 $\pm$ .04                                               \\ \cline{2-7} 
                                          & \textbf{OpenPCS}                 & \textbf{.82 $\pm$ .04}                                             & \textbf{.90 $\pm$ .02}                                           & .71 $\pm$ .06                                                    & .60 $\pm$ .07                                                & \textbf{.81 $\pm$ .04}                                      \\ \hline
    \end{tabular}
    }
\end{table}

\begin{table}[!ht]
    \centering
    \caption{AUC values for each UUC of the \textbf{Postdam} dataset. Bold values indicate the best results when comparing \softmaxt{}, OpenFCN, OpenPCS and OpenIPCS for each network and UUC. Results followed by $\dagger$ represent the best overall results for a certain UUC.}
    \label{tab:results_potsdam_auc_per_class}
    \resizebox{\textwidth}{!}{%
    \begin{tabular}{|c|c|c|c|c|c|c|}
        \hline
        \textbf{Backbone}                  & \textbf{Methods}                 & \textbf{\begin{tabular}[c]{@{}c@{}}UUC:\\ Imp. Surf.\end{tabular}} & \textbf{\begin{tabular}[c]{@{}c@{}}UUC:\\ Building\end{tabular}} & \textbf{\begin{tabular}[c]{@{}c@{}}UUC:\\ Low Veg.\end{tabular}} & \textbf{\begin{tabular}[c]{@{}c@{}}UUC:\\ Tree\end{tabular}} & \textbf{\begin{tabular}[c]{@{}c@{}}UUC:\\ Car\end{tabular}} \\ \hline
        \multirow{4}{*}{\textbf{DenseNet}} & \textbf{SoftMax$^{\mathcal{T}}$} & .66 $\pm$ .09                                                      & .49 $\pm$ .07                                                    & \textbf{.53 $\pm$ .13}                                           & \textbf{.64 $\pm$ .08$\dagger$}                              & .76 $\pm$ .02                                               \\ \cline{2-7} 
                                           & \textbf{OpenFCN}                 & .66 $\pm$ .09                                                      & .50 $\pm$ .06                                                    & .51 $\pm$ .13                                                    & .62 $\pm$ .07                                                & .80 $\pm$ .03                                               \\ \cline{2-7} 
                                           & \textbf{OpenPCS}                 & .83 $\pm$ .06                                                      & .87 $\pm$ .13                                                    & .28 $\pm$ .03                                                    & .54 $\pm$ .11                                                & .77 $\pm$ .04                                               \\ \cline{2-7} 
                                           & \textbf{OpenIPCS}                & \textbf{.84 $\pm$ .07$\dagger$}                                    & \textbf{.88 $\pm$ .14$\dagger$}                                  & .41 $\pm$ .07                                                    & .54 $\pm$ .09                                                & \textbf{.91 $\pm$ .04$\dagger$}                             \\ \hline
        \multirow{4}{*}{\textbf{WRN}}      & \textbf{SoftMax$^{\mathcal{T}}$} & .68 $\pm$ .06                                                      & .53 $\pm$ .06                                                    & .60 $\pm$ .12                                                    & \textbf{.62 $\pm$ .07}                                       & .66 $\pm$ .10                                               \\ \cline{2-7} 
                                           & \textbf{OpenFCN}                 & \textbf{.69 $\pm$ .05}                                             & .54 $\pm$ .05                                                    & .58 $\pm$ .12                                                    & .60 $\pm$ .07                                                & .69 $\pm$ .09                                               \\ \cline{2-7} 
                                           & \textbf{OpenPCS}                 & .66 $\pm$ .08                                                      & \textbf{.86 $\pm$ .10}                                           & .71 $\pm$ .08                                                    & .57 $\pm$ .10                                                & \textbf{.86 $\pm$ .04}                                      \\ \cline{2-7} 
                                           & \textbf{OpenIPCS}                & .64 $\pm$ .09                                                      & \textbf{.86 $\pm$ .10}                                           & \textbf{.76 $\pm$ .07$\dagger$}                                  & .55 $\pm$ .10                                                & .83 $\pm$ .04                                               \\ \hline
    \end{tabular}
    }
\end{table}

Analysis of Table~\ref{tab:results_vaihingen_auc_per_class} reveal that OpenPCS excels in almost all architectures and UUCs in the LOCO protocol. The overall best results for four out of five UUCs in the LOCO protocol (\textit{Impervious Surfaces}, \textit{Building}, \textit{Tree} and \textit{Car}) pertain to OpenPCS. Most AUCs for OpenPCS in these UUCs achieve values larger than 0.80, peaking at 0.94 in the class \textit{Building}. Not coincidentally, these classes are also the most semantically and visually distinct from the other ones. More detailed insights regarding the nature of the superiority of OpenPCS over the other methods will be discussed further in Section~\ref{sec:baseline}.

The outlier in this pattern is the class \textit{Low Vegetation}, as its samples present a much larger intraclass variability than the others, encompassing sidewalks, bushes, grass fields in the interior and exterior of buildings, gardens, backyards, some sections of parking lots and even train tracks partially covered with vegetation. Due to this higher variation in samples, the lower-dimensional multivariate gaussian fitting of OpenPCS was not able to properly map the whole variation in the data, as was the simple Weibull fitting of OpenFCN. Thus, \softmaxt{} overcame both proposed methods and achieved a peak AUC of 0.81 on ResNet-50. Even though the best result in the UUC \textit{Tree} was achieved by OpenPCS, the peak AUC for this class was also 0.81, this time using a DenseNet-121 backbone. The main visual distinction between \textit{Trees} and \textit{Low Vegetation} is not in the visible spectrum, though. While textures for most patches of both classes are rather similar when taking into account grass fields and bushes for \textit{Low Vegetation}, the DSM data is more visually distinct than the IRRG data, as the peak altitude for tree tops are above the mostly plain areas of \textit{Low Vegetation}. This implies that using the inputs channels (IRRG and DSM) coupled with the middle and later activations of the DNNs on the computation of PCAs for OpenPCS could have improved the performance of both classes. However, more research on this must be done to either confirm or deny this hypothesis.

\begin{figure*}[!ht]
	\centering
	\renewcommand{\currprop}{0.25\textwidth}
	\subfloat[Imp. Surf.]{
		\includegraphics[width=\currprop]{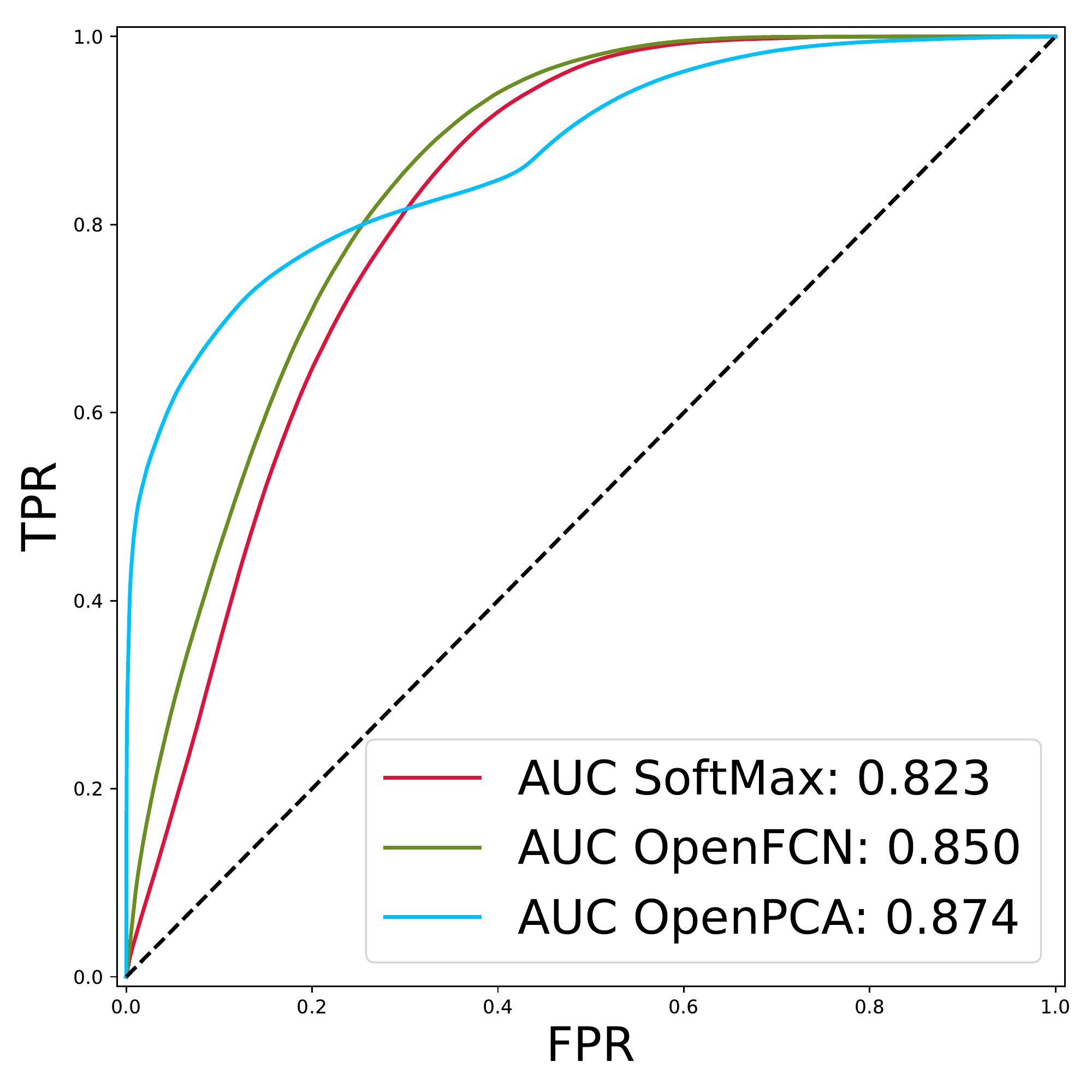}
		\label{fig:roc_fcndensenet121_vaihingen_0_area11}
	}
	\subfloat[Building]{
		\includegraphics[width=\currprop]{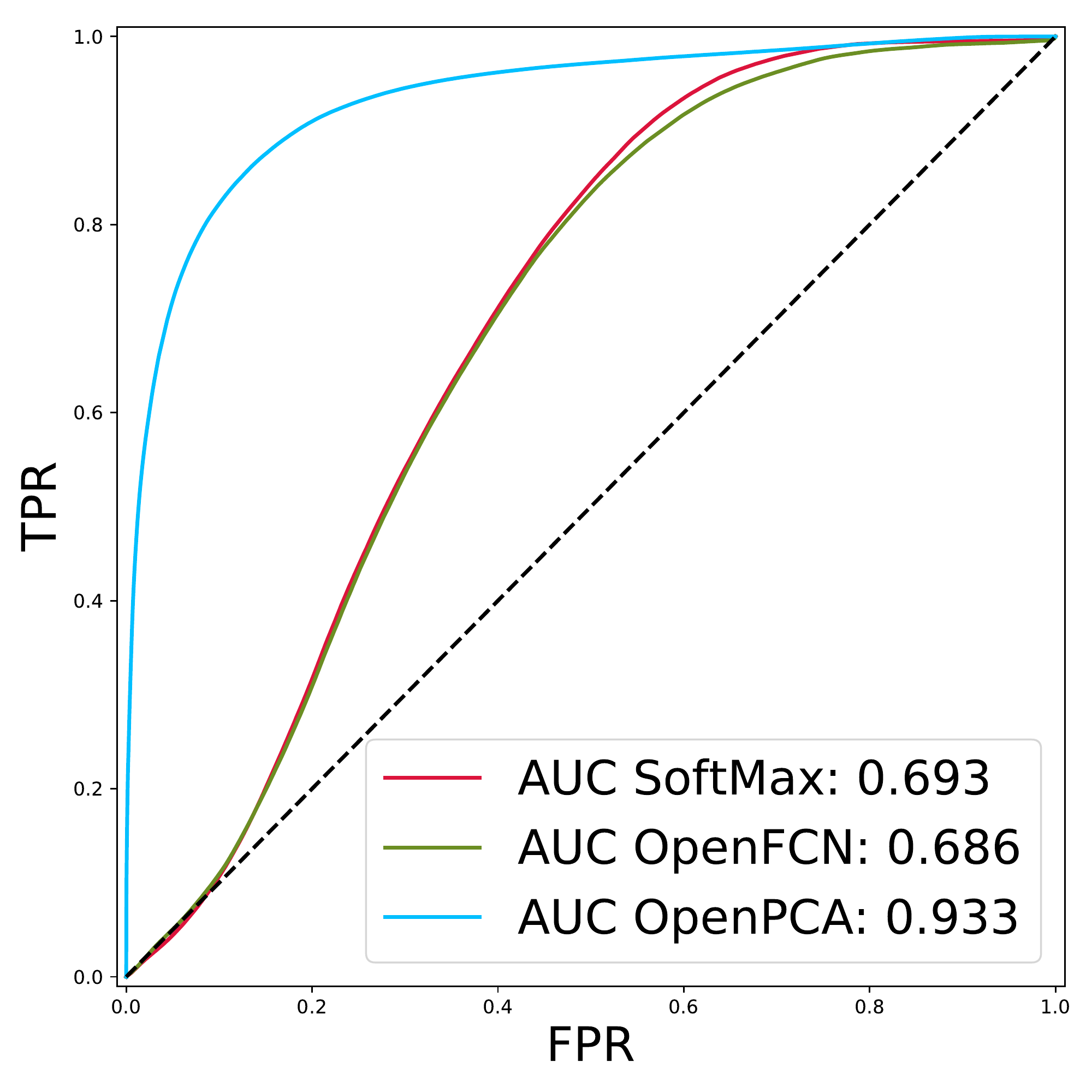}
		\label{fig:roc_fcndensenet121_vaihingen_1_area11}
	}
	\subfloat[Low Veg.]{
		\includegraphics[width=\currprop]{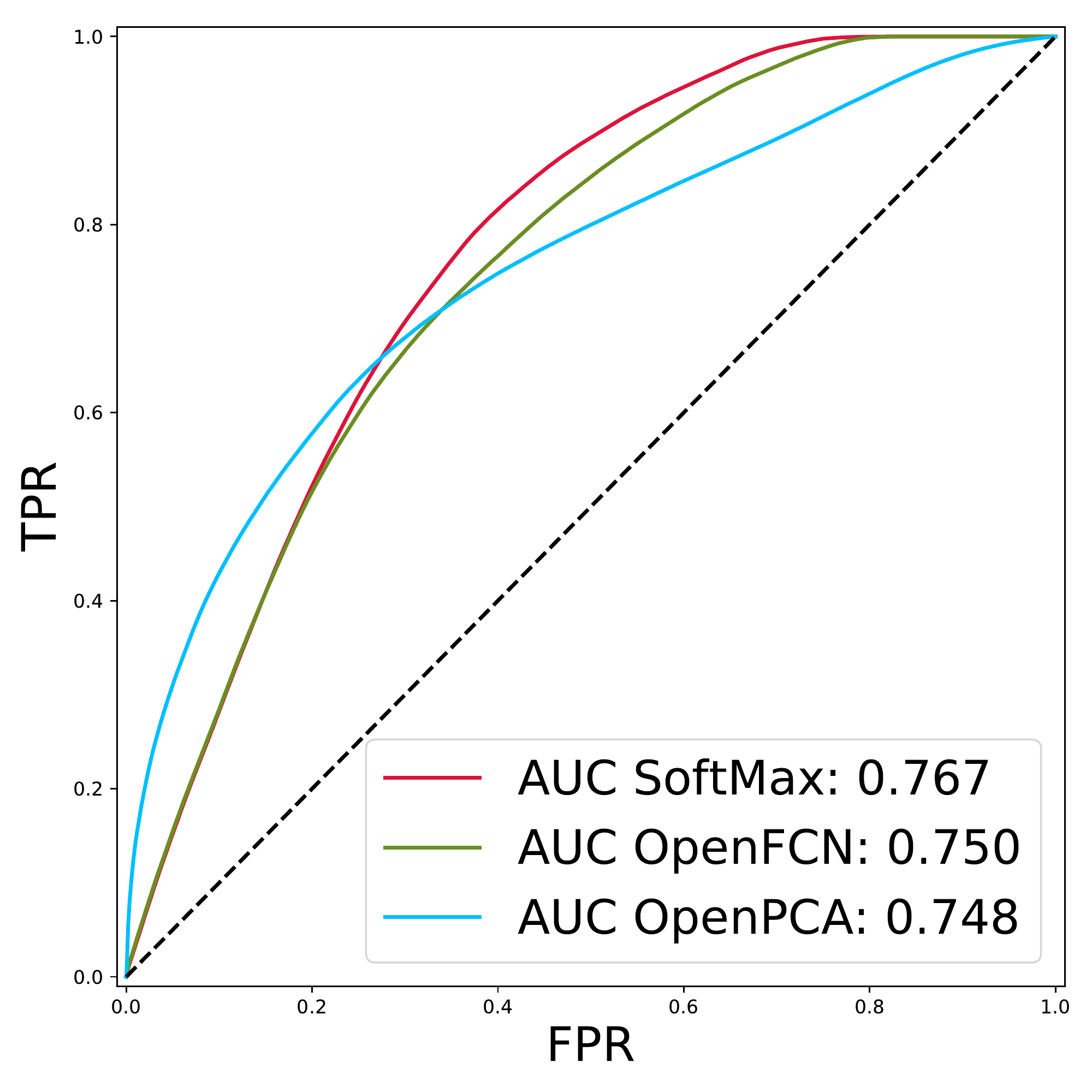}
		\label{fig:roc_fcndensenet121_vaihingen_2_area11}
	}
	\\
	\subfloat[Tree]{
		\includegraphics[width=\currprop]{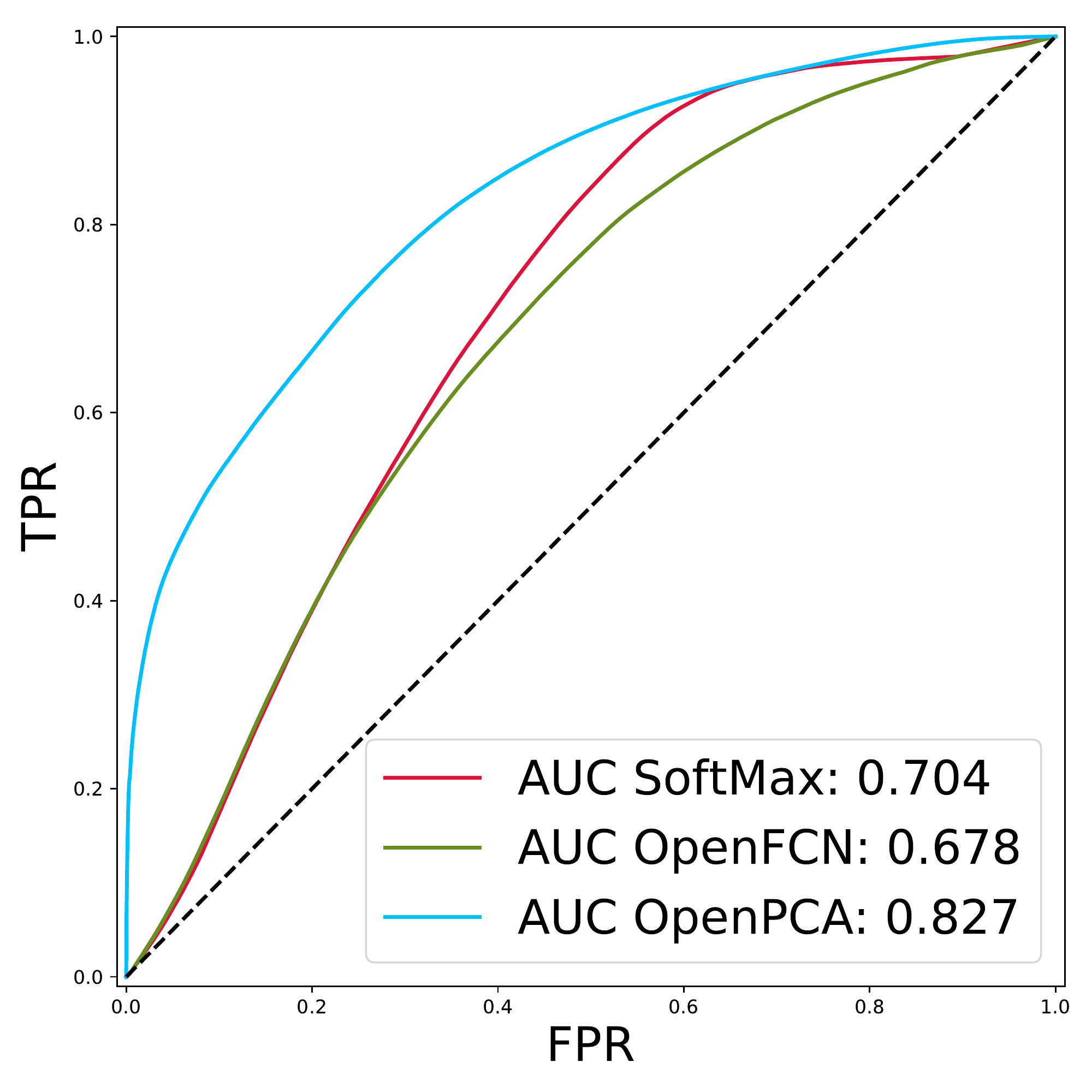}
		\label{fig:roc_fcndensenet121_vaihingen_3_area11}
	}
	\subfloat[Car]{
		\includegraphics[width=\currprop]{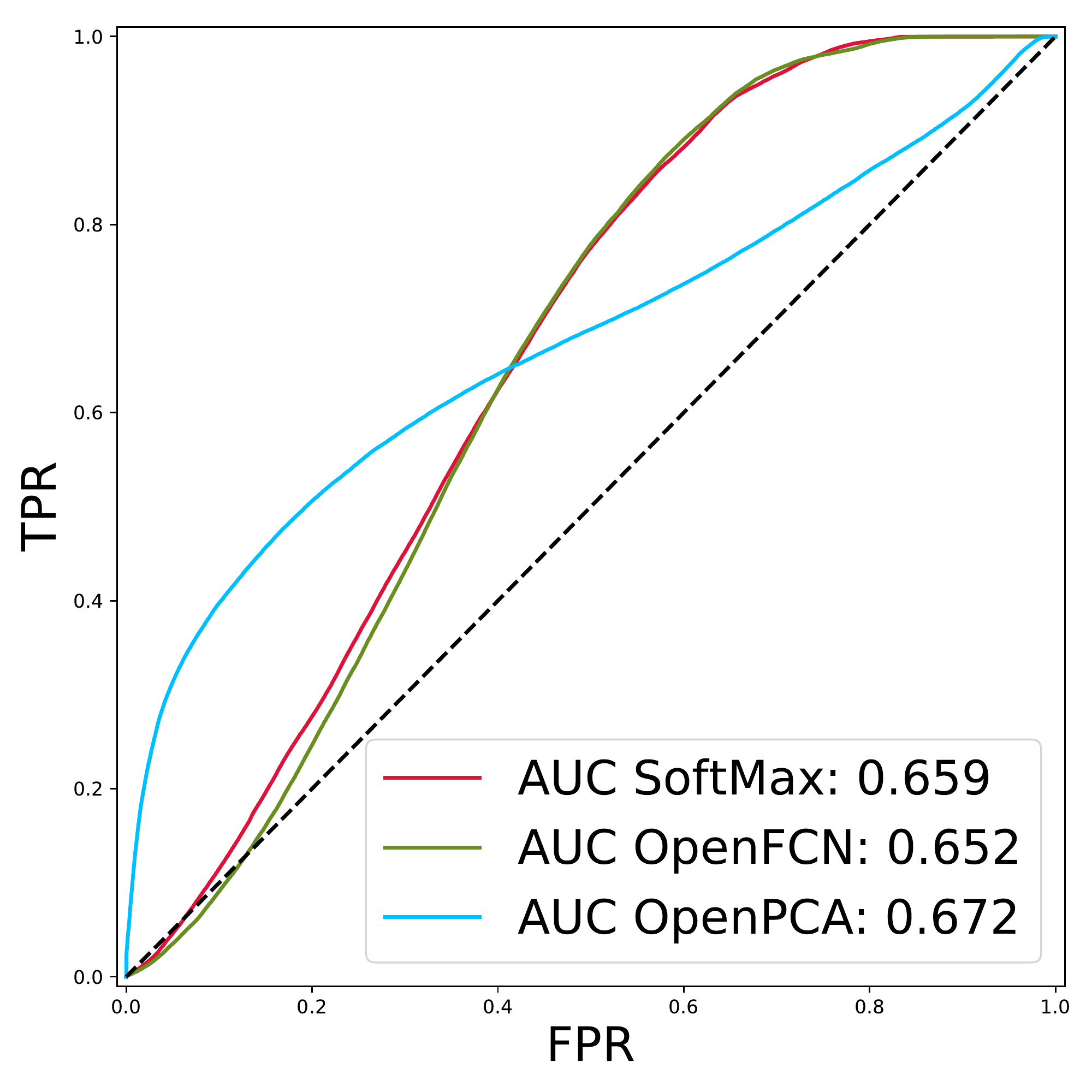}
		\label{fig:roc_fcndensenet121_vaihingen_4_area11}
	}
	\caption{ROC curves for \softmaxt{}, OpenFCN and OpenPCS on sample $11$ from \textbf{Vaihingen} using a \textbf{DenseNet-121} backbone.}
	\label{fig:roc_vaihingen}
\end{figure*}

\begin{figure*}[!ht]
	\centering
	\renewcommand{\currprop}{0.25\textwidth}
	\subfloat[Imp. Surf.]{
		\includegraphics[width=\currprop]{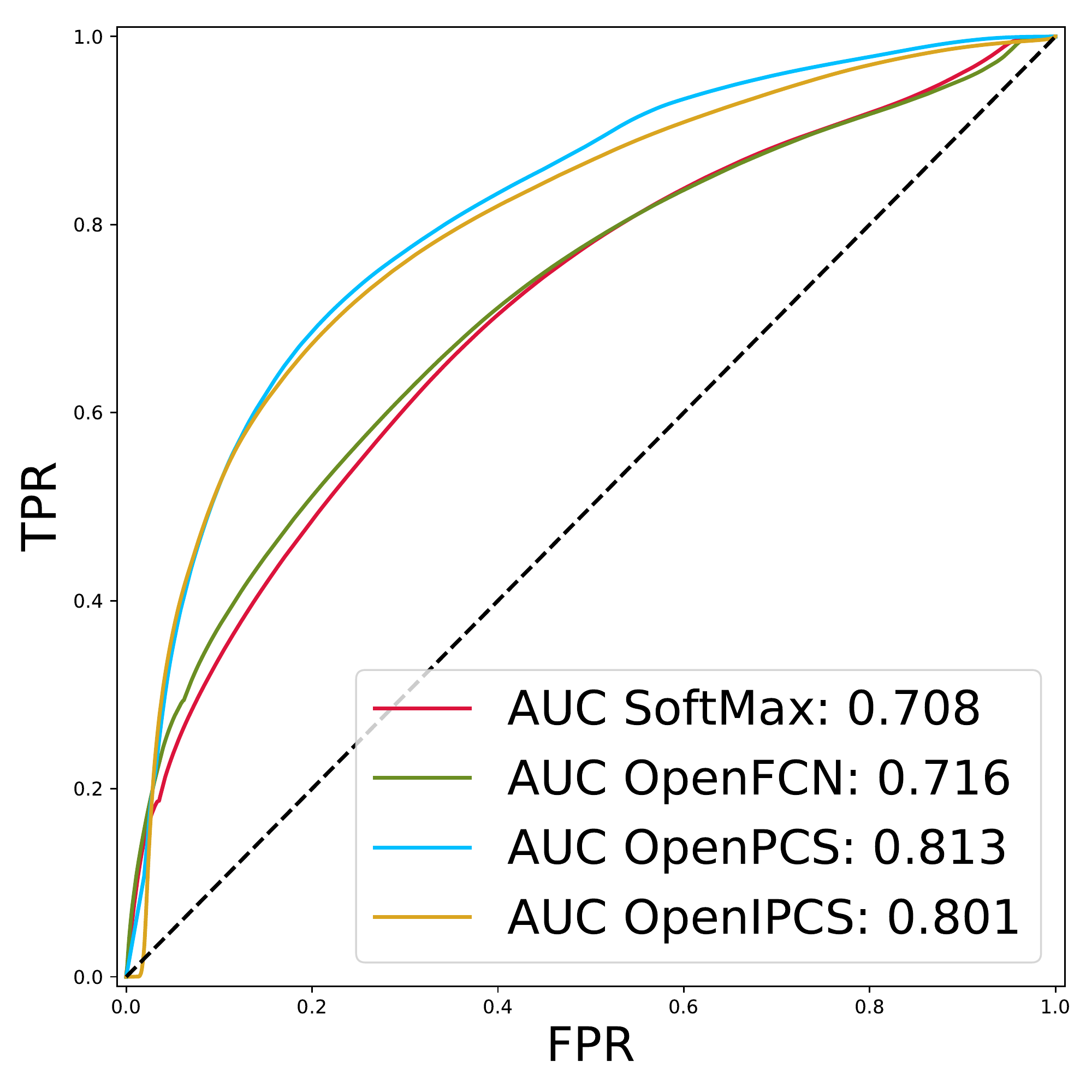}
		\label{fig:roc_fcnwideresnet50_potsdam_0_area_6_7}
	}
	\subfloat[Building]{
		\includegraphics[width=\currprop]{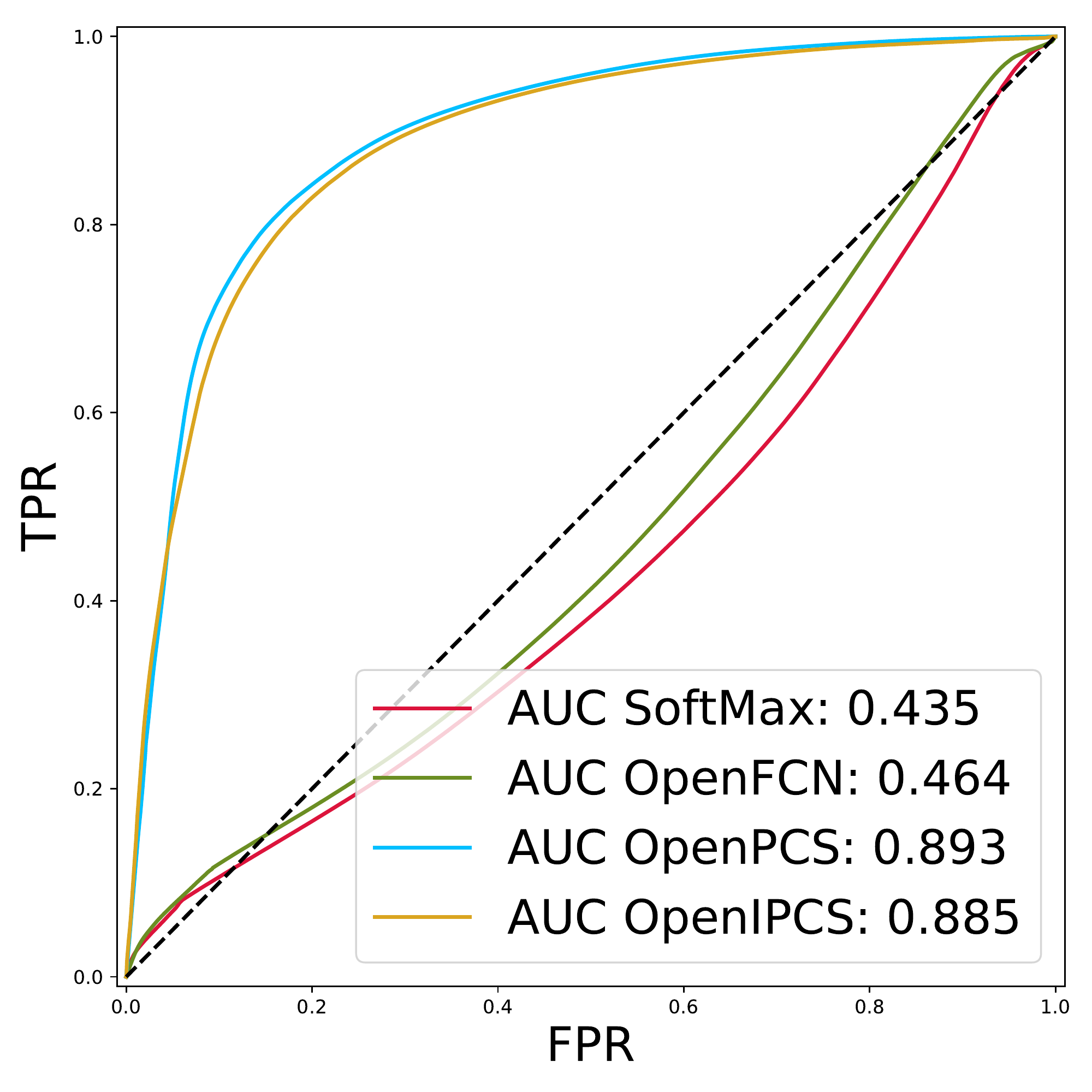}
		\label{fig:roc_fcnwideresnet50_potsdam_1_area_6_7}
	}
	\subfloat[Low Veg.]{
		\includegraphics[width=\currprop]{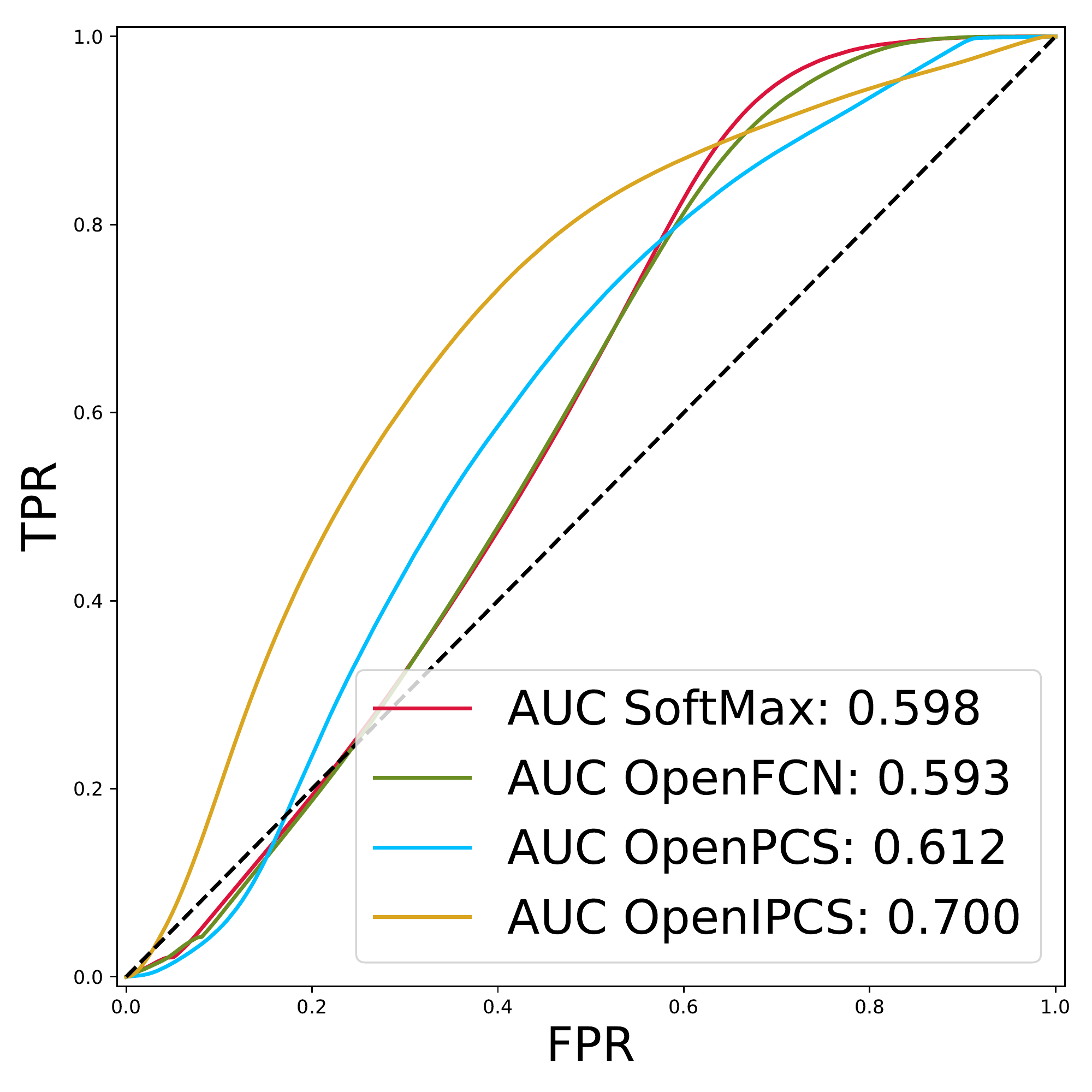}
		\label{fig:roc_fcnwideresnet50_potsdam_2_area_6_7}
	}
	\\
	\subfloat[Tree]{
		\includegraphics[width=\currprop]{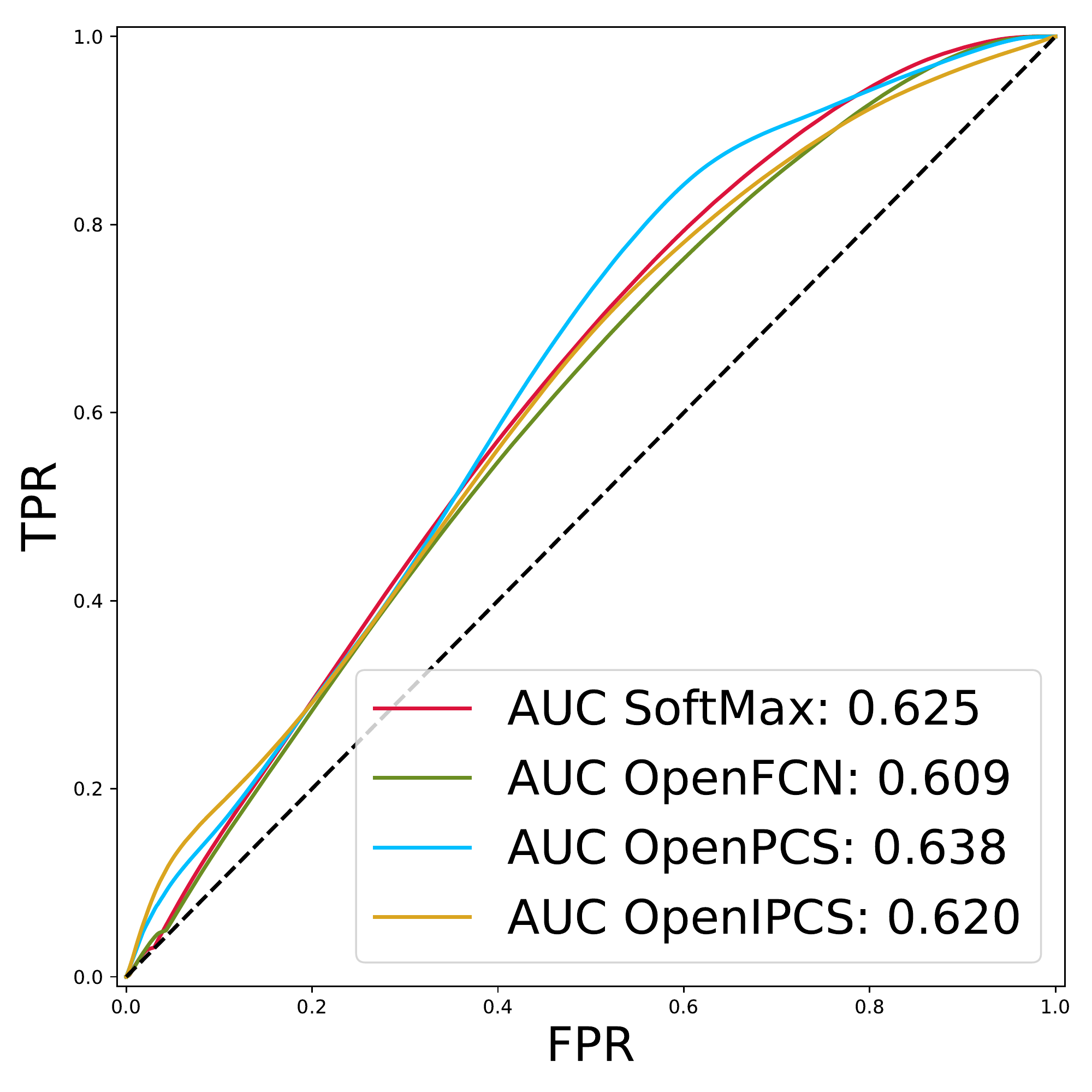}
		\label{fig:roc_fcnwideresnet50_potsdam_3_area_6_7}
	}
	\subfloat[Car]{
		\includegraphics[width=\currprop]{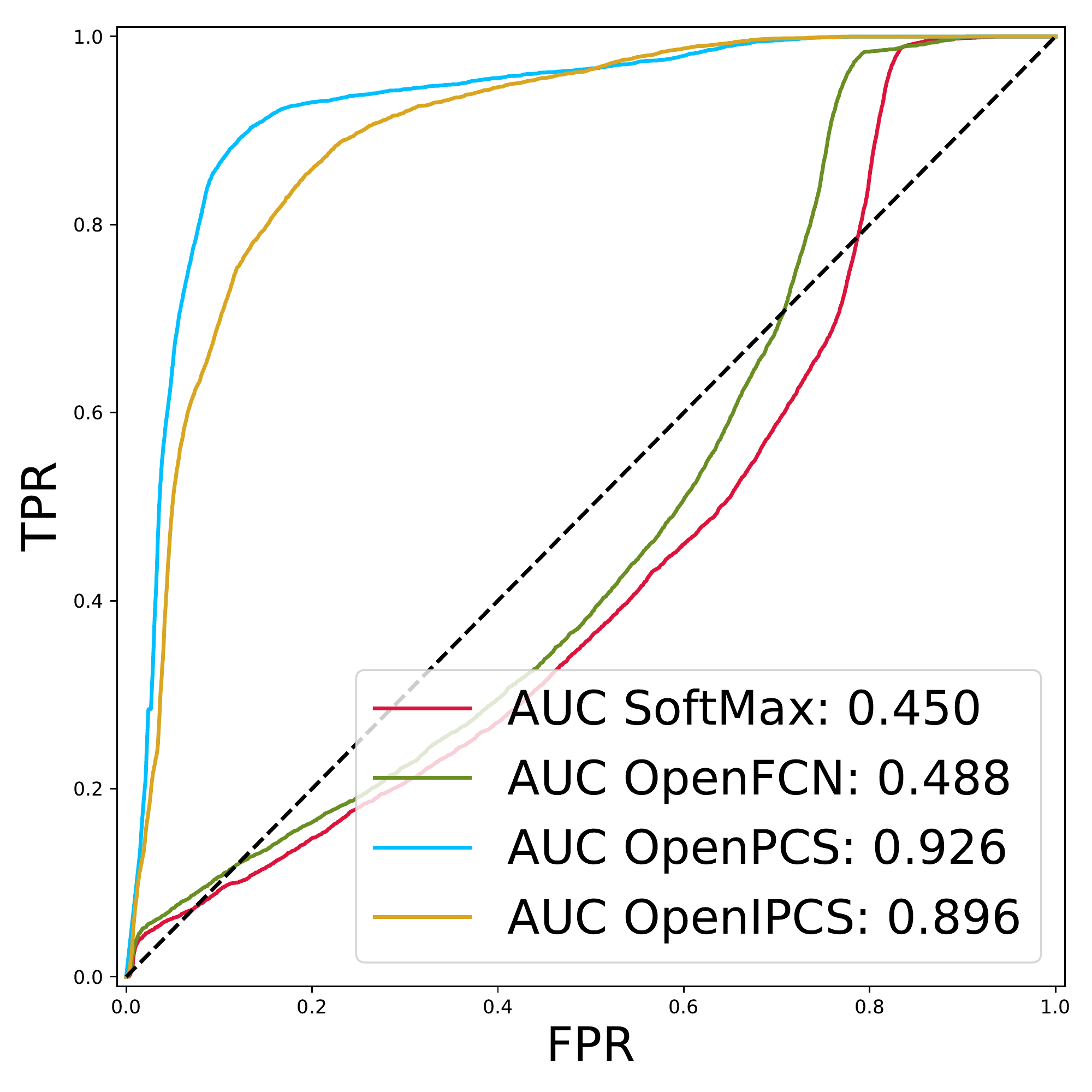}
		\label{fig:roc_fcnwideresnet50_potsdam_4_area_6_7}
	}
	\caption{ROC curves for \softmaxt{}, OpenFCN, OpenPCS and OpenIPCS on sample $6\_7$ from \textbf{Potsdam} using a \textbf{WRN-50} backbone.}
	\label{fig:roc_potsdam}
\end{figure*}

To better demonstrate the differences in performance between the techniques, Figures~\ref{fig:roc_vaihingen} and~\ref{fig:roc_potsdam} present the ROC curves for samples from the Vaihingen and Potsdam datasets, respectively. It is important to emphasize that each figure was created using a binary mask that separates KKCs from the evaluated UUC. Through Figures~\ref{fig:roc_vaihingen}, it is possible to observe that, in general, OpenPCS produces better results with lower FPRs for the same TPRs. It is also possible to visually assess that the detection capabilities of the methods for UUC \textit{Impervious Surfaces} were considerably superior to other classes. The ROCs for \textit{Buildings}, \textit{Tree} and \textit{Car} in Vaihingen also highlight the superiority of OpenPCS when compared to the other approaches.

Figure~\ref{fig:roc_potsdam} provides distinct insights, as Potsdam is a much harder dataset than Vaihingen. Figures~\ref{fig:roc_fcnwideresnet50_potsdam_1_area_6_7} and~\ref{fig:roc_fcnwideresnet50_potsdam_4_area_6_7} show that OpenPCS and its online/incremental implementation (OpenIPCS) behaves dramatically better than both OpenFCN and \softmaxt{}. OpenFCN and \softmaxt{} stay mostly close to randomly selecting pixels in the images, which is delineated in the figures by the 45 degree dashed line in the plot. For \textit{Building} and \textit{Car}, the ROCs and AUCs show that \softmaxt{} and OpenFCN achieved worse than random results, while OpenPCS and OpenIPCS reached close to or above 0.90 in AUC.

Analysis of the per-class ROCs and Tables reveal that OpenPCS (and its variant OpenIPCS) behave considerably better than both \softmaxt{} and OpenFCN in the binary task of Pixel Anomaly Detection for most UUCs. One can clearly see the distinction of the distinct methods in Pixel Anomaly Detection performance between UUCs in Figures~\ref{fig:roc_vaihingen} and~\ref{fig:roc_potsdam}. OpenPCS excels by a large margin on some UUCS, especially \textit{Building} and \textit{Tree} on Vaihingen, presenting lower FPR values than OpenFCN and \softmaxt{} by almost any TPR threshold in The ROC. The evaluation on the other classes is more nuanced, as the plots cross at some points. OpenPCS tends to perform better on the lower end of FPR detections, providing the first evidence that OSR using Principal Components lessens the effect of UUC segmentation on the KKC classification performance, as smaller FPRs indicate that a smaller number of KKC pixels were misclassified as UUCs. This relation between UUCs and their distinction in performance on both KKCs and UUCs will further explored in Sections~\ref{sec:quantitative_vaihingen} and~\ref{sec:quantitative_potsdam}.

\subsubsection{Performance on Vaihingen for KKCs and UUCs}
\label{sec:quantitative_vaihingen}

OSR tasks are inherently multi-objective, as Open Set algorithms must be able to successfully discern UUCs from KKCs, while still being able to correctly classify samples from KKCs. Tables~\ref{tab:tpr_vaihingen_wideresnet50} and~\ref{tab:tpr_vaihingen_densenet121} show results from $Acc^{K}$, $Pre^{U}$ and $\kappa$ in the Vaihingen dataset using FCNs with WRN-50 and DenseNet-121 backbones, respectively.

\begin{table}[!ht]
    \centering
    \caption{Results for different unknown TPR thresholds applied to \softmaxt{}, OpenFCN and OpenPCS from an FCN with \textbf{WRN-50} backbone in \textbf{Vaihingen}. Bold values indicate the best overall $Acc^{K}$, $Pre^{U}$ and $\kappa$ results for a certain UUC. ``Closed'' rows mean unknown TPRs of 0.0, which is equivalent to Closed Set segmentation.}
    \label{tab:tpr_vaihingen_wideresnet50}
    \resizebox{\textwidth}{!}{%
    \begin{tabular}{|c|c|c|c|c|c|c|c|c|c|}
        \hline
        \multicolumn{10}{|c|}{\textbf{UUC: Impervious Surfaces}}                                                                                                                                                                                    \\ \hline
        \multirow{2}{*}{\textbf{TPR}} & \multicolumn{3}{c|}{\textbf{SoftMax$^{\mathcal{T}}$}}              & \multicolumn{3}{c|}{\textbf{OpenFCN}}                            & \multicolumn{3}{c|}{\textbf{OpenPCS}}                               \\ \cline{2-10} 
                                      & \textbf{$Acc^{K}$}   & \textbf{$Pre^{U}$}            & \textbf{$\kappa$}    & \textbf{$Acc^{K}$}   & \textbf{$Pre^{U}$} & \textbf{$\kappa$}    & \textbf{$Acc^{K}$}   & \textbf{$Pre^{U}$}    & \textbf{$\kappa$}    \\ \hline
        \textbf{Closed}               & \textbf{.86$\pm$.01} & .00$\pm$.00          & .53$\pm$.04          & \textbf{.86$\pm$.01} & .00$\pm$.00        & .53$\pm$.04          & \textbf{.86$\pm$.01} & .00$\pm$.00           & .53$\pm$.04          \\ \hline
        \textbf{0.10}                 & .85$\pm$.01          & .61$\pm$.11          & .55$\pm$.03          & .85$\pm$.01          & .77$\pm$.07        & .56$\pm$.04          & \textbf{.86$\pm$.01} & \textbf{1.00$\pm$.00} & .56$\pm$.04          \\ \hline
        \textbf{0.30}                 & .81$\pm$.02          & .52$\pm$.10          & .57$\pm$.02          & .83$\pm$.02          & .63$\pm$.10        & .59$\pm$.02          & \textbf{.86$\pm$.01} & .95$\pm$.03           & .62$\pm$.02          \\ \hline
        \textbf{0.50}                 & .76$\pm$.02          & .50$\pm$.09          & .59$\pm$.02          & .78$\pm$.02          & .55$\pm$.10        & .61$\pm$.02          & .84$\pm$.02          & .85$\pm$.08           & \textbf{.68$\pm$.02} \\ \hline
        \textbf{0.70}                 & .70$\pm$.03          & .47$\pm$.10          & .59$\pm$.03          & .72$\pm$.03          & .50$\pm$.10        & .62$\pm$.02          & .78$\pm$.04          & .64$\pm$.13           & .67$\pm$.04          \\ \hline
        \textbf{0.90}                 & .56$\pm$.05          & .42$\pm$.10          & .52$\pm$.06          & .59$\pm$.04          & .44$\pm$.09        & .56$\pm$.05          & .53$\pm$.07          & .41$\pm$.10           & .49$\pm$.08          \\ \hline
        \multicolumn{10}{|c|}{\textbf{UUC: Building}}                                                                                                                                                                                               \\ \hline
        \multirow{2}{*}{\textbf{TPR}} & \multicolumn{3}{c|}{\textbf{SoftMax$^{\mathcal{T}}$}}              & \multicolumn{3}{c|}{\textbf{OpenFCN}}                            & \multicolumn{3}{c|}{\textbf{OpenPCS}}                               \\ \cline{2-10} 
                                      & \textbf{$Acc^{K}$}   & \textbf{$Pre^{U}$}   & \textbf{$\kappa$}    & \textbf{$Acc^{K}$}   & \textbf{$Pre^{U}$} & \textbf{$\kappa$}    & \textbf{$Acc^{K}$}   & \textbf{$Pre^{U}$}    & \textbf{$\kappa$}    \\ \hline
        \textbf{Closed}               & \textbf{.82$\pm$.02} & .00$\pm$.00          & .49$\pm$.05          & \textbf{.82$\pm$.02} & .00$\pm$.00        & .49$\pm$.05          & \textbf{.82$\pm$.02} & .00$\pm$.00           & .49$\pm$.05          \\ \hline
        \textbf{0.10}                 & .79$\pm$.02          & .34$\pm$.07          & .49$\pm$.05          & .79$\pm$.02          & .34$\pm$.08        & .49$\pm$.05          & \textbf{.82$\pm$.02} & \textbf{1.00$\pm$.00} & .53$\pm$.05          \\ \hline
        \textbf{0.30}                 & .73$\pm$.03          & .36$\pm$.06          & .49$\pm$.04          & .73$\pm$.03          & .36$\pm$.07        & .49$\pm$.04          & \textbf{.82$\pm$.02} & .99$\pm$.01           & .59$\pm$.04          \\ \hline
        \textbf{0.50}                 & .65$\pm$.03          & .36$\pm$.05          & .48$\pm$.04          & .65$\pm$.03          & .36$\pm$.06        & .49$\pm$.04          & \textbf{.82$\pm$.02} & .93$\pm$.04           & .65$\pm$.03          \\ \hline
        \textbf{0.70}                 & .52$\pm$.05          & .33$\pm$.04          & .42$\pm$.04          & .54$\pm$.03          & .35$\pm$.05        & .44$\pm$.03          & .80$\pm$.03          & .85$\pm$.06           & \textbf{.70$\pm$.03} \\ \hline
        \textbf{0.90}                 & .23$\pm$.12          & .28$\pm$.04          & .20$\pm$.11          & .34$\pm$.05          & .31$\pm$.05        & .30$\pm$.04          & .71$\pm$.04          & .64$\pm$.09           & .68$\pm$.04          \\ \hline
        \multicolumn{10}{|c|}{\textbf{UUC: Low Vegetation}}                                                                                                                                                                                         \\ \hline
        \multirow{2}{*}{\textbf{TPR}} & \multicolumn{3}{c|}{\textbf{SoftMax$^{\mathcal{T}}$}}              & \multicolumn{3}{c|}{\textbf{OpenFCN}}                            & \multicolumn{3}{c|}{\textbf{OpenPCS}}                               \\ \cline{2-10} 
                                      & \textbf{$Acc^{K}$}   & \textbf{$Pre^{U}$}   & \textbf{$\kappa$}    & \textbf{$Acc^{K}$}   & \textbf{$Pre^{U}$} & \textbf{$\kappa$}    & \textbf{$Acc^{K}$}   & \textbf{$Pre^{U}$}    & \textbf{$\kappa$}    \\ \hline
        \textbf{Closed}               & \textbf{.92$\pm$.01} & .00$\pm$.00          & .61$\pm$.04          & \textbf{.92$\pm$.01} & .00$\pm$.00        & .61$\pm$.04          & \textbf{.92$\pm$.01} & .00$\pm$.00           & .61$\pm$.04          \\ \hline
        \textbf{0.10}                 & .91$\pm$.01          & \textbf{.47$\pm$.05} & .62$\pm$.04          & .90$\pm$.01          & .46$\pm$.05        & .62$\pm$.04          & .90$\pm$.02          & .46$\pm$.12           & .62$\pm$.04          \\ \hline
        \textbf{0.30}                 & .86$\pm$.02          & .45$\pm$.05          & \textbf{.64$\pm$.04} & .86$\pm$.02          & .44$\pm$.06        & .63$\pm$.04          & .84$\pm$.04          & \textbf{.47$\pm$.12}  & .62$\pm$.05          \\ \hline
        \textbf{0.50}                 & .79$\pm$.03          & .43$\pm$.05          & .63$\pm$.04          & .77$\pm$.04          & .41$\pm$.06        & .61$\pm$.05          & .77$\pm$.05          & .44$\pm$.08           & .61$\pm$.05          \\ \hline
        \textbf{0.70}                 & .66$\pm$.07          & .39$\pm$.05          & .56$\pm$.07          & .63$\pm$.07          & .36$\pm$.06        & .53$\pm$.08          & .67$\pm$.05          & .41$\pm$.06           & .57$\pm$.05          \\ \hline
        \textbf{0.90}                 & .44$\pm$.09          & .32$\pm$.05          & .40$\pm$.09          & .43$\pm$.08          & .32$\pm$.04        & .39$\pm$.08          & .49$\pm$.04          & .35$\pm$.05           & .45$\pm$.04          \\ \hline
        \multicolumn{10}{|c|}{\textbf{UUC: Tree}}                                                                                                                                                                                                   \\ \hline
        \multirow{2}{*}{\textbf{TPR}} & \multicolumn{3}{c|}{\textbf{SoftMax$^{\mathcal{T}}$}}              & \multicolumn{3}{c|}{\textbf{OpenFCN}}                            & \multicolumn{3}{c|}{\textbf{OpenPCS}}                               \\ \cline{2-10} 
                                      & \textbf{$Acc^{K}$}   & \textbf{$Pre^{U}$}   & \textbf{$\kappa$}    & \textbf{$Acc^{K}$}   & \textbf{$Pre^{U}$} & \textbf{$\kappa$}    & \textbf{$Acc^{K}$}   & \textbf{$Pre^{U}$}    & \textbf{$\kappa$}    \\ \hline
        \textbf{Closed}               & \textbf{.88$\pm$.02} & .00$\pm$.00          & .54$\pm$.07          & \textbf{.88$\pm$.02} & .00$\pm$.00        & .54$\pm$.07          & \textbf{.88$\pm$.02} & .00$\pm$.00           & .54$\pm$.07          \\ \hline
        \textbf{0.10}                 & .81$\pm$.04          & .22$\pm$.05          & .50$\pm$.08          & .81$\pm$.04          & .23$\pm$.04        & .51$\pm$.08          & \textbf{.88$\pm$.02} & \textbf{.88$\pm$.08}  & \textbf{.57$\pm$.06} \\ \hline
        \textbf{0.30}                 & .70$\pm$.05          & .28$\pm$.06          & .46$\pm$.08          & .69$\pm$.06          & .27$\pm$.05        & .45$\pm$.08          & .81$\pm$.07          & .62$\pm$.17           & \textbf{.57$\pm$.10} \\ \hline
        \textbf{0.50}                 & .62$\pm$.06          & .32$\pm$.06          & .45$\pm$.07          & .60$\pm$.06          & .31$\pm$.06        & .43$\pm$.07          & .61$\pm$.14          & .37$\pm$.08           & .44$\pm$.16          \\ \hline
        \textbf{0.70}                 & .53$\pm$.06          & .35$\pm$.07          & .43$\pm$.06          & .51$\pm$.06          & .34$\pm$.07        & .41$\pm$.06          & .38$\pm$.07          & .29$\pm$.05           & .28$\pm$.07          \\ \hline
        \textbf{0.90}                 & .40$\pm$.05          & .34$\pm$.06          & .37$\pm$.05          & .37$\pm$.05          & .33$\pm$.06        & .33$\pm$.05          & .17$\pm$.06          & .28$\pm$.05           & .14$\pm$.06          \\ \hline
        \multicolumn{10}{|c|}{\textbf{UUC: Car}}                                                                                                                                                                                                    \\ \hline
        \multirow{2}{*}{\textbf{TPR}} & \multicolumn{3}{c|}{\textbf{SoftMax$^{\mathcal{T}}$}}              & \multicolumn{3}{c|}{\textbf{OpenFCN}}                            & \multicolumn{3}{c|}{\textbf{OpenPCS}}                               \\ \cline{2-10} 
                                      & \textbf{$Acc^{K}$}   & \textbf{$Pre^{U}$}   & \textbf{$\kappa$}    & \textbf{$Acc^{K}$}   & \textbf{$Pre^{U}$} & \textbf{$\kappa$}    & \textbf{$Acc^{K}$}   & \textbf{$Pre^{U}$}    & \textbf{$\kappa$}    \\ \hline
        \textbf{Closed}               & \textbf{.84$\pm$.02} & .00$\pm$.00          & \textbf{.77$\pm$.02} & \textbf{.84$\pm$.02} & .00$\pm$.00        & \textbf{.77$\pm$.02} & \textbf{.84$\pm$.02} & .00$\pm$.00           & \textbf{.77$\pm$.02} \\ \hline
        \textbf{0.10}                 & .77$\pm$.05          & .01$\pm$.01          & .69$\pm$.05          & .77$\pm$.04          & .01$\pm$.01        & .70$\pm$.04          & \textbf{.84$\pm$.02} & \textbf{.27$\pm$.06}  & \textbf{.77$\pm$.02} \\ \hline
        \textbf{0.30}                 & .64$\pm$.06          & .01$\pm$.01          & .56$\pm$.06          & .64$\pm$.06          & .01$\pm$.01        & .56$\pm$.06          & .83$\pm$.02          & .25$\pm$.05           & \textbf{.77$\pm$.02} \\ \hline
        \textbf{0.50}                 & .54$\pm$.05          & .01$\pm$.01          & .46$\pm$.05          & .55$\pm$.06          & .01$\pm$.01        & .47$\pm$.05          & .81$\pm$.02          & .13$\pm$.04           & .75$\pm$.02          \\ \hline
        \textbf{0.70}                 & .45$\pm$.04          & .02$\pm$.01          & .38$\pm$.04          & .46$\pm$.05          & .02$\pm$.01        & .39$\pm$.04          & .72$\pm$.04          & .05$\pm$.02           & .64$\pm$.04          \\ \hline
        \textbf{0.90}                 & .35$\pm$.04          & .02$\pm$.01          & .28$\pm$.03          & .34$\pm$.04          & .02$\pm$.01        & .28$\pm$.03          & .56$\pm$.05          & .03$\pm$.01           & .48$\pm$.05          \\ \hline
    \end{tabular}
    }
\end{table}

\begin{table}[!ht]
    \centering
    \caption{Results for different unknown TPR thresholds applied to \softmaxt{}, OpenFCN and OpenPCS from an FCN with \textbf{DenseNet-121} backbone in \textbf{Vaihingen}. Bold values indicate the best overall $Acc^{K}$, $Pre^{U}$ and $\kappa$ results for a certain UUC. ``Closed'' rows mean unknown TPRs of 0.0, which is equivalent to Closed Set segmentation.}
    \label{tab:tpr_vaihingen_densenet121}
    \resizebox{\textwidth}{!}{%
    \begin{tabular}{|c|c|c|c|c|c|c|c|c|c|}
        \hline
        \multicolumn{10}{|c|}{\textbf{UUC: Impervious Surfaces}}                                                                                                                                                                                 \\ \hline
        \multirow{2}{*}{\textbf{TPR}} & \multicolumn{3}{c|}{\textbf{SoftMax$^{\mathcal{T}}$}}            & \multicolumn{3}{c|}{\textbf{OpenFCN}}                            & \multicolumn{3}{c|}{\textbf{OpenPCS}}                              \\ \cline{2-10} 
                                      & \textbf{$Acc^{K}$}   & \textbf{$Pre^{U}$}          & \textbf{$\kappa$}    & \textbf{$Acc^{K}$}   & \textbf{$Pre^{U}$} & \textbf{$\kappa$}    & \textbf{$Acc^{K}$}   & \textbf{$Pre^{U}$}   & \textbf{$\kappa$}    \\ \hline
        \textbf{Closed}               & \textbf{.85$\pm$.02} & .00$\pm$.00        & .52$\pm$.04          & \textbf{.85$\pm$.02} & .00$\pm$.00        & .52$\pm$.04          & \textbf{.85$\pm$.02} & .00$\pm$.00          & .52$\pm$.04          \\ \hline
        \textbf{0.10}                 & .83$\pm$.03          & .47$\pm$.11        & .53$\pm$.04          & \textbf{.85$\pm$.02} & .67$\pm$.12        & .55$\pm$.04          & \textbf{.85$\pm$.02} & \textbf{.97$\pm$.02} & .55$\pm$.04          \\ \hline
        \textbf{0.30}                 & .79$\pm$.03          & .48$\pm$.10        & .56$\pm$.02          & .81$\pm$.03          & .56$\pm$.11        & .58$\pm$.02          & \textbf{.85$\pm$.03} & .91$\pm$.07          & .61$\pm$.03          \\ \hline
        \textbf{0.50}                 & .74$\pm$.03          & .47$\pm$.11        & .57$\pm$.02          & .76$\pm$.03          & .52$\pm$.11        & .59$\pm$.02          & .82$\pm$.04          & .77$\pm$.14          & \textbf{.65$\pm$.03} \\ \hline
        \textbf{0.70}                 & .67$\pm$.04          & .45$\pm$.10        & .57$\pm$.04          & .70$\pm$.04          & .49$\pm$.11        & .59$\pm$.03          & .71$\pm$.09          & .57$\pm$.15          & .61$\pm$.09          \\ \hline
        \textbf{0.90}                 & .54$\pm$.05          & .41$\pm$.09        & .51$\pm$.06          & .58$\pm$.04          & .43$\pm$.10        & .54$\pm$.05          & .38$\pm$.06          & .35$\pm$.08          & .35$\pm$.06          \\ \hline
        \multicolumn{10}{|c|}{\textbf{UUC: Building}}                                                                                                                                                                                            \\ \hline
        \multirow{2}{*}{\textbf{TPR}} & \multicolumn{3}{c|}{\textbf{SoftMax$^{\mathcal{T}}$}}            & \multicolumn{3}{c|}{\textbf{OpenFCN}}                            & \multicolumn{3}{c|}{\textbf{OpenPCS}}                              \\ \cline{2-10} 
                                      & \textbf{$Acc^{K}$}   & \textbf{$Pre^{U}$} & \textbf{$\kappa$}    & \textbf{$Acc^{K}$}   & \textbf{$Pre^{U}$} & \textbf{$\kappa$}    & \textbf{$Acc^{K}$}   & \textbf{$Pre^{U}$}   & \textbf{$\kappa$}    \\ \hline
        \textbf{Closed}               & \textbf{.83$\pm$.03} & .00$\pm$.00        & .50$\pm$.06          & \textbf{.83$\pm$.03} & .00$\pm$.00        & .50$\pm$.06          & \textbf{.83$\pm$.03} & .00$\pm$.00          & .50$\pm$.06          \\ \hline
        \textbf{0.10}                 & .77$\pm$.03          & .26$\pm$.06        & .48$\pm$.05          & .78$\pm$.03          & .28$\pm$.07        & .48$\pm$.05          & .82$\pm$.03          & \textbf{.98$\pm$.02} & .53$\pm$.05          \\ \hline
        \textbf{0.30}                 & .70$\pm$.03          & .32$\pm$.06        & .47$\pm$.04          & .71$\pm$.03          & .32$\pm$.06        & .47$\pm$.04          & .82$\pm$.03          & .96$\pm$.02          & .59$\pm$.05          \\ \hline
        \textbf{0.50}                 & .62$\pm$.03          & .33$\pm$.06        & .45$\pm$.04          & .62$\pm$.03          & .34$\pm$.06        & .46$\pm$.04          & .82$\pm$.03          & .94$\pm$.03          & .65$\pm$.04          \\ \hline
        \textbf{0.70}                 & .50$\pm$.06          & .33$\pm$.06        & .40$\pm$.06          & .51$\pm$.06          & .33$\pm$.06        & .41$\pm$.06          & .80$\pm$.03          & .87$\pm$.06          & \textbf{.70$\pm$.03} \\ \hline
        \textbf{0.90}                 & .30$\pm$.13          & .30$\pm$.06        & .26$\pm$.13          & .29$\pm$.15          & .30$\pm$.06        & .25$\pm$.14          & .71$\pm$.04          & .64$\pm$.07          & .68$\pm$.03          \\ \hline
        \multicolumn{10}{|c|}{\textbf{UUC: Low Vegetation}}                                                                                                                                                                                      \\ \hline
        \multirow{2}{*}{\textbf{TPR}} & \multicolumn{3}{c|}{\textbf{SoftMax$^{\mathcal{T}}$}}            & \multicolumn{3}{c|}{\textbf{OpenFCN}}                            & \multicolumn{3}{c|}{\textbf{OpenPCS}}                              \\ \cline{2-10} 
                                      & \textbf{$Acc^{K}$}   & \textbf{$Pre^{U}$} & \textbf{$\kappa$}    & \textbf{$Acc^{K}$}   & \textbf{$Pre^{U}$} & \textbf{$\kappa$}    & \textbf{$Acc^{K}$}   & \textbf{$Pre^{U}$}   & \textbf{$\kappa$}    \\ \hline
        \textbf{Closed}               & \textbf{.92$\pm$.01} & .00$\pm$.00        & .61$\pm$.04          & \textbf{.92$\pm$.01} & .00$\pm$.00        & .61$\pm$.04          & \textbf{.92$\pm$.01} & .00$\pm$.00          & .61$\pm$.04          \\ \hline
        \textbf{0.10}                 & .90$\pm$.01          & .45$\pm$.08        & .62$\pm$.04          & .90$\pm$.01          & .46$\pm$.09        & .62$\pm$.04          & \textbf{.92$\pm$.01} & \textbf{.70$\pm$.06} & .64$\pm$.04          \\ \hline
        \textbf{0.30}                 & .85$\pm$.02          & .43$\pm$.07        & .63$\pm$.04          & .85$\pm$.03          & .43$\pm$.08        & .63$\pm$.04          & .88$\pm$.02          & .56$\pm$.05          & \textbf{.66$\pm$.04} \\ \hline
        \textbf{0.50}                 & .77$\pm$.04          & .41$\pm$.06        & .60$\pm$.05          & .76$\pm$.05          & .40$\pm$.06        & .60$\pm$.06          & .77$\pm$.06          & .44$\pm$.07          & .61$\pm$.07          \\ \hline
        \textbf{0.70}                 & .64$\pm$.06          & .37$\pm$.04        & .54$\pm$.06          & .61$\pm$.07          & .35$\pm$.05        & .51$\pm$.07          & .55$\pm$.14          & .33$\pm$.07          & .45$\pm$.14          \\ \hline
        \textbf{0.90}                 & .42$\pm$.07          & .31$\pm$.04        & .38$\pm$.07          & .39$\pm$.06          & .30$\pm$.04        & .35$\pm$.06          & .23$\pm$.11          & .26$\pm$.04          & .19$\pm$.11          \\ \hline
        \multicolumn{10}{|c|}{\textbf{UUC: Tree}}                                                                                                                                                                                                \\ \hline
        \multirow{2}{*}{\textbf{TPR}} & \multicolumn{3}{c|}{\textbf{SoftMax$^{\mathcal{T}}$}}            & \multicolumn{3}{c|}{\textbf{OpenFCN}}                            & \multicolumn{3}{c|}{\textbf{OpenPCS}}                              \\ \cline{2-10} 
                                      & \textbf{$Acc^{K}$}   & \textbf{$Pre^{U}$} & \textbf{$\kappa$}    & \textbf{$Acc^{K}$}   & \textbf{$Pre^{U}$} & \textbf{$\kappa$}    & \textbf{$Acc^{K}$}   & \textbf{$Pre^{U}$}   & \textbf{$\kappa$}    \\ \hline
        \textbf{Closed}               & \textbf{.87$\pm$.02} & .00$\pm$.00        & .53$\pm$.05          & \textbf{.87$\pm$.02} & .00$\pm$.00        & .53$\pm$.05          & \textbf{.87$\pm$.02} & .00$\pm$.00          & .53$\pm$.05          \\ \hline
        \textbf{0.10}                 & .84$\pm$.02          & .33$\pm$.09        & .53$\pm$.05          & .84$\pm$.02          & .34$\pm$.09        & .53$\pm$.05          & \textbf{.87$\pm$.02} & \textbf{.97$\pm$.04} & .56$\pm$.05          \\ \hline
        \textbf{0.30}                 & .77$\pm$.04          & .37$\pm$.09        & .53$\pm$.05          & .77$\pm$.04          & .37$\pm$.09        & .53$\pm$.06          & .86$\pm$.02          & .82$\pm$.11          & .62$\pm$.05          \\ \hline
        \textbf{0.50}                 & .68$\pm$.05          & .37$\pm$.09        & .51$\pm$.06          & .68$\pm$.06          & .37$\pm$.08        & .50$\pm$.06          & .80$\pm$.04          & .64$\pm$.10          & \textbf{.63$\pm$.06} \\ \hline
        \textbf{0.70}                 & .57$\pm$.06          & .37$\pm$.08        & .46$\pm$.06          & .54$\pm$.07          & .35$\pm$.08        & .44$\pm$.07          & .70$\pm$.07          & .50$\pm$.04          & .60$\pm$.08          \\ \hline
        \textbf{0.90}                 & .35$\pm$.10          & .33$\pm$.09        & .31$\pm$.11          & .30$\pm$.08          & .31$\pm$.08        & .27$\pm$.08          & .47$\pm$.10          & .38$\pm$.04          & .43$\pm$.09          \\ \hline
        \multicolumn{10}{|c|}{\textbf{UUC: Car}}                                                                                                                                                                                                 \\ \hline
        \multirow{2}{*}{\textbf{TPR}} & \multicolumn{3}{c|}{\textbf{SoftMax$^{\mathcal{T}}$}}            & \multicolumn{3}{c|}{\textbf{OpenFCN}}                            & \multicolumn{3}{c|}{\textbf{OpenPCS}}                              \\ \cline{2-10} 
                                      & \textbf{$Acc^{K}$}   & \textbf{$Pre^{U}$} & \textbf{$\kappa$}    & \textbf{$Acc^{K}$}   & \textbf{$Pre^{U}$} & \textbf{$\kappa$}    & \textbf{$Acc^{K}$}   & \textbf{$Pre^{U}$}   & \textbf{$\kappa$}    \\ \hline
        \textbf{Closed}               & \textbf{.84$\pm$.02} & .00$\pm$.00        & \textbf{.77$\pm$.03} & \textbf{.84$\pm$.02} & .00$\pm$.00        & \textbf{.77$\pm$.03} & \textbf{.84$\pm$.02} & .00$\pm$.00          & \textbf{.77$\pm$.03} \\ \hline
        \textbf{0.10}                 & .77$\pm$.04          & .01$\pm$.01        & .69$\pm$.04          & .76$\pm$.04          & .01$\pm$.00        & .68$\pm$.04          & .83$\pm$.02          & \textbf{.17$\pm$.08} & .76$\pm$.02          \\ \hline
        \textbf{0.30}                 & .66$\pm$.06          & .01$\pm$.01        & .58$\pm$.06          & .65$\pm$.05          & .01$\pm$.01        & .57$\pm$.05          & .80$\pm$.04          & .07$\pm$.03          & .72$\pm$.04          \\ \hline
        \textbf{0.50}                 & .55$\pm$.07          & .01$\pm$.01        & .47$\pm$.06          & .56$\pm$.05          & .01$\pm$.01        & .48$\pm$.05          & .68$\pm$.07          & .03$\pm$.01          & .60$\pm$.08          \\ \hline
        \textbf{0.70}                 & .45$\pm$.06          & .02$\pm$.01        & .38$\pm$.06          & .47$\pm$.06          & .02$\pm$.01        & .39$\pm$.05          & .40$\pm$.10          & .02$\pm$.01          & .33$\pm$.09          \\ \hline
        \textbf{0.90}                 & .32$\pm$.05          & .02$\pm$.01        & .26$\pm$.04          & .34$\pm$.05          & .02$\pm$.01        & .27$\pm$.04          & .13$\pm$.05          & .01$\pm$.00          & .10$\pm$.04          \\ \hline
    \end{tabular}
    }
\end{table}

Tables~\ref{tab:tpr_vaihingen_wideresnet50}, and~\ref{tab:tpr_vaihingen_densenet121} show the performance of the OSR segmentation approaches on both KKCs and UUCs. The comparison of TPRs larger than one -- that is, thresholds that allow for OSR -- with their Closed Set counterparts (TPR = 0) using $\kappa$ reveals that, in many cases, assuming openness improves object recognition in scenarios where full knowledge of the world is not possible. The most noticeable improvements in $\kappa$ happen in UUCs that are visually and semantically more distinct from the other classes in the dataset: \textit{Impervious Surfaces}, \textit{Building}, \textit{Tree} and \textit{Car}. In the following paragraphs we will discuss the performance of the methods for each UUC in LOCO, starting from \textit{Impervious Surfaces}.

\textit{Impervious Surfaces} pixels in both Vaihingen and Potsdam are mainly composed of streets and driveways in residential/commercial buildings. These surfaces are considerably distinct from vegetation areas, as the ``gray'' signature in IRRG bands of asphalt is considerably distinct from the redness of plants due to their large reflection of near-infrared radiation. On Vaihingen, $\kappa$ using fully closed DNNs achieve 0.53 with WRN-50 and and 0.52 with DenseNet-121. By adding openness to these DNNs via OpenPCS, we were able to increase these $\kappa$ metrics to 0.68 and 0.65, respectively, representing gains in $\kappa$ between 0.13 and 0.15. The OSR methods allowed the methods to correctly delineate most streets and driveways in the samples -- as can also be seen in a qualitative manner in Section~\ref{sec:qualitative}. \softmaxt{} and OpenFCN also improved the performance of the closed DNNs for \textit{Impervious Surfaces}, albeit in a smaller magnitude when compared to OpenPCS, achieving gains of between 0.05 and 0.09.

Closed Set WRN-50 achieved an $Acc^{K}$ of 0.86, followed closely by DenseNet-121 with 0.85 when \textit{Impervious Surface} pixels were set as UUC in LOCO. For the same TPR, $Acc^{K}$ was only barely affected by the openness of OpenPCS, maintaining their original levels of 0.86 and 0.85 up until a TPR of 0.30. In contrast, $Acc^{K}$ was considerably sacrificed in \softmaxt{} and OpenFCN for TPRs larger than 0.10, meaning that pixels that were correctly classified by the original DNNs were cast as UUCs by the OSR post-processing. OpenPCS maintained large $Acc^{K}$ values up until TPRs of 0.50, with a large drop only being seen in a TPR of 0.70. These results imply that OpenPCS is more accuracy efficient when identifying OOD \textit{Impervious Surface} pixels, barely sacrificing KKC segmentation performances. The precision of unknowns ($Pre^{U}$) is another evidence for the superiority of OpenPCS compared to OpenFCN and \softmaxt{}, as it reaches a perfect precision of 1.00 in WRN-50 for a TPR (recall) of 0.10, followed closely by DenseNet's $Pre^{U}$ of 0.97 for the same TPR. $Pre^{U}$ remains larger than 0.90 for both WRN-50 and DenseNet-121 up to a TPR of 0.30, meaning that UUC predictions for \textit{Impervious Surfaces} from OpenPCS were highly reliable, even allowing for real-world applications of the method. Similarly to $Acc^{K}$, the precision of unknowns $Pre^{U}$ was significantly lower from the start on \softmaxt{} and OpenFCN, with the best results for these methods peaking (with 0.10 TPR using WRN-50) on 0.61 and 0.77, respectively.

The UUC \textit{Building} yielded the most drastic gains in $\kappa$ when OpenPCS was introduced, as showed by the increase from 0.50 to 0.70 with the DenseNet-121 backbone (Table~\ref{tab:tpr_vaihingen_densenet121}). \textit{Building} pixels are certainly among the most visually distinct features in the IRRG images in both texture and shape. In addition to that, these structures are rather distinct in the DSM data, as they can be easily seen on depth maps as sudden variations in altitude in comparison with the surrounding terrain. Finally, context also plays a larger role in \textit{Bulding} detection, as both commercial and residential structures reside close to driveways or parkways (both \textit{Street} samples) and are often surrounded by vegetation in suburban areas as Vaihingen and Potsdam. Predictions for UUC \textit{Building} using a TPR of 0.10 were also highly precise, with $Pre^{U}$ values of 1.0 (perfect scoring) and 0.98 (near perfect) for WRN-50 and DenseNet-121, respectively.

Overall results from UUC \textit{Tree} followed similar patterns to \textit{Street} and \textit{Building}, but in a smaller scale. $\kappa$ gains using OpenPCS with WRN-50 and DenseNet-121 backbones were in the scale of 0.03 and 0.10, respectively, showing a noticeable boost in performance mainly in DenseNet-121. In contrast, the best $\kappa$ values for OpenFCN and \softmaxt{} were the Closed Set ones, with $\kappa$ quickly dropping even since TPR 0.10 on WRN-50. The best $Pre^{U}$ was also achieved by OpenPCS on both backbones, reaching a maximum of 0.97 for a TPR of 0.10 on DenseNet-121, while no noticeable drop in $Acc^{K}$ could be seen in this setting. $Acc^{K}$ remained relatively close to the Closed Set KKC classification performance up to 0.30 of TPR with DenseNet-121 and up to 0.10 on WRN-50.

\textit{Car} pixels represent a tiny proportion of samples in both Potsdam and Vaihingen. Thus, analysis of Open vs. Closed Set methods by only using $\kappa$ would yield basically no gain in adding OSR to the pipeline, as can be seen in Tables~\ref{tab:tpr_vaihingen_wideresnet50} and~\ref{tab:tpr_vaihingen_densenet121}. This is due to the benefits of detecting UUCs not outweighing the added errors in KKC classification in metrics that measure general multiclass performance as Balanced Accuracy or $\kappa$. Instead, a more accurate picture of \textit{Car} segmentation can be seen when taking into account $Pre^{U}$ and $Acc^{K}$, together with the TPR, which is automatically given because it served to compute the thresholds for the methods. One can easily see that OpenPCS' $Acc^{K}$ performance was barely degraded for TPR values up to 0.50 in WRN-50, while also remaining considerably close to the Closed Set accuracy in DenseNet-121 until a TPR of 0.30. In other words, there was virtually no harm to KKC classification in WRN-50 in exchange for the correct identification of 50\% of \textit{Car} pixels, while DenseNet-121 also kept high $Acc^{K}$ values for 30\% of the \textit{Car} pixels being correctly segmented.

$Pre^{U}$ values were not particularly high in either method, even for small TPRs, reaching a peak of 0.27 in WRN-50 with a TPR of 0.10 -- that is, only approximately one in four pixels predicted to be pertaining to a vehicle in this setting was really from a vehicle. The low $Pre^{U}$ of UUC \textit{Car} can be explained by the relatively high intra-class variability in class \textit{Building}, as most of the pixels misclassified as \textit{Car} were, in fact, from parts of \textit{Building} structures that were not common in the rest of the houses and warehouses in the data. We reached this conclusion after the qualitative evaluation shown in Section~\ref{sec:qualitative} that can be fully appreciated in the project's webpage. It is natural that houses have distinct shapes, sizes and rooftop textures, resulting. OpenPCS' $Pre^{U}$ results, though indeed small, are still more than one order of magnitude larger than both OpenFCN and \softmaxt{}.

Lastly, one important distinction between segmentation and detection must be highlighted in the case for \textit{Car}: while many vehicles were not perfectly segmented, almost all automobiles were at least partially identified by OpenPCS. Qualitative results available at the project's webpage show that in both datasets a large proportion of the automobiles were correctly identified by having a large subset of their pixels correctly identified as pertaining to a UUC. The same cannot be said for OpenFCN and \softmaxt{}, which correctly detected (even if only partially) a much lower proportion of vehicles. In addition to that, one can see that even for small TPRs of 0.10 OpenFCN and \softmaxt{} presented large drops to both $Acc^{K}$ and $\kappa$ performances.

\subsubsection{Performance on Potsdam for KKCs and UUCs}
\label{sec:quantitative_potsdam}

Similarly to the presentation of the results from Vaihingen, Tables~\ref{tab:tpr_potsdam_wideresnet50} and~\ref{tab:tpr_potsdam_densenet121} show the same per-class threshold-dependent metrics ($\kappa$, $Acc^{K}$ and $Pre^{U}$) in Potsdam for WRN-50 and DenseNet-121, respectively. As Potsdam is a considerably harder dataset when compared to Vaihingen -- mainly due to the large imbalance and much larger spatial resolution -- almost all metrics achieve lower values than the ones from Tables~\ref{tab:tpr_vaihingen_wideresnet50} and~\ref{tab:tpr_vaihingen_densenet121}. We highlight that the standard OpenPCS with random patch selection for the training of PCA was also computed for Potsdam (as shown in Figure~\ref{fig:roc_potsdam}), but this section only reports OpenIPCS values. The incremental training of PCA showed to be much more stable in the larger Potsdam dataset, while we choose to not shot the original OpenPCS results mostly due to spatial constraints and organization

\begin{table}[!ht]
    \centering
    \caption{Results for different unknown TPR thresholds applied to \softmaxt{}, OpenFCN and OpenIPCS from an FCN with \textbf{WRN-50} backbone in \textbf{Potsdam}. Bold values indicate the best overall $Acc^{K}$, $Pre^{U}$ and $\kappa$ results for a certain UUC. ``Closed'' rows mean unknown TPRs of 0.0, which is equivalent to Closed Set segmentation.}
    \label{tab:tpr_potsdam_wideresnet50}
    \resizebox{\textwidth}{!}{%
    \begin{tabular}{|c|c|c|c|c|c|c|c|c|c|}
        \hline
        \multicolumn{10}{|c|}{\textbf{UUC: Impervious Surfaces}}                                                                                                                                                                                     \\ \hline
        \multirow{2}{*}{\textbf{TPR}} & \multicolumn{3}{c|}{\textbf{SoftMax$^{\mathcal{T}}$}}              & \multicolumn{3}{c|}{\textbf{OpenFCN}}                              & \multicolumn{3}{c|}{\textbf{OpenIPCS}}                             \\ \cline{2-10} 
                                      & \textbf{$Acc^{K}$}   & \textbf{$Pre^{U}$}   & \textbf{$\kappa$}    & \textbf{$Acc^{K}$}   & \textbf{$Pre^{U}$}   & \textbf{$\kappa$}    & \textbf{$Acc^{K}$}   & \textbf{$Pre^{U}$}   & \textbf{$\kappa$}    \\ \hline
        \textbf{Closed}               & \textbf{.73$\pm$.09} & .00$\pm$.00          & .40$\pm$.09          & \textbf{.73$\pm$.09} & .00$\pm$.00          & .40$\pm$.09          & \textbf{.73$\pm$.09} & .00$\pm$.00          & .40$\pm$.09          \\ \hline
        \textbf{0.10}                 & .71$\pm$.09          & .58$\pm$.14          & .41$\pm$.08          & .72$\pm$.09          & \textbf{.70$\pm$.13} & .42$\pm$.08          & .71$\pm$.08          & .54$\pm$.18          & .41$\pm$.08          \\ \hline
        \textbf{0.30}                 & .66$\pm$.09          & .46$\pm$.11          & .43$\pm$.08          & .68$\pm$.09          & .51$\pm$.10          & \textbf{.44$\pm$.08} & .64$\pm$.08          & .47$\pm$.09          & .40$\pm$.07          \\ \hline
        \textbf{0.50}                 & .59$\pm$.10          & .40$\pm$.11          & .41$\pm$.09          & .60$\pm$.10          & .42$\pm$.11          & .42$\pm$.09          & .54$\pm$.08          & .40$\pm$.09          & .36$\pm$.07          \\ \hline
        \textbf{0.70}                 & .47$\pm$.11          & .35$\pm$.11          & .37$\pm$.10          & .48$\pm$.11          & .36$\pm$.11          & .37$\pm$.10          & .39$\pm$.08          & .33$\pm$.09          & .28$\pm$.07          \\ \hline
        \textbf{0.90}                 & .29$\pm$.11          & .31$\pm$.11          & .25$\pm$.10          & .28$\pm$.12          & .31$\pm$.11          & .25$\pm$.11          & .16$\pm$.04          & .28$\pm$.08          & .13$\pm$.03          \\ \hline
        \multicolumn{10}{|c|}{\textbf{UUC: Building}}                                                                                                                                                                                                \\ \hline
        \multirow{2}{*}{\textbf{TPR}} & \multicolumn{3}{c|}{\textbf{SoftMax$^{\mathcal{T}}$}}              & \multicolumn{3}{c|}{\textbf{OpenFCN}}                              & \multicolumn{3}{c|}{\textbf{OpenIPCS}}                             \\ \cline{2-10} 
                                      & \textbf{$Acc^{K}$}   & \textbf{$Pre^{U}$}   & \textbf{$\kappa$}    & \textbf{$Acc^{K}$}   & \textbf{$Pre^{U}$}   & \textbf{$\kappa$}    & \textbf{$Acc^{K}$}   & \textbf{$Pre^{U}$}   & \textbf{$\kappa$}    \\ \hline
        \textbf{Closed}               & \textbf{.77$\pm$.04} & .00$\pm$.00          & .44$\pm$.08          & \textbf{.77$\pm$.04} & .00$\pm$.00          & .44$\pm$.08          & \textbf{.77$\pm$.04} & .00$\pm$.00          & .44$\pm$.08          \\ \hline
        \textbf{0.10}                 & .74$\pm$.04          & .34$\pm$.15          & .43$\pm$.07          & .75$\pm$.04          & .38$\pm$.16          & .44$\pm$.07          & \textbf{.77$\pm$.04} & \textbf{.90$\pm$.06} & .46$\pm$.07          \\ \hline
        \textbf{0.30}                 & .64$\pm$.07          & .29$\pm$.15          & .40$\pm$.06          & .65$\pm$.07          & .30$\pm$.15          & .40$\pm$.06          & .76$\pm$.04          & .82$\pm$.05          & .51$\pm$.06          \\ \hline
        \textbf{0.50}                 & .52$\pm$.10          & .28$\pm$.15          & .35$\pm$.08          & .54$\pm$.09          & .29$\pm$.15          & .36$\pm$.07          & .74$\pm$.04          & .77$\pm$.08          & .56$\pm$.05          \\ \hline
        \textbf{0.70}                 & .37$\pm$.10          & .27$\pm$.14          & .26$\pm$.08          & .39$\pm$.09          & .28$\pm$.14          & .28$\pm$.07          & .71$\pm$.05          & .69$\pm$.14          & \textbf{.59$\pm$.06} \\ \hline
        \textbf{0.90}                 & .16$\pm$.07          & .26$\pm$.13          & .13$\pm$.06          & .18$\pm$.08          & .26$\pm$.13          & .14$\pm$.07          & .56$\pm$.13          & .48$\pm$.20          & .50$\pm$.14          \\ \hline
        \multicolumn{10}{|c|}{\textbf{UUC: Low Vegetation}}                                                                                                                                                                                          \\ \hline
        \multirow{2}{*}{\textbf{TPR}} & \multicolumn{3}{c|}{\textbf{SoftMax$^{\mathcal{T}}$}}              & \multicolumn{3}{c|}{\textbf{OpenFCN}}                              & \multicolumn{3}{c|}{\textbf{OpenIPCS}}                             \\ \cline{2-10} 
                                      & \textbf{$Acc^{K}$}   & \textbf{$Pre^{U}$}   & \textbf{$\kappa$}    & \textbf{$Acc^{K}$}   & \textbf{$Pre^{U}$}   & \textbf{$\kappa$}    & \textbf{$Acc^{K}$}   & \textbf{$Pre^{U}$}   & \textbf{$\kappa$}    \\ \hline
        \textbf{Closed}               & \textbf{.86$\pm$.04} & .00$\pm$.00          & .46$\pm$.14          & \textbf{.86$\pm$.04} & .00$\pm$.00          & .46$\pm$.14          & \textbf{.86$\pm$.04} & .00$\pm$.00          & .46$\pm$.14          \\ \hline
        \textbf{0.10}                 & .81$\pm$.04          & .37$\pm$.11          & .45$\pm$.14          & .81$\pm$.04          & .35$\pm$.11          & .44$\pm$.15          & .83$\pm$.05          & .55$\pm$.14          & .47$\pm$.14          \\ \hline
        \textbf{0.30}                 & .71$\pm$.09          & .39$\pm$.12          & .43$\pm$.16          & .69$\pm$.11          & .38$\pm$.12          & .41$\pm$.18          & .78$\pm$.07          & \textbf{.56$\pm$.15} & .49$\pm$.13          \\ \hline
        \textbf{0.50}                 & .60$\pm$.14          & .40$\pm$.12          & .40$\pm$.18          & .58$\pm$.15          & .39$\pm$.12          & .38$\pm$.19          & .72$\pm$.08          & .55$\pm$.15          & .51$\pm$.12          \\ \hline
        \textbf{0.70}                 & .49$\pm$.15          & .40$\pm$.13          & .37$\pm$.16          & .47$\pm$.15          & .39$\pm$.14          & .35$\pm$.16          & .64$\pm$.10          & .53$\pm$.15          & \textbf{.52$\pm$.11} \\ \hline
        \textbf{0.90}                 & .34$\pm$.13          & .39$\pm$.16          & .30$\pm$.11          & .31$\pm$.11          & .38$\pm$.16          & .27$\pm$.10          & .50$\pm$.13          & .47$\pm$.15          & .47$\pm$.10          \\ \hline
        \multicolumn{10}{|c|}{\textbf{UUC: Tree}}                                                                                                                                                                                                    \\ \hline
        \multirow{2}{*}{\textbf{TPR}} & \multicolumn{3}{c|}{\textbf{SoftMax$^{\mathcal{T}}$}}              & \multicolumn{3}{c|}{\textbf{OpenFCN}}                              & \multicolumn{3}{c|}{\textbf{OpenIPCS}}                             \\ \cline{2-10} 
                                      & \textbf{$Acc^{K}$}   & \textbf{$Pre^{U}$}   & \textbf{$\kappa$}    & \textbf{$Acc^{K}$}   & \textbf{$Pre^{U}$}   & \textbf{$\kappa$}    & \textbf{$Acc^{K}$}   & \textbf{$Pre^{U}$}   & \textbf{$\kappa$}    \\ \hline
        \textbf{Closed}               & \textbf{.89$\pm$.05} & .00$\pm$.00          & \textbf{.60$\pm$.07} & \textbf{.89$\pm$.05} & .00$\pm$.00          & \textbf{.60$\pm$.07} & \textbf{.89$\pm$.05} & .00$\pm$.00          & \textbf{.60$\pm$.07} \\ \hline
        \textbf{0.10}                 & .85$\pm$.05          & .26$\pm$.04          & .58$\pm$.07          & .85$\pm$.05          & .25$\pm$.04          & .58$\pm$.07          & .80$\pm$.06          & \textbf{.28$\pm$.17} & .53$\pm$.10          \\ \hline
        \textbf{0.30}                 & .77$\pm$.05          & \textbf{.28$\pm$.05} & .55$\pm$.08          & .76$\pm$.05          & .27$\pm$.05          & .54$\pm$.08          & .67$\pm$.09          & .23$\pm$.09          & .45$\pm$.12          \\ \hline
        \textbf{0.50}                 & .65$\pm$.08          & .26$\pm$.05          & .48$\pm$.11          & .63$\pm$.10          & .25$\pm$.06          & .46$\pm$.12          & .55$\pm$.11          & .23$\pm$.07          & .38$\pm$.13          \\ \hline
        \textbf{0.70}                 & .50$\pm$.12          & .25$\pm$.05          & .39$\pm$.13          & .47$\pm$.13          & .24$\pm$.05          & .35$\pm$.14          & .39$\pm$.13          & .21$\pm$.06          & .27$\pm$.13          \\ \hline
        \textbf{0.90}                 & .29$\pm$.11          & .22$\pm$.04          & .23$\pm$.11          & .24$\pm$.11          & .21$\pm$.04          & .18$\pm$.10          & .16$\pm$.09          & .19$\pm$.04          & .11$\pm$.08          \\ \hline
        \multicolumn{10}{|c|}{\textbf{UUC: Car}}                                                                                                                                                                                                     \\ \hline
        \multirow{2}{*}{\textbf{TPR}} & \multicolumn{3}{c|}{\textbf{SoftMax$^{\mathcal{T}}$}}              & \multicolumn{3}{c|}{\textbf{OpenFCN}}                              & \multicolumn{3}{c|}{\textbf{OpenIPCS}}                             \\ \cline{2-10} 
                                      & \textbf{$Acc^{K}$}   & \textbf{$Pre^{U}$}   & \textbf{$\kappa$}    & \textbf{$Acc^{K}$}   & \textbf{$Pre^{U}$}   & \textbf{$\kappa$}    & \textbf{$Acc^{K}$}   & \textbf{$Pre^{U}$}   & \textbf{$\kappa$}    \\ \hline
        \textbf{Closed}               & \textbf{.75$\pm$.07} & .00$\pm$.00          & \textbf{.64$\pm$.09} & \textbf{.75$\pm$.07} & .00$\pm$.00          & \textbf{.64$\pm$.09} & \textbf{.75$\pm$.07} & .00$\pm$.00          & \textbf{.64$\pm$.09} \\ \hline
        \textbf{0.10}                 & .74$\pm$.08          & .22$\pm$.13          & .63$\pm$.10          & \textbf{.75$\pm$.08} & .28$\pm$.16          & \textbf{.64$\pm$.09} & \textbf{.75$\pm$.07} & \textbf{.29$\pm$.15} & \textbf{.64$\pm$.09} \\ \hline
        \textbf{0.30}                 & .69$\pm$.12          & .04$\pm$.03          & .58$\pm$.12          & .70$\pm$.12          & .06$\pm$.03          & .60$\pm$.12          & .74$\pm$.07          & .14$\pm$.08          & .63$\pm$.09          \\ \hline
        \textbf{0.50}                 & .59$\pm$.13          & .02$\pm$.02          & .49$\pm$.12          & .62$\pm$.13          & .03$\pm$.02          & .52$\pm$.12          & .71$\pm$.06          & .07$\pm$.05          & .60$\pm$.08          \\ \hline
        \textbf{0.70}                 & .47$\pm$.13          & .02$\pm$.01          & .38$\pm$.11          & .51$\pm$.12          & .02$\pm$.01          & .41$\pm$.10          & .62$\pm$.07          & .04$\pm$.03          & .51$\pm$.09          \\ \hline
        \textbf{0.90}                 & .31$\pm$.13          & .02$\pm$.01          & .25$\pm$.10          & .34$\pm$.12          & .02$\pm$.01          & .27$\pm$.09          & .45$\pm$.09          & .02$\pm$.02          & .35$\pm$.09          \\ \hline
    \end{tabular}
    }
\end{table}

\begin{table}[!ht]
    \centering
    \caption{Results for different unknown TPR thresholds applied to \softmaxt{}, OpenFCN and OpenIPCS from an FCN with \textbf{DenseNet-121} backbone in \textbf{Potsdam}. Bold values indicate the best overall $Acc^{K}$, $Pre^{U}$ and $\kappa$ results for a certain UUC. ``Closed'' rows mean unknown TPRs of 0.0, which is equivalent to Closed Set segmentation.}
    \label{tab:tpr_potsdam_densenet121}
    \resizebox{\textwidth}{!}{%
    \begin{tabular}{|c|c|c|c|c|c|c|c|c|c|}
        \hline
        \multicolumn{10}{|c|}{\textbf{UUC: Impervious Surfaces}}                                                                                                                                                                                   \\ \hline
        \multirow{2}{*}{\textbf{TPR}} & \multicolumn{3}{c|}{\textbf{SoftMax$^{\mathcal{T}}$}}              & \multicolumn{3}{c|}{\textbf{OpenFCN}}                            & \multicolumn{3}{c|}{\textbf{OpenIPCS}}                             \\ \cline{2-10} 
                                      & \textbf{$Acc^{K}$}   & \textbf{$Pre^{U}$}   & \textbf{$\kappa$}    & \textbf{$Acc^{K}$}   & \textbf{$Pre^{U}$} & \textbf{$\kappa$}    & \textbf{$Acc^{K}$}   & \textbf{$Pre^{U}$}   & \textbf{$\kappa$}    \\ \hline
        \textbf{Closed}               & \textbf{.77$\pm$.08} & .00$\pm$.00          & .42$\pm$.10          & \textbf{.77$\pm$.08} & .00$\pm$.00        & .42$\pm$.10          & \textbf{.77$\pm$.08} & .00$\pm$.00          & .42$\pm$.10          \\ \hline
        \textbf{0.10}                 & .74$\pm$.08          & .44$\pm$.11          & .43$\pm$.09          & .74$\pm$.08          & .45$\pm$.11        & .43$\pm$.09          & .76$\pm$.08          & \textbf{.87$\pm$.11} & .45$\pm$.10          \\ \hline
        \textbf{0.30}                 & .68$\pm$.09          & .40$\pm$.14          & .43$\pm$.09          & .69$\pm$.09          & .42$\pm$.14        & .44$\pm$.09          & .75$\pm$.09          & .84$\pm$.10          & .51$\pm$.09          \\ \hline
        \textbf{0.50}                 & .59$\pm$.11          & .37$\pm$.14          & .41$\pm$.10          & .60$\pm$.10          & .37$\pm$.14        & .42$\pm$.10          & .73$\pm$.10          & .76$\pm$.12          & \textbf{.55$\pm$.09} \\ \hline
        \textbf{0.70}                 & .47$\pm$.14          & .35$\pm$.14          & .36$\pm$.13          & .47$\pm$.14          & .35$\pm$.14        & .36$\pm$.13          & .66$\pm$.10          & .63$\pm$.13          & \textbf{.55$\pm$.09} \\ \hline
        \textbf{0.90}                 & .29$\pm$.18          & .31$\pm$.13          & .25$\pm$.17          & .28$\pm$.18          & .31$\pm$.13        & .25$\pm$.16          & .46$\pm$.12          & .42$\pm$.11          & .40$\pm$.09          \\ \hline
        \multicolumn{10}{|c|}{\textbf{UUC: Building}}                                                                                                                                                                                              \\ \hline
        \multirow{2}{*}{\textbf{TPR}} & \multicolumn{3}{c|}{\textbf{SoftMax$^{\mathcal{T}}$}}              & \multicolumn{3}{c|}{\textbf{OpenFCN}}                            & \multicolumn{3}{c|}{\textbf{OpenIPCS}}                             \\ \cline{2-10} 
                                      & \textbf{$Acc^{K}$}   & \textbf{$Pre^{U}$}   & \textbf{$\kappa$}    & \textbf{$Acc^{K}$}   & \textbf{$Pre^{U}$} & \textbf{$\kappa$}    & \textbf{$Acc^{K}$}   & \textbf{$Pre^{U}$}   & \textbf{$\kappa$}    \\ \hline
        \textbf{Closed}               & \textbf{.74$\pm$.08} & .00$\pm$.00          & .41$\pm$.07          & \textbf{.74$\pm$.08} & .00$\pm$.00        & .41$\pm$.07          & \textbf{.74$\pm$.08} & .00$\pm$.00          & .41$\pm$.07          \\ \hline
        \textbf{0.10}                 & .71$\pm$.08          & .37$\pm$.17          & .41$\pm$.06          & .72$\pm$.08          & .40$\pm$.18        & .42$\pm$.06          & \textbf{.74$\pm$.08} & \textbf{.96$\pm$.05} & .44$\pm$.07          \\ \hline
        \textbf{0.30}                 & .61$\pm$.10          & .30$\pm$.17          & .37$\pm$.06          & .62$\pm$.09          & .32$\pm$.17        & .38$\pm$.06          & .74$\pm$.08          & .91$\pm$.11          & .49$\pm$.06          \\ \hline
        \textbf{0.50}                 & .46$\pm$.12          & .27$\pm$.15          & .29$\pm$.09          & .47$\pm$.11          & .28$\pm$.15        & .30$\pm$.08          & .73$\pm$.08          & .85$\pm$.13          & .55$\pm$.06          \\ \hline
        \textbf{0.70}                 & .25$\pm$.14          & .25$\pm$.14          & .16$\pm$.11          & .26$\pm$.12          & .25$\pm$.14        & .17$\pm$.10          & .69$\pm$.09          & .73$\pm$.17          & \textbf{.57$\pm$.09} \\ \hline
        \textbf{0.90}                 & .09$\pm$.07          & .24$\pm$.13          & .06$\pm$.06          & .07$\pm$.05          & .24$\pm$.12        & .05$\pm$.04          & .55$\pm$.22          & .53$\pm$.21          & .50$\pm$.22          \\ \hline
        \multicolumn{10}{|c|}{\textbf{UUC: Low Vegetation}}                                                                                                                                                                                        \\ \hline
        \multirow{2}{*}{\textbf{TPR}} & \multicolumn{3}{c|}{\textbf{SoftMax$^{\mathcal{T}}$}}              & \multicolumn{3}{c|}{\textbf{OpenFCN}}                            & \multicolumn{3}{c|}{\textbf{OpenIPCS}}                             \\ \cline{2-10} 
                                      & \textbf{$Acc^{K}$}   & \textbf{$Pre^{U}$}   & \textbf{$\kappa$}    & \textbf{$Acc^{K}$}   & \textbf{$Pre^{U}$} & \textbf{$\kappa$}    & \textbf{$Acc^{K}$}   & \textbf{$Pre^{U}$}   & \textbf{$\kappa$}    \\ \hline
        \textbf{Closed}               & \textbf{.83$\pm$.03} & .00$\pm$.00          & \textbf{.44$\pm$.14} & \textbf{.83$\pm$.03} & .00$\pm$.00        & \textbf{.44$\pm$.14} & \textbf{.83$\pm$.03} & .00$\pm$.00          & \textbf{.44$\pm$.14} \\ \hline
        \textbf{0.10}                 & .78$\pm$.04          & .35$\pm$.13          & .43$\pm$.15          & .78$\pm$.05          & .34$\pm$.13        & .42$\pm$.15          & \textbf{.74$\pm$.06} & .24$\pm$.06          & .38$\pm$.17          \\ \hline
        \textbf{0.30}                 & .65$\pm$.13          & \textbf{.36$\pm$.14} & .37$\pm$.19          & .63$\pm$.15          & .35$\pm$.14        & .36$\pm$.21          & .57$\pm$.09          & .28$\pm$.11          & .30$\pm$.17          \\ \hline
        \textbf{0.50}                 & .53$\pm$.16          & \textbf{.36$\pm$.14} & .33$\pm$.19          & .51$\pm$.16          & .35$\pm$.14        & .31$\pm$.19          & .40$\pm$.08          & .30$\pm$.14          & .20$\pm$.13          \\ \hline
        \textbf{0.70}                 & .39$\pm$.16          & \textbf{.36$\pm$.15} & .27$\pm$.17          & .36$\pm$.15          & .35$\pm$.15        & .24$\pm$.16          & .21$\pm$.06          & .31$\pm$.15          & .10$\pm$.08          \\ \hline
        \textbf{0.90}                 & .20$\pm$.12          & .35$\pm$.16          & .16$\pm$.12          & .18$\pm$.10          & .35$\pm$.16        & .14$\pm$.09          & .06$\pm$.02          & .32$\pm$.16          & .02$\pm$.03          \\ \hline
        \multicolumn{10}{|c|}{\textbf{UUC: Tree}}                                                                                                                                                                                                  \\ \hline
        \multirow{2}{*}{\textbf{TPR}} & \multicolumn{3}{c|}{\textbf{SoftMax$^{\mathcal{T}}$}}              & \multicolumn{3}{c|}{\textbf{OpenFCN}}                            & \multicolumn{3}{c|}{\textbf{OpenIPCS}}                             \\ \cline{2-10} 
                                      & \textbf{$Acc^{K}$}   & \textbf{$Pre^{U}$}   & \textbf{$\kappa$}    & \textbf{$Acc^{K}$}   & \textbf{$Pre^{U}$} & \textbf{$\kappa$}    & \textbf{$Acc^{K}$}   & \textbf{$Pre^{U}$}   & \textbf{$\kappa$}    \\ \hline
        \textbf{Closed}               & \textbf{.80$\pm$.10} & .00$\pm$.00          & .51$\pm$.11          & \textbf{.80$\pm$.10} & .00$\pm$.00        & .51$\pm$.11          & \textbf{.80$\pm$.10} & .00$\pm$.00          & .51$\pm$.11          \\ \hline
        \textbf{0.10}                 & .77$\pm$.10          & .25$\pm$.05          & .50$\pm$.11          & .77$\pm$.09          & .26$\pm$.07        & .50$\pm$.10          & .79$\pm$.09          & \textbf{.34$\pm$.12} & \textbf{.52$\pm$.10} \\ \hline
        \textbf{0.30}                 & .69$\pm$.10          & .25$\pm$.05          & .47$\pm$.11          & .69$\pm$.09          & .25$\pm$.06        & .48$\pm$.10          & .69$\pm$.09          & .26$\pm$.08          & .47$\pm$.13          \\ \hline
        \textbf{0.50}                 & .59$\pm$.11          & .25$\pm$.05          & .42$\pm$.11          & .59$\pm$.09          & .25$\pm$.05        & .43$\pm$.10          & .54$\pm$.13          & .23$\pm$.07          & .36$\pm$.16          \\ \hline
        \textbf{0.70}                 & .49$\pm$.13          & .25$\pm$.05          & .38$\pm$.12          & .47$\pm$.11          & .24$\pm$.05        & .36$\pm$.10          & .31$\pm$.13          & .19$\pm$.05          & .19$\pm$.12          \\ \hline
        \textbf{0.90}                 & .39$\pm$.12          & .25$\pm$.04          & .34$\pm$.10          & .32$\pm$.10          & .23$\pm$.04        & .27$\pm$.09          & .10$\pm$.06          & .18$\pm$.04          & .06$\pm$.04          \\ \hline
        \multicolumn{10}{|c|}{\textbf{UUC: Car}}                                                                                                                                                                                                   \\ \hline
        \multirow{2}{*}{\textbf{TPR}} & \multicolumn{3}{c|}{\textbf{SoftMax$^{\mathcal{T}}$}}              & \multicolumn{3}{c|}{\textbf{OpenFCN}}                            & \multicolumn{3}{c|}{\textbf{OpenIPCS}}                             \\ \cline{2-10} 
                                      & \textbf{$Acc^{K}$}   & \textbf{$Pre^{U}$}   & \textbf{$\kappa$}    & \textbf{$Acc^{K}$}   & \textbf{$Pre^{U}$} & \textbf{$\kappa$}    & \textbf{$Acc^{K}$}   & \textbf{$Pre^{U}$}   & \textbf{$\kappa$}    \\ \hline
        \textbf{Closed}               & \textbf{.68$\pm$.11} & .00$\pm$.00          & .56$\pm$.13          & \textbf{.68$\pm$.11} & .00$\pm$.00        & .56$\pm$.13          & \textbf{.68$\pm$.11} & .00$\pm$.00          & .56$\pm$.13          \\ \hline
        \textbf{0.10}                 & \textbf{.68$\pm$.12} & .15$\pm$.10          & .56$\pm$.14          & \textbf{.68$\pm$.11} & .16$\pm$.09        & .56$\pm$.13          & \textbf{.68$\pm$.11} & \textbf{.58$\pm$.29} & \textbf{.57$\pm$.13} \\ \hline
        \textbf{0.30}                 & .65$\pm$.12          & .05$\pm$.03          & .53$\pm$.13          & .65$\pm$.11          & .06$\pm$.03        & .54$\pm$.13          & \textbf{.68$\pm$.11} & .35$\pm$.19          & \textbf{.57$\pm$.14} \\ \hline
        \textbf{0.50}                 & .60$\pm$.12          & .03$\pm$.02          & .49$\pm$.13          & .62$\pm$.10          & .04$\pm$.02        & .51$\pm$.12          & .67$\pm$.11          & .20$\pm$.11          & .56$\pm$.14          \\ \hline
        \textbf{0.70}                 & .52$\pm$.10          & .03$\pm$.01          & .42$\pm$.10          & .56$\pm$.09          & .03$\pm$.02        & .45$\pm$.10          & .65$\pm$.11          & .10$\pm$.06          & .54$\pm$.13          \\ \hline
        \textbf{0.90}                 & .38$\pm$.08          & .02$\pm$.01          & .30$\pm$.07          & .44$\pm$.08          & .02$\pm$.01        & .34$\pm$.08          & .58$\pm$.10          & .05$\pm$.03          & .47$\pm$.11          \\ \hline
    \end{tabular}
    }
\end{table}

Discussion regarding Potsdam results closely resembles the ones presented from Vaihingen on Section~\ref{sec:quantitative_vaihingen}, albeit with one major distinction: relatively poor performance in comparison with Vaihingen due to the hardships encountered in processing Potsdam samples. These difficulties include the massive size of the dataset, restricting tuning of the networks' hyperparameters due to computational constraints; the larger spatial resolution ($6000 \times 6000$ per image) resulting from smaller physical areas covered by each pixel in comparison with Vaihingen, which also accentuate the downsides of introducing a larger correlation between adjacent pixels and providing less context on a $224 \times 224$ patch; and an extremely imbalanced set of labels.

One can indeed observe significant $\kappa$ improvements by introducing OpenIPCS to the segmentation procedure in UUCs \textit{Impervious Surfaces}, \textit{Building}, \textit{Tree} and \textit{Car}, mainly with a DenseNet-121 backbone (Table~\ref{tab:tpr_potsdam_densenet121}). Closed Set $\kappa$ values for these classes were 0.42, 0.41, 0.51 and 0.56, while the best OSR results (assuming optimal TPR choice) increased these values to 0.55, 0.57, 0.52 and 0.57. Analogously to Vaihingen, the increases to $\kappa$ were larger for \textit{Impervious Surfaces} and \textit{Building} (between 0.13 and 0.16), while \textit{Tree} and \textit{Car} achieve much more modest gains of only 0.01, which are not statistically significant. \textit{Building} again achieved a close to perfect $Pre^{U}$ (0.96) using DenseNet-121 and TPR 0.10, followed by the also high \textit{Impervious Surfaces} $Pre^{U}$ of 0.87 also with 0.10 of TPR.

\textit{Low Vegetation} again was an outlier in the LOCO protocol using DenseNet-121, with \softmaxt{} and OpenFCN achieving less degradation in $Acc^{K}$ and a larger $Pre^{U}$ for the same TPR values. Since \textit{Low Vegetation} fills a much larger proportion of pixels in Potsdam than in Vaihingen, it would be expected that adding OSR would considerably improve $\kappa$ values in comparison with Closed Set when this class was set to UUC in the LOCO protocol. However, the best $\kappa$ results were the Closed Set ones for all OSR methods on the DenseNet-121 backbone, implying that neither of the methods converged well enough on this architecture to compensate for the KKC prediction losses. \textit{Low Vegetation} results were opposite on the WRN-50 backbone in Potsdam: $\kappa$ increased by 0.06 (for TPR 0.70); the best $Pre^{U}$ results were indeed from OpenIPCS, peaking at 0.56 with TPR 0.30 and $Acc^{K}$ degradation was much slower than OpenFCN and \softmaxt{} as TPRs were allowed to grow. These discrepancies between architectures highlights that, even though OpenPCS and OpenIPCS were found to be the current state-of-the-art for OSR segmentation, they can be highly dependent on the DNN architecture, hyperparameters and proved to be rather unstable, being still a work in progress.

For almost all UUCs, $Acc^{K}$ degraded much faster on Potsdam than on Vaihingen, as larger TPRs were evaluated. One important outlier in almost all aspects presented previously was \textit{Car} on Potsdam. While on Vaihingen the $\kappa$ did not increase at all in comparison to the Closet Set DNNs with the introduction of OpenPCS, the $\kappa$ value for \textit{Car} had a slight, but noticeable, increase from 0.56 to 0.57. $Pre^{U}$ values for Car segmentation were also considerably larger on Potsdam, peaking at 0.58 for TPR 0.10 using DenseNet-121 and 0.29 using WRN-50. At the same time, $Acc^{K}$ remained close to the Closed Set counterpart with TPRs up to 0.70, only observing a noticeable decrease in KKC classification performance on TPR 0.90. As for OpenFCN and \softmaxt{}, $Pre^{U}$ results for \textit{Car} were also higher and $Acc^{K}$ also degraded slowly with the increase of TPR (comparing with results from Vaihingen), even if these metrics did not achieve the same gains as the ones from OpenPCS.

\subsubsection{Qualitative Analysis}
\label{sec:qualitative}

\begin{figure*}[!ht]
	\centering
	\includegraphics[width=\textwidth]{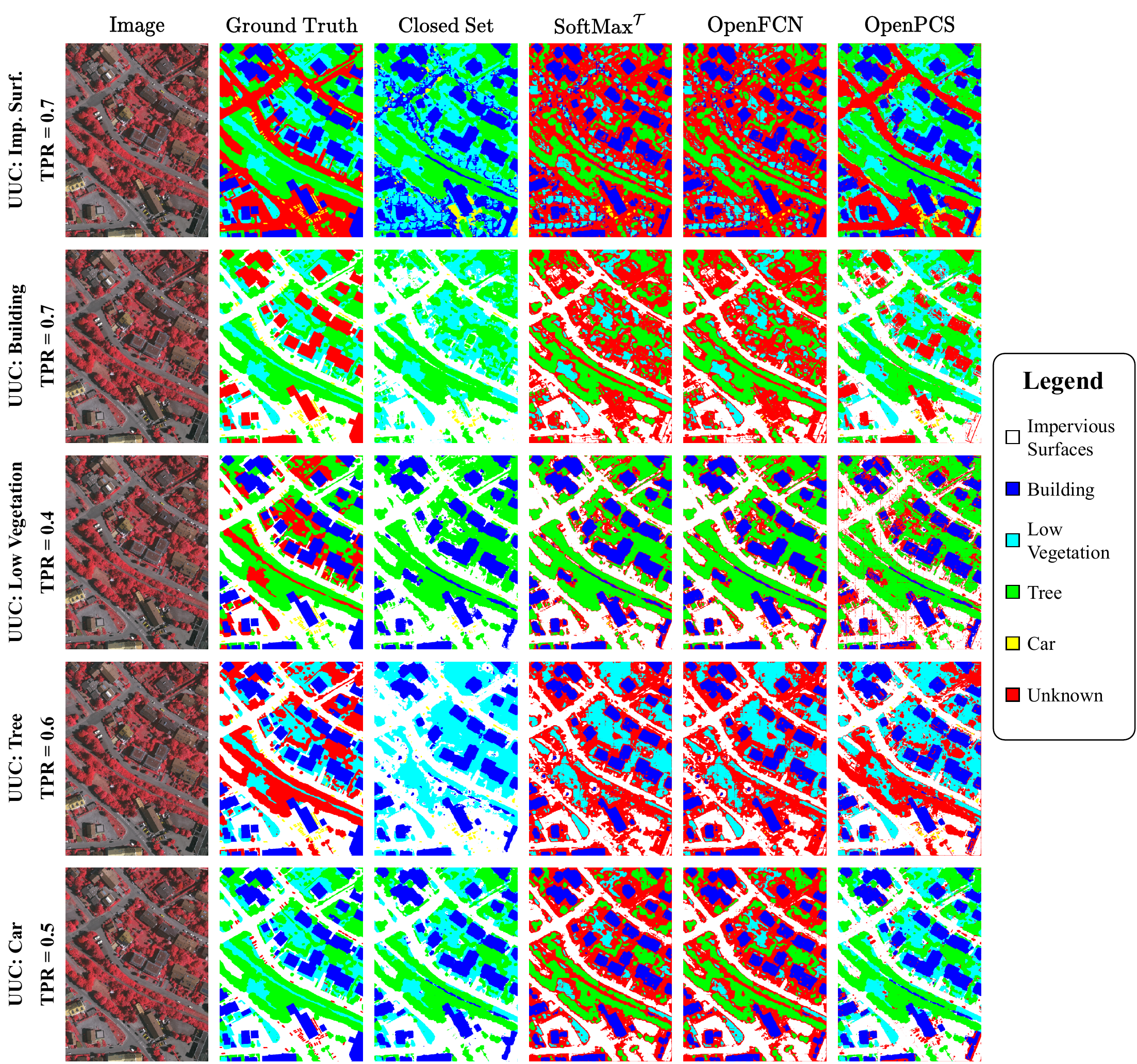}
	\caption{Some visual result samples obtained for the \textbf{Vaihingen} dataset according to distinct UUCs in the LOCO protocol and distinct TPR thresholds for \softmaxt{}, OpenFCN and OpenPCS.}
	\label{fig:qualitative_per_uucs_vaihingen}
\end{figure*}

\begin{figure*}[!ht]
	\centering
	\includegraphics[width=\textwidth]{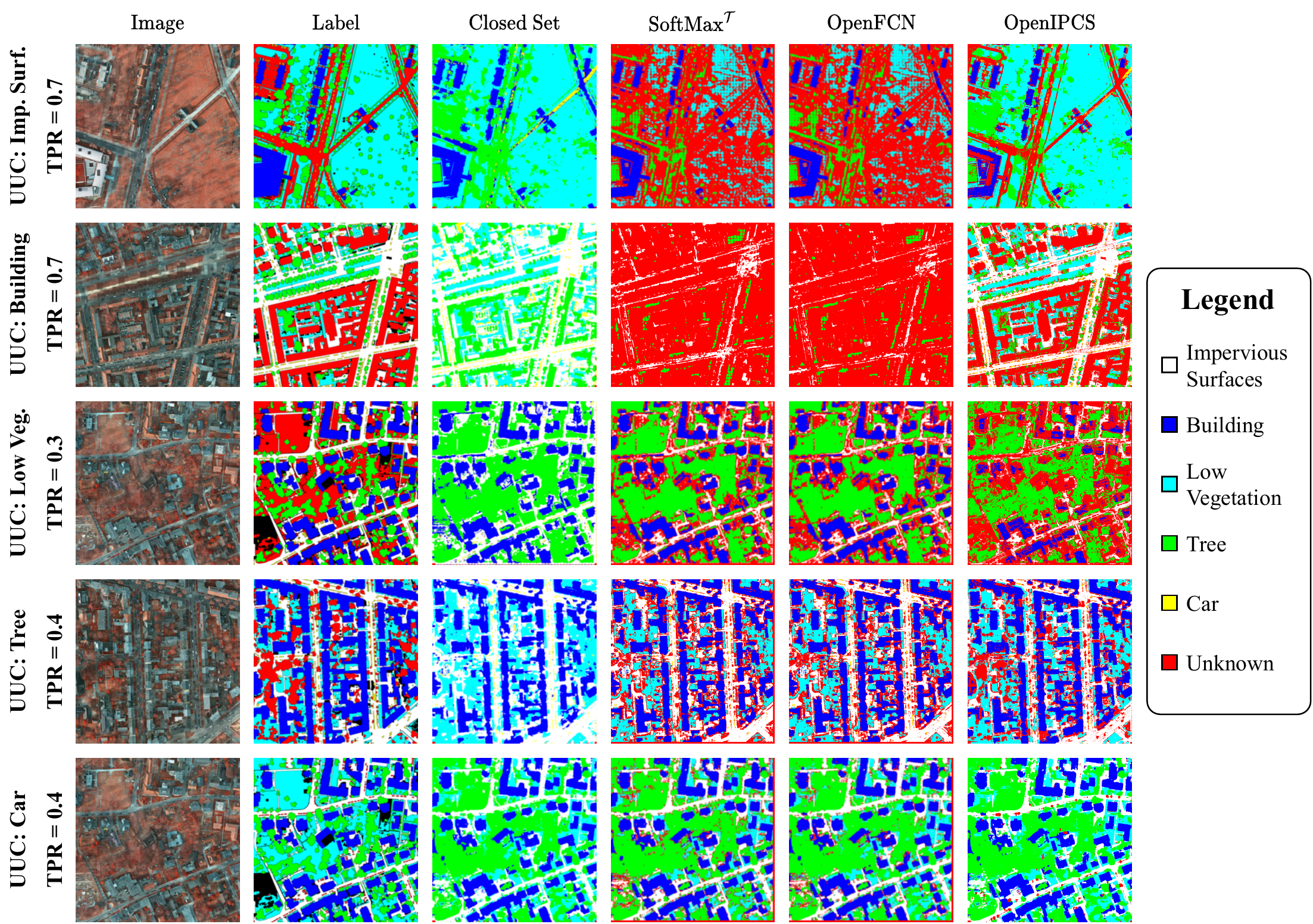}
	\caption{Some visual result samples obtained for the \textbf{Potsdam} dataset according to distinct UUCs in the LOCO protocol and distinct TPR thresholds for \softmaxt{}, OpenFCN and OpenIPCS.}
	\label{fig:qualitative_per_uucs_potsdam}
\end{figure*}

Figures~\ref{fig:qualitative_per_uucs_vaihingen} and~\ref{fig:qualitative_per_uucs_potsdam} present some visual examples of the results generated by the FCN with DenseNet-121 backbone in the Vaihingen and Potsdam datasets respectively. Qualitatively, the effectiveness of OpenPCS to distinguish KKCs from UUCs is even clearer. As can be observed, OpenPCS is capable of producing more accurate UUC identification for both KKCs and UUCs, when compared to the ground-truth label. On the other hand, outcomes generated by the \softmaxt{} and OpenFCN are very similar to each other, but not very alike to the ground-truth label, mainly for the UUCs. This corroborates with previous analysis and conclusions about the effectiveness of the OpenPCS mainly for discriminating UUCs.

Performing a deep analysis of the qualitative results, we can see that the proposed OpenPCS identified the UUC \textit{Building} almost perfectly, while the \softmaxt{} and OpenFCN techniques quite confused this UUC with other KCCs using the same threshold TPR. This same outcome can be seen for the UUC \textit{Car}. Although this UCC comprises only a tiny percentage of the total amount of pixels and does not contribute a lot in terms of $\kappa$, it was much better identified by the OpenPCS than by other approaches. In fact, the above outcomes are repeated for all UUCs, except for the \textit{Low-Vegetation} one, which has enormous intra-class variability (with pixels of grass fields, sidewalk-like areas and other structures) and, consequently, natural erratic behaviour. For a more detailed qualitative analysis, please, check the project's webpage.

\subsubsection{Runtime Performance Analysis}
\label{sec:time_comparison}

\begin{figure*}[!ht]
	\centering
	\renewcommand{\currprop}{0.49\textwidth}
	\subfloat[Linear time Vaihingen]{
		\includegraphics[width=\currprop]{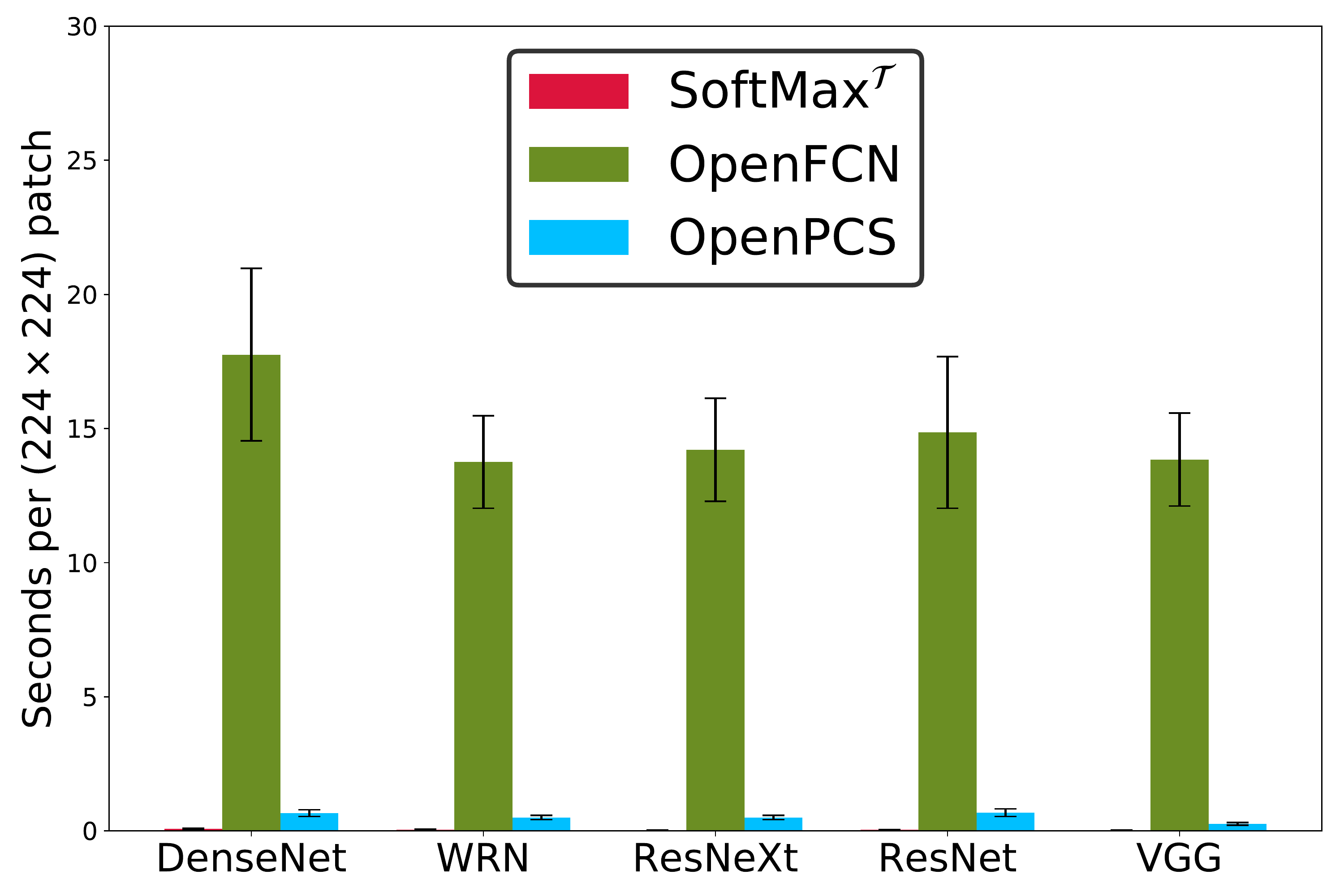}
		\label{fig:time_comparison_linear_vaihingen}
	}
	\subfloat[Linear time Potsdam]{
		\includegraphics[width=\currprop]{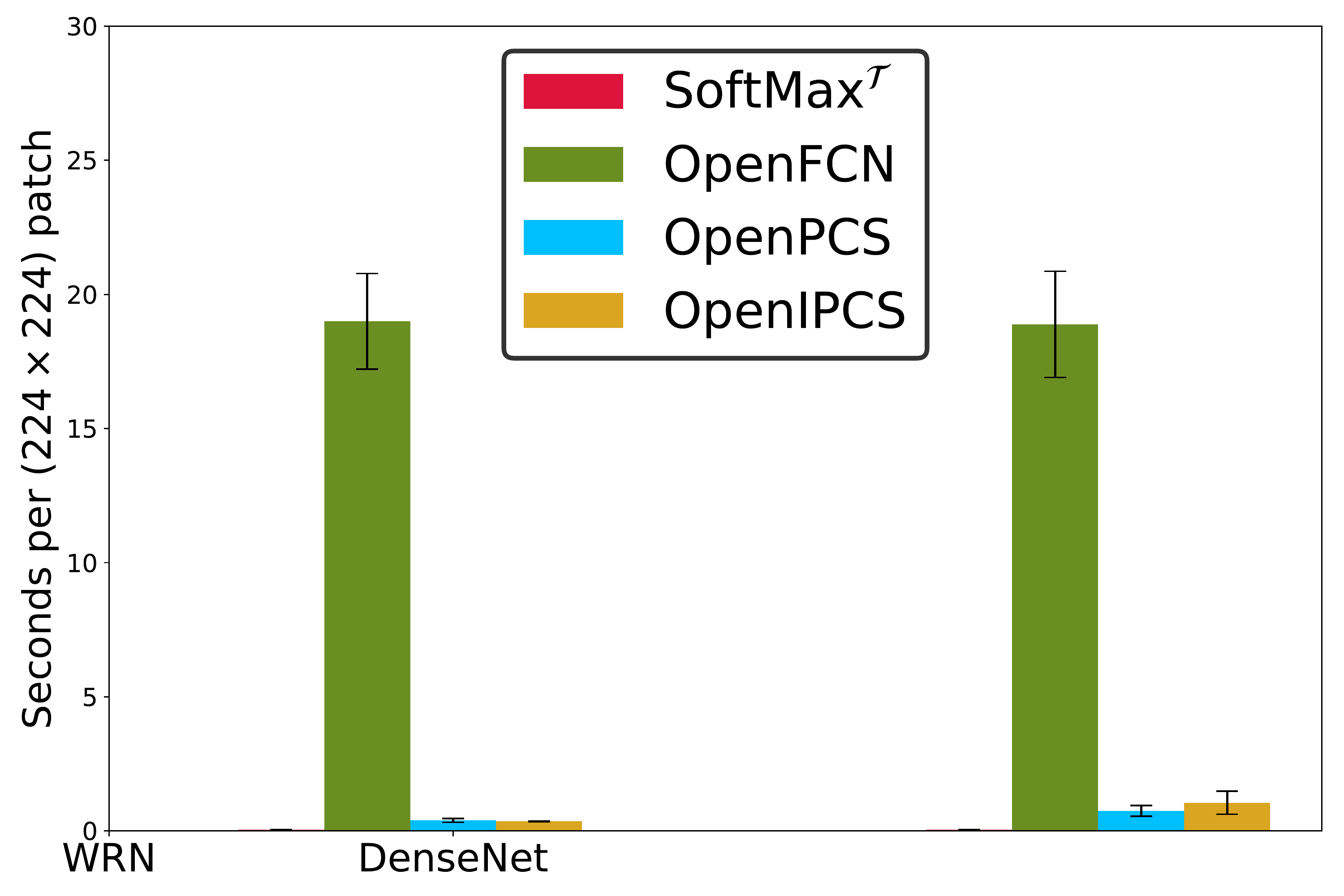}
		\label{fig:time_comparison_linear_potsdam}
	}
	\\
	\subfloat[Log time Vaihingen]{
		\includegraphics[width=\currprop]{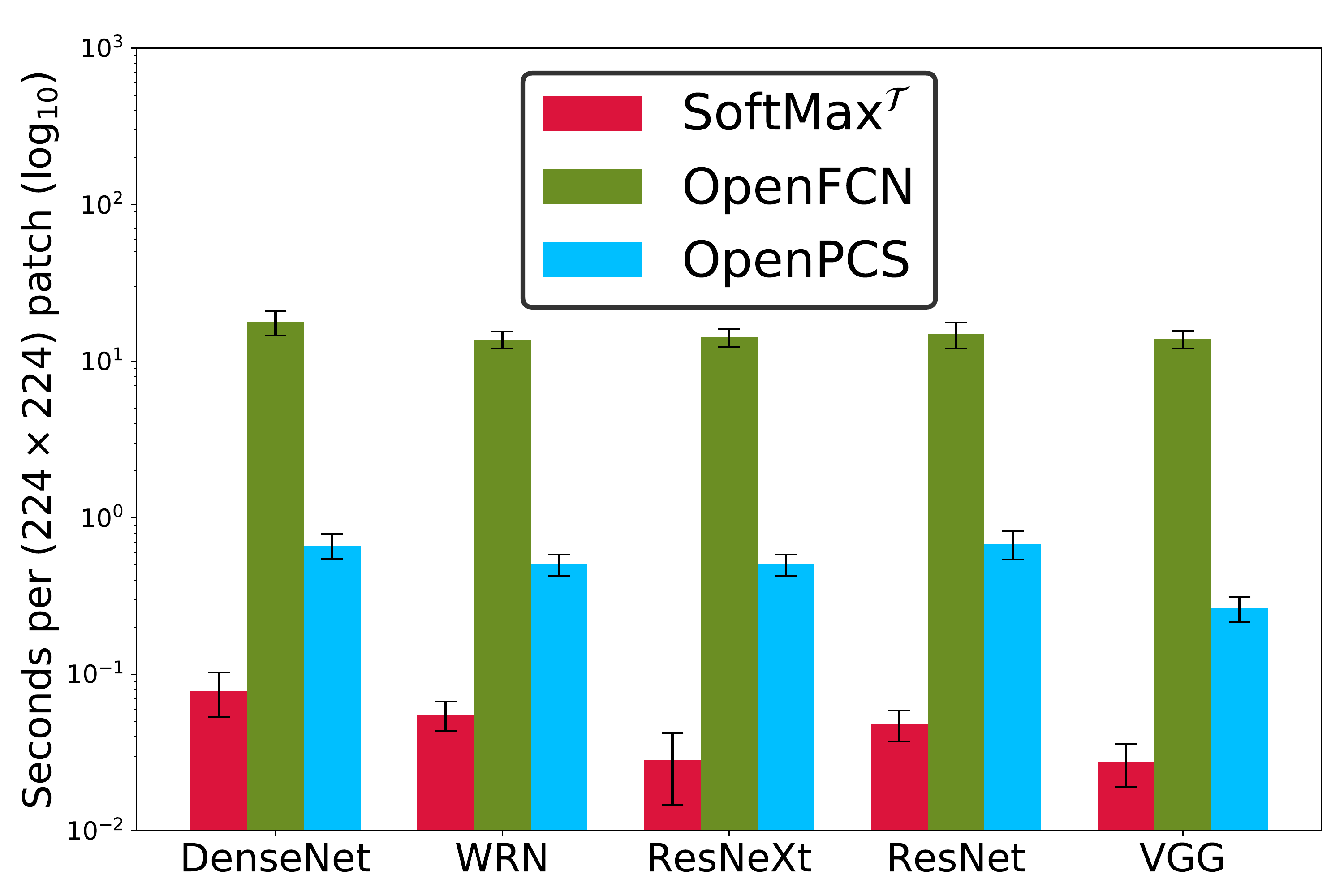}
		\label{fig:time_comparison_log_vaihingen}
	}
	\subfloat[Log time Potsdam]{
		\includegraphics[width=\currprop]{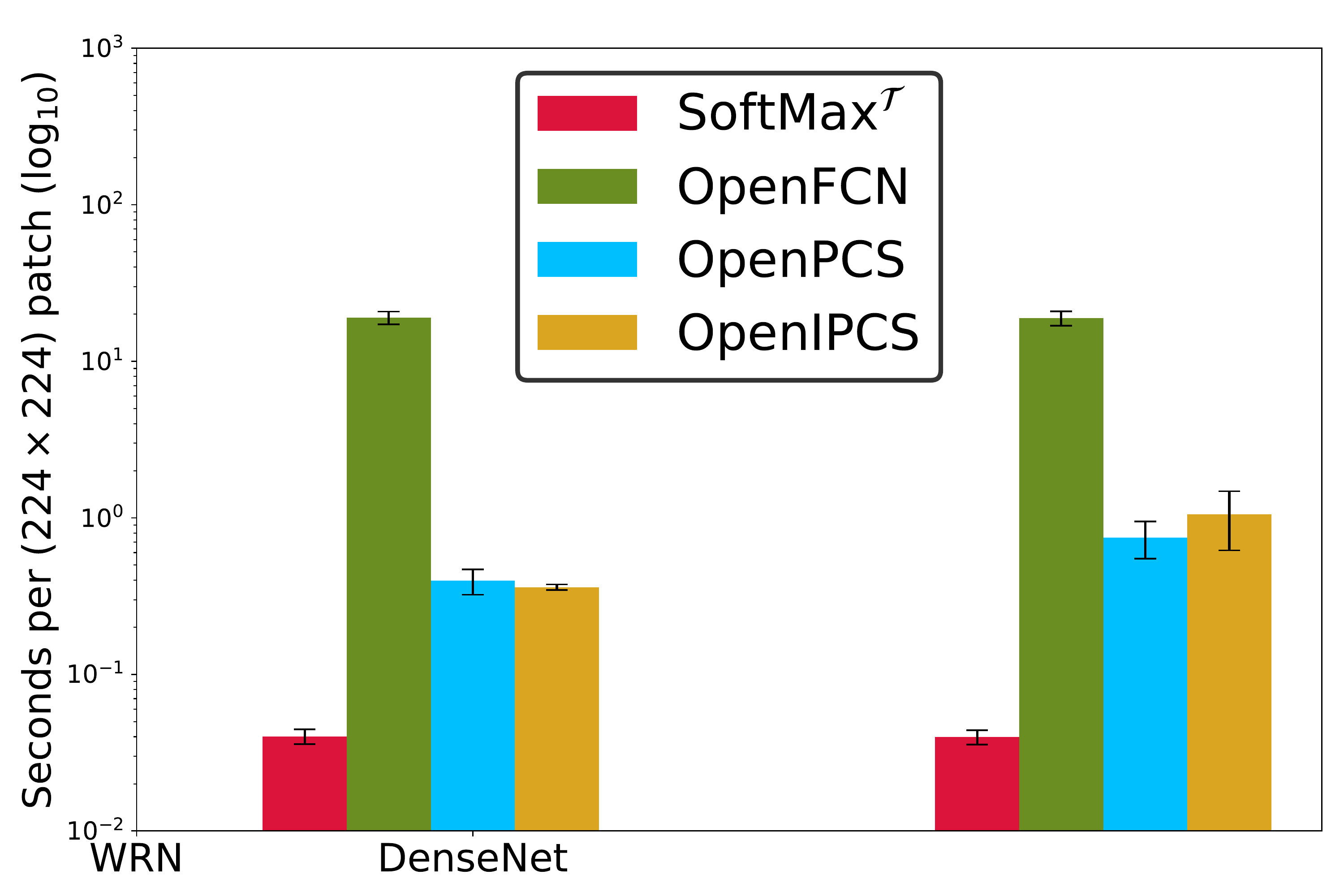}
		\label{fig:time_comparison_log_potsdam}
	}
	\caption{Per-patch time comparison between the proposed approaches. The time presented in the $y$-axis is shown in seconds for each ($224 \times 224$) patch across all test patches. The left part of the figure (a, c) shows times for Vaihingen, while the rightmost figures (b, d) depict times for Potsdam. Results are also shown in linear (a, b) and $log_{10}$ scale (c, d) in order to show the exponential distinction between execution times across \softmaxt{}, OpenFCN and OpenPCS. Confidence intervals for each plot are shown as error bars computed with the t-Student distribution on the average execution runtimes for patches over 5 runs of the LOCO protocol (one for each class set as UUC) for each backbone.}
	\label{fig:time_comparison}
\end{figure*}

One of the most important motivations for proposing OpenPCS was the observation that OpenMax did not scale well to dense labeling, being prohibitively expensive when operating in a pixelwise fashion, while the simple \softmaxt{} performed exponentially faster. In order to properly quantify the time differences between \softmaxt{}, OpenFCN and OpenPCS, we computed the per-patch runtimes of each method for $224 \times 224$ patch resolution. These results are shown in Figure~\ref{fig:time_comparison} for both Vaihingen and Potsdam (Figures~\ref{fig:time_comparison_linear_vaihingen} and Figures~\ref{fig:time_comparison_linear_potsdam}, respectively) and these same results are also presented in $\log_{10}$ scale (Figures~\ref{fig:time_comparison_log_vaihingen} and Figures~\ref{fig:time_comparison_log_potsdam}, respectively), as the linear time comparisons severely hampered the visualization of \softmaxt{}'s performance when compared to OpenFCN.

Visual analysis of Figure~\ref{fig:time_comparison} reveals the discrepancies between OSR methods, with \softmaxt{} inference being the fastest, usually taking between 0.03 and 0.1 second per $224 \times 224$ patch. On the opposite side, OpenFCN was observed to be by far the slowest method, with runtimes for one single patch in the range between 10 and 20 seconds. OpenFCN's runtime makes it highly unsuitable for real-time computer vision applications.

On the faster end of the spectrum between \softmaxt{} and OpenFCN, there were OpenPCS and OpenIPCS, with execution times between 0.25 and 0.7 seconds per patch. Both the online and offline PCAs used for the inference of UUCs from the proposed methods are highly vectorized operations, which allows them to be parallelized into several processing cores and be faster even in single-core architectures. While we did not tune the algorithms for this purpose, OpenPCS' and OpenIPCS' runtimes allow for near real-time inference on applications as self-driving cars or autonomous drone control. In contrast, the original implementation of OpenMax from libMR is not naturally vectorized, requiring inferences to be performed linearly on pixels and severely hampering OpenFCN's performance.

\section{Conclusion}
\label{sec:conclusion}

In this manuscript we introduced, formally defined and explored the problem of Open Set Segmentation and proposed two approaches for solving OSR segmentation tasks: 1) OpenFCN, which is based on the well-known OpenMax~\cite{Bendale:2016} approach; and 2) OpenPCS, a completely novel method for Open Set Semantic Segmentation. A comprehensive evaluation protocol based on a set of standard objective metrics to be used on this novel field that takes into account KKCs and UUCs was proposed, executed and discussed for the proposed methods and the main baseline: \softmaxt{}.

OpenPCS -- and its more scalable variant referred to as OpenIPCS -- yielded significantly better results than OpenFCN and \softmaxt{}, suggesting that using middle-level feature information is highly useful for OSR segmentation. OpenFCN showed relatively poor performance in most experiments, as it is limited to using the activations of the last layer to detect OODs. This scheme works well for sparse labeling scenarios~\cite{Bendale:2016} (i.e. image classification), however, dense labeling inherently consists of highly correlated samples -- in this case, pixels -- and labels -- in this case, semantic maps -- with border regions naturally presenting lower confidence than the interior pixels of objects. This distinction between sparse and dense labeling severely hampered the capabilities of OpenFCN and \softmaxt{} of detecting OOD samples.

As expected, UUCs with larger semantic and visual distinctions in relation to the remaining KKCs also showed considerably better classification performance in the LOCO protocol. Specifically, \textit{Low Vegetation} samples have a high intra-class variability, representing grass fields, sidewalk-like areas and other structures in the Vaihingen and Potsdam datasets, rendering them similar to samples from the classes \textit{Street} and \textit{Tree}. This resulted in lower performances in almost all metrics for all methods assessed in our experimental procedure. Closed Set predictions by the DNN architectures evaluated in Section~\ref{sec:baseline} generally resulted in lower overall $\kappa$ results than when OSR was added in the form of OpenPCS mainly, with gains up to 0.20 in this metric for the \textit{Building} UUC. The analysis conducted in Section~\ref{sec:time_comparison} also revealed the discrepancies between OSR methods in inference runtime, with \softmaxt{} being the fastest method, OpenFCN being the exponentially slower and OpenPCS being a reasonable compromise, being able to run close to real-time, depending on GPU memory for larger patches and/or image size.

As OpenFCN is simply a fully convolutional version of OpenPixel~\cite{Silva:2020}, also based on OpenMax~\cite{Bendale:2016}, we can confidently infer that even OpenFCN performed better than OpenPixel. That is because OpenPixel was based on a traditional CNNs trained in a patchwise fashion for the classification of the central pixel and fully convolutional training was already shown to improve both segmentation performance and dramatically improve runtime efficiency of patchwise approaches in the pattern recognition literature~\cite{Long:2015}. While OpenFCN differs in this aspect, it even uses the same meta-recognition lib and some of the code of OpenPixel for the computation of OpenMax. One important variation of OpenPixel's approach that still needs to be verified in the context of fully convolutional training is the post processing using morphology, called Morph-OpenPixel by Silva \textit{et al.}~\cite{Silva:2020}. This post-processing was shown to considerably improve the performance of OpenPixel. At last, even though time comparisons were not made available by Silva \textit{et al.}~\cite{Silva:2020}, it is also possible to infer that OSR segmentation using even OpenFCN is exponentially faster than patchwise training also due to the experiments performed on the original FCN paper~\cite{Long:2015}.

Approaches as CGDL~\cite{Sun:2020} and C2AE~\cite{Oza:2019} couple the training of the generative model (usually a VAE) with the supervised training for KKC classification. These approaches allow for and end-to-end training of DNNs for OSR image classification tasks, and are currently considered the state-of-the-art for this task. Future works for OpenPCS include the adaptation of these methods for Open Set Semantic Segmentation, which would incorporate low-, middle- and high-level semantic features into the recognition of OOD samples, possibly allowing for an end-to-end training of Pixel Anomaly Detection and OSR segmentation. We also aim to investigate the capabilities of other lighter and/or more robust generative models in the OpenPCS pipeline, such as One-Class SVM (OCSVM)~\cite{Scholkopf:2001,Scheirer:2012} or Gaussian Mixture Models (GMMs)~\cite{Bishop:2006,Attias:2000} instead of the simpler Principal Components used for computing OpenPCS likelihoods.

Authors also aim to perform future analysis by varying the Openness of the problems -- that is, setting more than one class to UUC in a variation of the LOCO protocol. Early experiments indicate that the superiority of OpenPCS might be even more pronounced in settings with larger Openness and/or larger number of known classes. 

\newpage
\section*{Acknowledgments}
The authors would like to thank NVIDIA for the donation of the GPUs that allowed the execution of all experiments in this paper.





\bibliographystyle{spmpsci}      
\bibliography{refs}   

\clearpage

\end{document}